%% file: main.tex
\documentclass[10pt,twocolumn,letterpaper]{article}

\usepackage[pagenumbers]{cvpr} % To force page numbers, e.g. for an arXiv version

\input{preamble}

\definecolor{cvprblue}{rgb}{0.21,0.49,0.74}
\usepackage[pagebackref,breaklinks,colorlinks,allcolors=cvprblue]{hyperref}

\def\paperID{} % *** Enter the Paper ID here
\def\confName{CVPR}
\def\confYear{2026}

\title{Gen-n-Val: Agentic Image Data Generation and Validation}

\author{Jing-En Huang$^{\ast, \dagger, \ddagger}$ \quad I-Sheng Fang$^{\ast, \ddagger}$ \quad Tzuhsuan Huang$^\ddagger$ \\ 
Yu-Lun Liu$^\dagger$ \quad  Chih-Yu Wang$^\ddagger$ \quad Jun-Cheng Chen$^\ddagger$\\
$^\dagger$ National Yang Ming Chiao Tung University \\
$^\ddagger$Research Center for Information Technology Innovation, Academia Sinica 
\\
}

\newcommand\blfootnote[1]{%
  \begingroup
  \renewcommand\thefootnote{}\footnote{#1}%
  \addtocounter{footnote}{-1}%
  \endgroup
}

\begin{document}
% \maketitle

\twocolumn[{%
\renewcommand\twocolumn[1][]{#1}%
\maketitle
\vspace{-2em}
 \begin{center}

% \begin{tabular}{@{}c@{}c@{}c@{}}
\resizebox{\textwidth}{!}{
\begin{tabular}{cc}
 Our Synthetic Data & Original (a) \& Balanced (\textcolor{orange}{b}, \textcolor{SeaGreen}{c}, \textcolor{blue}{d}) LVIS Category Distribution (Log Scale)\\
 \includegraphics[width=.3\textwidth]{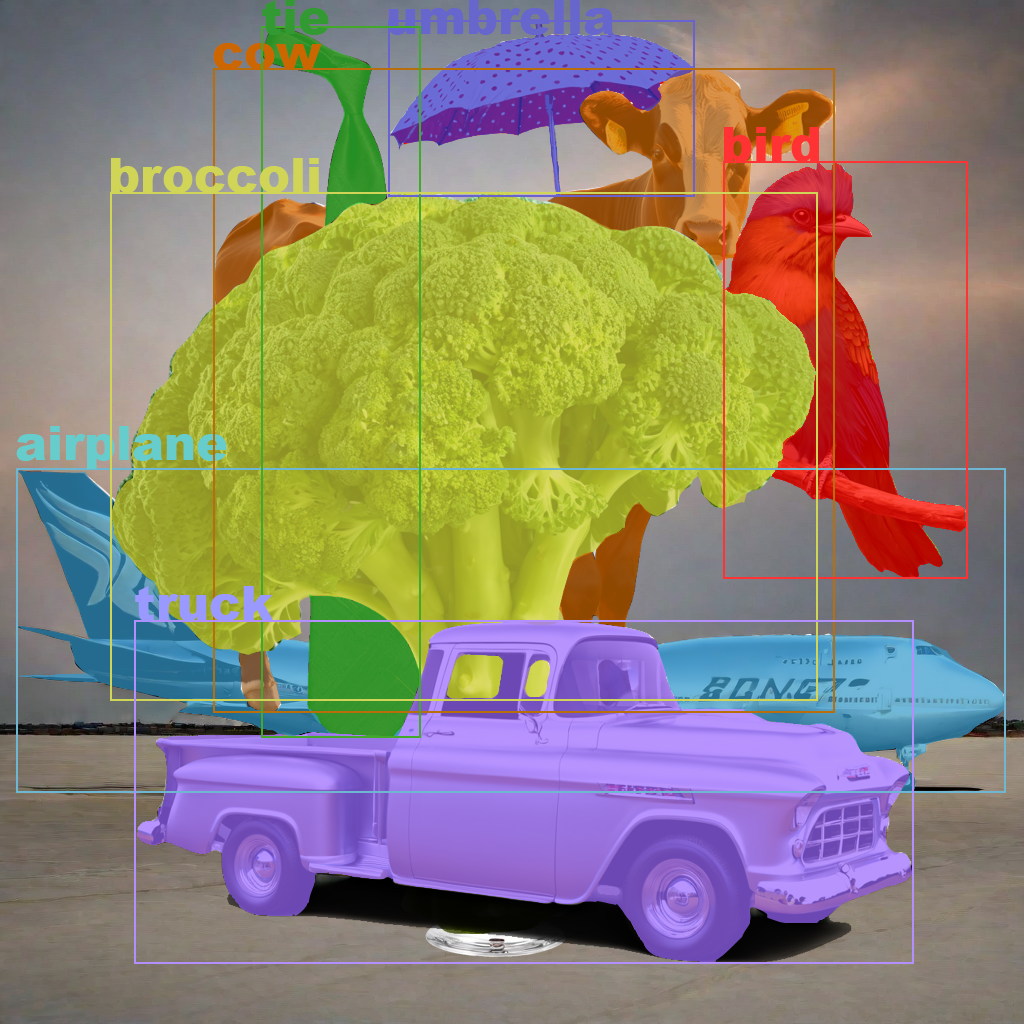} &  
 \includegraphics[width=.72\textwidth]{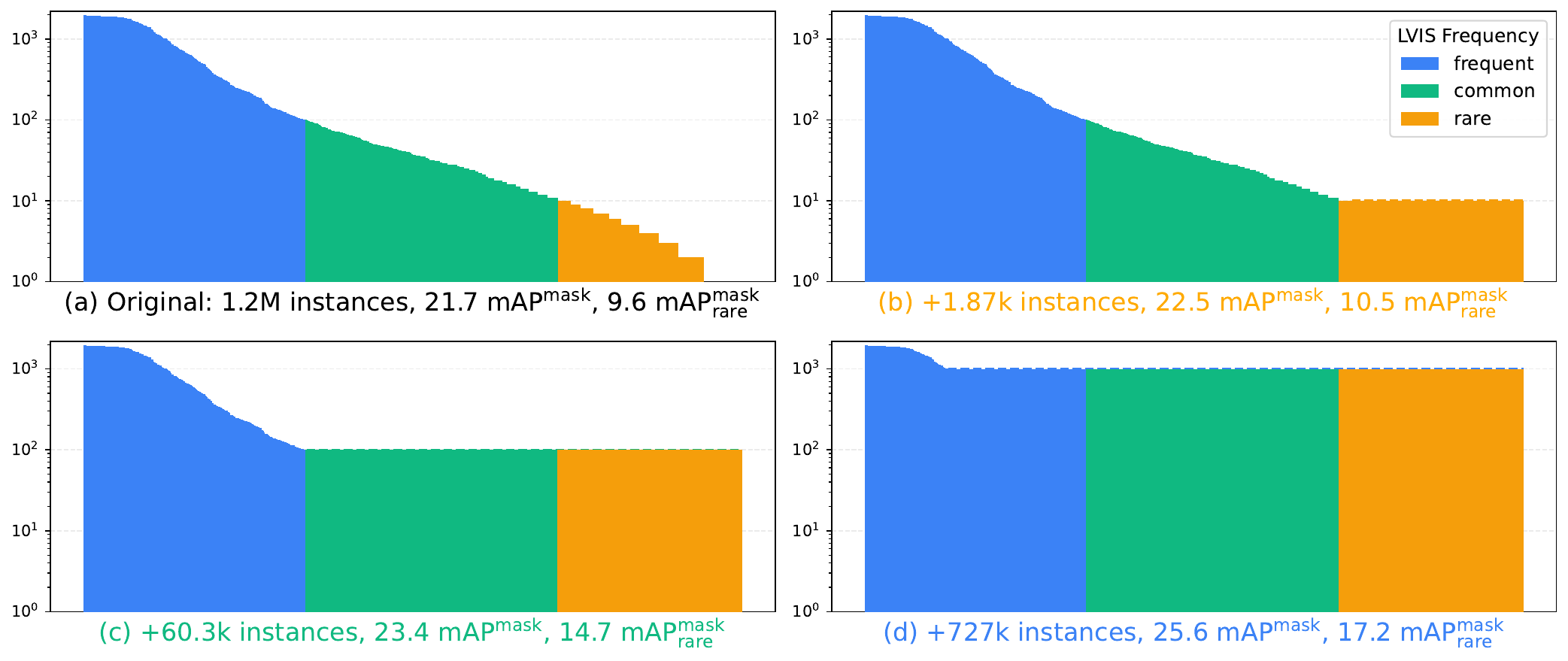} \\
\end{tabular}
}
    \vspace{-3mm}
\captionof{figure}{
\textbf{Gen-n-Val generates high-quality synthetic data to address long-tailed distribution in instance segmentation.}
\emph{(Left)} Sample synthetic data generated by Gen-n-Val with precise instance masks (different colors indicate different object categories).
\emph{(Right)} Log-scale category distribution of LVIS~\cite{gupta2019lvis}, a dataset known for its severe long-tailed distribution where rare categories (\textcolor{orange}{orange}) often have fewer than 10 images. With Gen-n-Val, we inject \textcolor{orange}{1,874}, \textcolor{SeaGreen}{60,306}, and \textcolor{blue}{727,393} additional instances for rarer categories, substantially balancing the long-tailed distribution such that each category has at least \textcolor{orange}{10}, \textcolor{SeaGreen}{100}, and \textcolor{blue}{1,000} images. This redistribution leads to gains of \textcolor{orange}{+0.8}, \textcolor{SeaGreen}{+1.7}, and \textcolor{blue}{+3.9} mask mAP overall and \textcolor{orange}{+0.9}, \textcolor{SeaGreen}{+5.1}, and \textcolor{blue}{+7.6} mask mAP on rare categories with Mask R-CNN~\cite{he2017maskrcnn}, demonstrating the effectiveness of our agentic data generation and validation pipeline in addressing data scarcity for underrepresented classes.
}
\label{fig:teaser}
\end{center}
}]

\input{sec/0_Abstract}
\vspace{-1em}
\input{sec/1_Introduction}
\input{sec/2_RelatedWork}
\input{sec/3_Method}
\input{sec/4_Experiments}
\input{sec/5_Conclusion}
\input{sec/6_Acknowledgement}

{
    \small
    \bibliographystyle{ieeenat_fullname.bst}
    \bibliography{main.bib}
}


\clearpage
\appendix
\setcounter{table}{0}
\renewcommand{\thetable}{S.\arabic{table}}
\setcounter{figure}{0}    
\renewcommand{\thefigure}{S.\arabic{figure}}
\setcounter{page}{1}
\input{sec/X_suppl}
\section*{Please check the following pages for more tables and figures.}
\input{sec/Y_suppl_large}

\end{document}

%% file: preamble.tex
\usepackage{colortbl}
\usepackage{multirow}
\usepackage{xcolor}
\usepackage{algorithm}
\usepackage{algpseudocode}
\usepackage{soul}

%% file: sec/0_Abstract.tex
\begin{abstract}
\blfootnote{$^\ast$ indicates equal contribution.}
The data scarcity, label noise, and long-tailed category imbalance remain important and unresolved challenges in many computer vision tasks, such as object detection and instance segmentation, especially on large-vocabulary benchmarks like LVIS, where most categories appear in only a few images.
Current synthetic data generation methods still suffer from multiple objects per mask, inaccurate segmentation, incorrect category labels, and other issues, limiting their effectiveness. To address these issues, we introduce Gen-n-Val, a novel agentic data generation framework that leverages Layer Diffusion (LD), a Large Language Model (LLM), and a Vision Large Language Model (VLLM) to produce high-quality and diverse instance masks and images for object detection and instance segmentation.
Gen-n-Val consists of two agents: (1) the LD prompt agent, an LLM, optimizes prompts to encourage LD to generate high-quality foreground single-object images and corresponding segmentation masks; and (2) the data validation agent, a VLLM, filters out low-quality synthetic instance images. The system prompts for both agents are optimized by TextGrad. Compared to state-of-the-art synthetic data approaches like MosaicFusion, our approach reduces invalid synthetic data from 50\% to 7\% and improves performance by 7.6\% on rare classes in LVIS instance segmentation with Mask R-CNN, and by 3.6\% mAP on rare classes in COCO instance segmentation with YOLOv9c and YOLO11m. Furthermore, Gen-n-Val shows significant improvements (7.1\% mAP) over YOLO-Worldv2-M in open-vocabulary object detection benchmarks with YOLO11m.
Moreover, Gen-n-Val has scalability in model capacity and dataset size. The code is available at \url{https://github.com/aiiu-lab/Gen-n-Val}.
\end{abstract}

%% file: sec/1_Introduction.tex
\begin{figure}[t]
    \centering
    \includegraphics[width=\columnwidth]{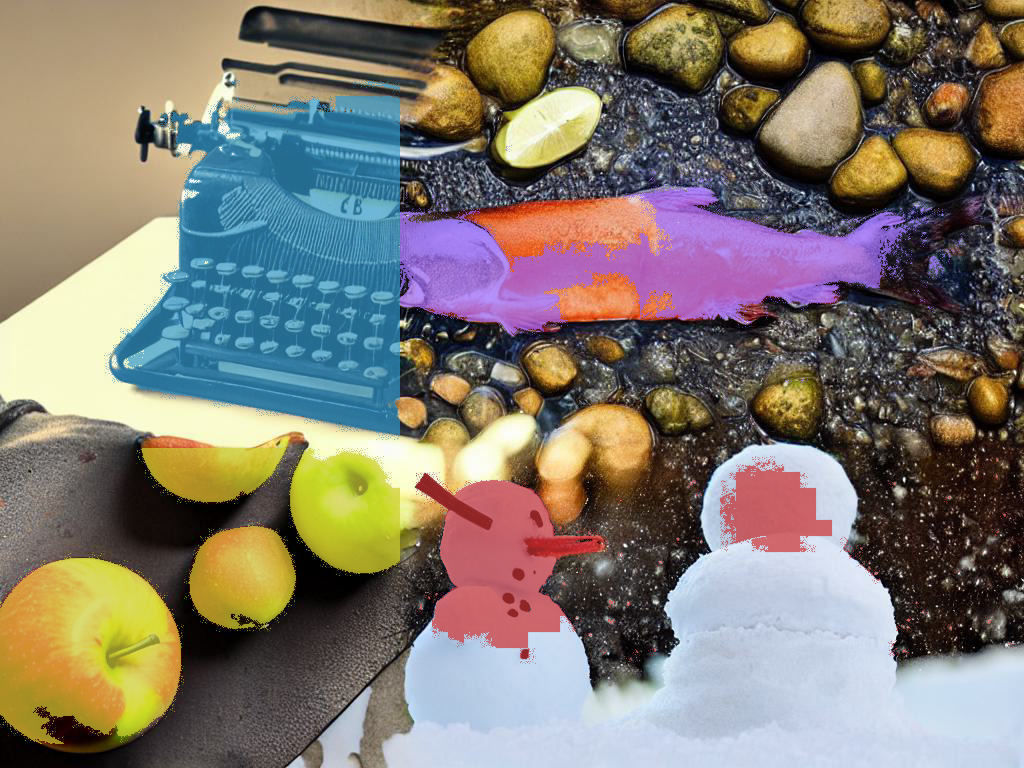}
    \vspace{-6mm}
    \caption{\textbf{Common failures in MosaicFusion~\cite{xie2024mosaicfusion} synthetic data.}
    Despite using prompt ``\texttt{a photo of a single <object>}'', the method produces: (1) incomplete objects (\textcolor{cyan}{blue}: typewriter cut off), (2) incomplete segmentation (\textcolor{violet}{purple}: salmon; \textcolor{red}{red}: snowmen), (3) multiple objects with single mask (\textcolor{red}{red}: two snowmen), and (4) wrong categories (\textcolor{yellow}{yellow}: whole apples labeled ``apple sauce'').}
    \label{fig:mosaic_issue}
    \vspace{-1em}
\end{figure}

\newcommand\LayerDiffIssueTextWidth{0.1\textwidth}
\newcommand\LayerDiffIssueBoxWidth{0.16 \textwidth}
\newcommand\LayerDiffIssueImageWidth{0.16\textwidth}

\begin{figure*}
    \resizebox{\textwidth}{!}{
    \small
    \begin{tabular}{cccccc}
    \parbox[c]{\LayerDiffIssueBoxWidth}{\includegraphics[width=\LayerDiffIssueImageWidth]{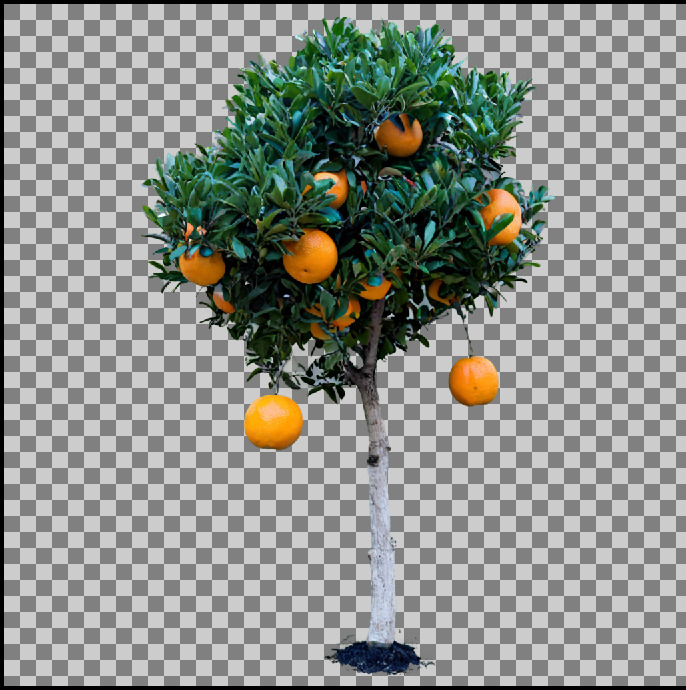}} & 
    \parbox[c]{\LayerDiffIssueBoxWidth}{\includegraphics[width=\LayerDiffIssueImageWidth]{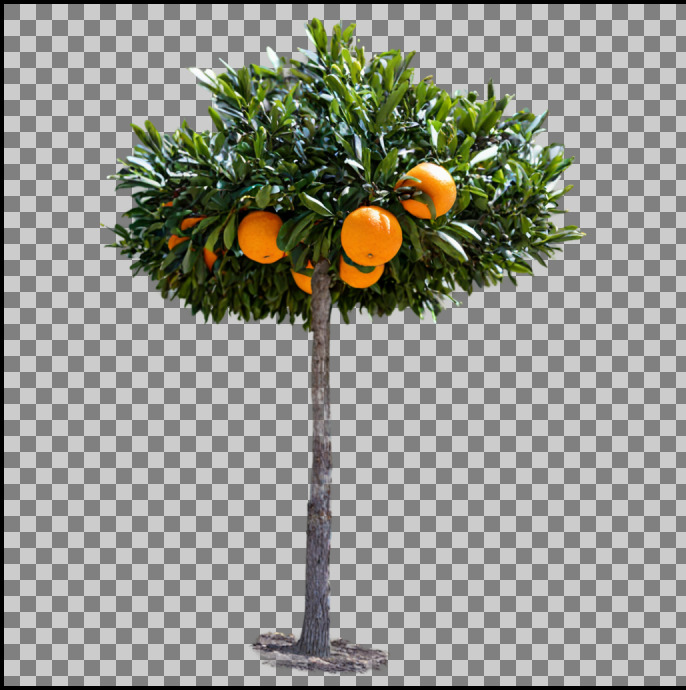}} &
    \parbox[c]{\LayerDiffIssueBoxWidth}{\includegraphics[width=\LayerDiffIssueImageWidth]{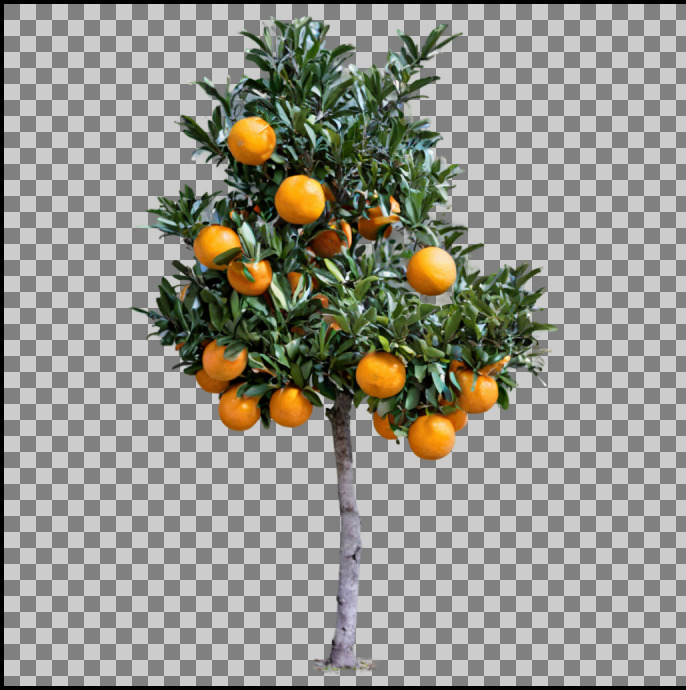}} &   
    \parbox[c]{\LayerDiffIssueBoxWidth}{\includegraphics[width=\LayerDiffIssueImageWidth]{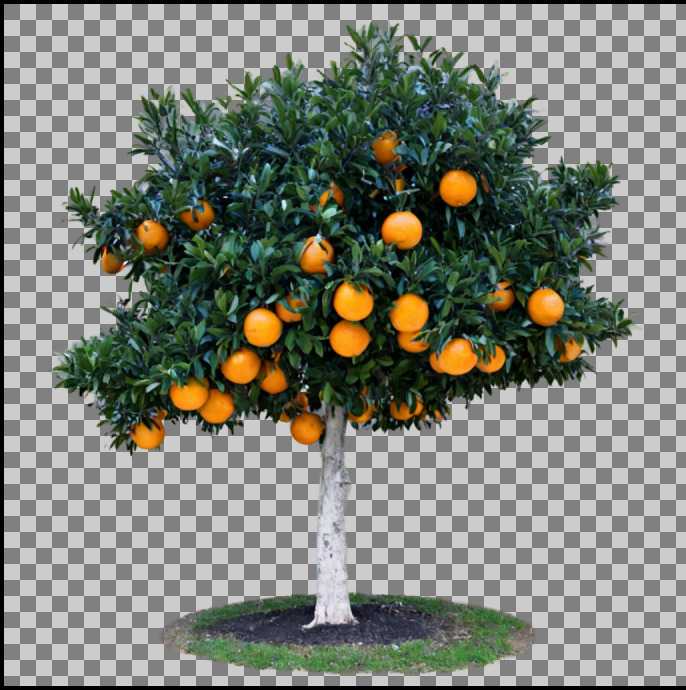}} &
    \parbox[c]{\LayerDiffIssueBoxWidth}{\includegraphics[width=\LayerDiffIssueImageWidth]{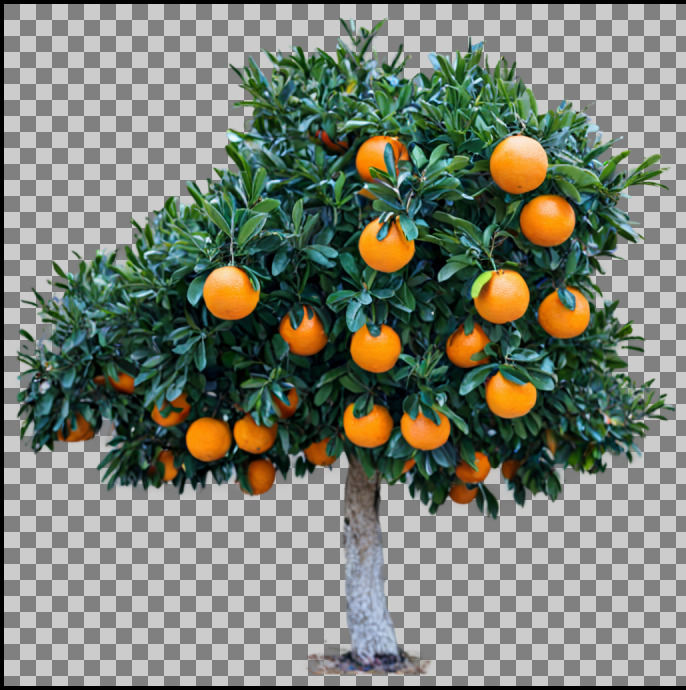}} &
    \parbox[c]{\LayerDiffIssueBoxWidth}{\includegraphics[width=\LayerDiffIssueImageWidth]{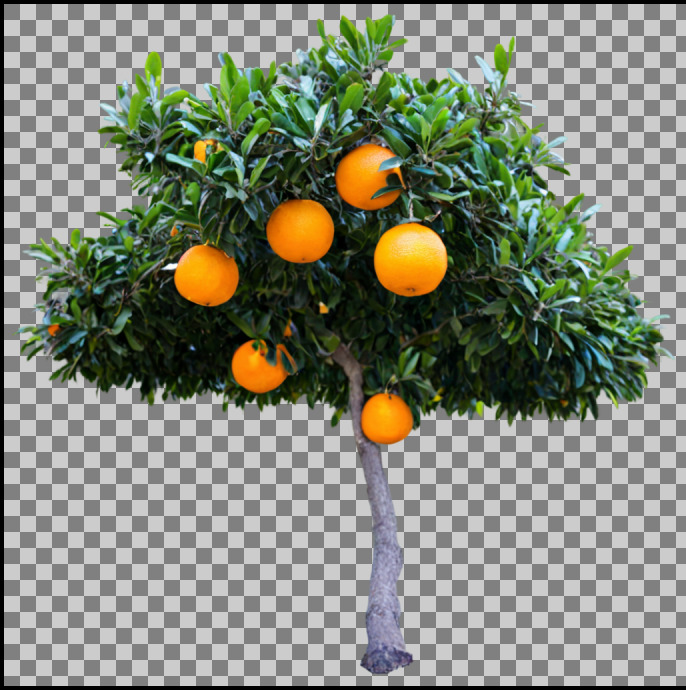}} \\
    \end{tabular}
    }
    \caption{\textbf{The samples generated by Layer Diffusion~\cite{zhang2024layerdiffusion} with \emph{standard} prompt.} ``\textit{An image of a single orange\_(fruit), orange (FRUIT of an orange tree).}''. These images, generated using standard prompts, suffer from low diversity and often contain multiple unintended objects due to their monotonous and ambiguous descriptions.}
    \label{fig:std_prompt_issue}
\end{figure*}

\begin{figure*}
    \centering
    \resizebox{\textwidth}{!}{
    \small
    \begin{tabular}{cccccc}
    \parbox[c]{\LayerDiffIssueBoxWidth}{\includegraphics[width=\LayerDiffIssueImageWidth]{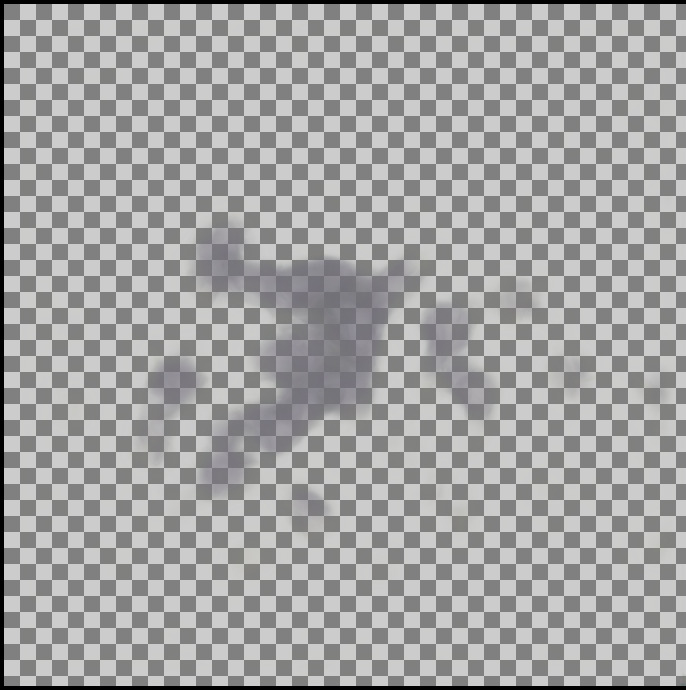}} & 
    \parbox[c]{\LayerDiffIssueBoxWidth}{\includegraphics[width=\LayerDiffIssueImageWidth]{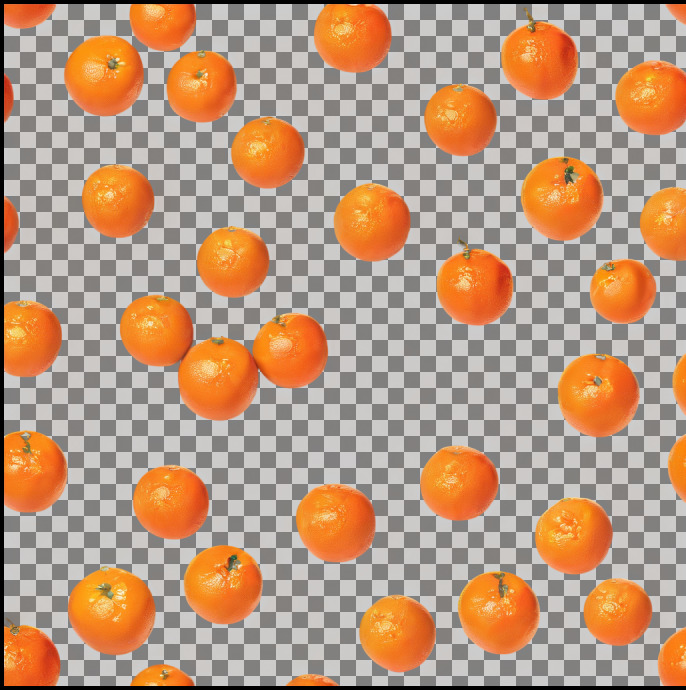}} &
    \parbox[c]{\LayerDiffIssueBoxWidth}{\includegraphics[width=\LayerDiffIssueImageWidth]{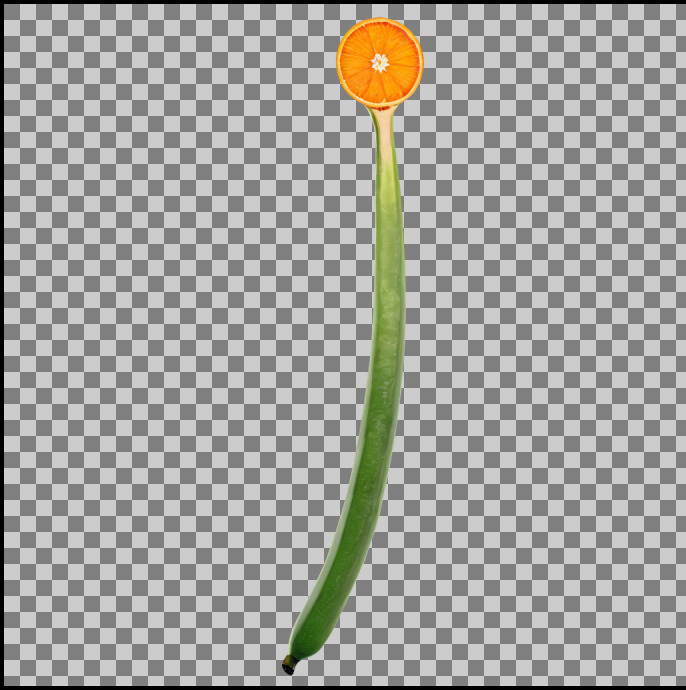}} &
    \parbox[c]{\LayerDiffIssueBoxWidth}{\includegraphics[width=\LayerDiffIssueImageWidth]{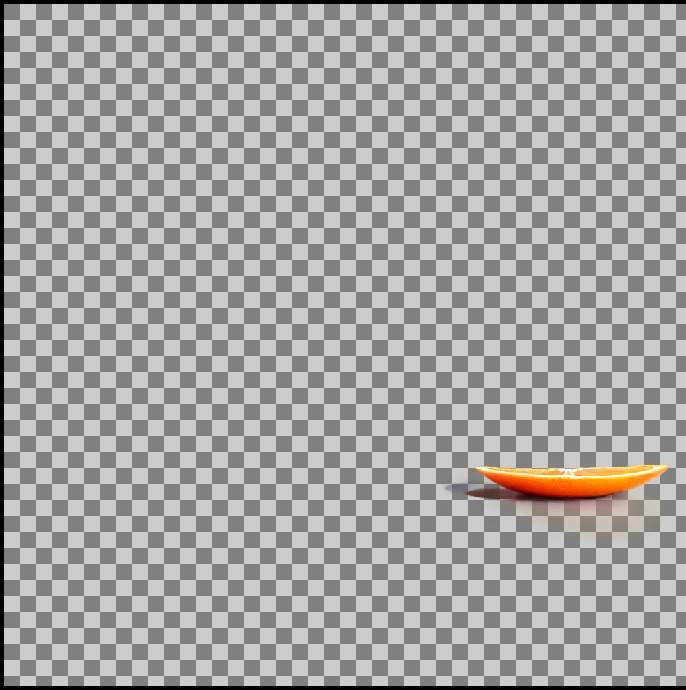}} &
    \parbox[c]{\LayerDiffIssueBoxWidth}{\includegraphics[width=\LayerDiffIssueImageWidth]{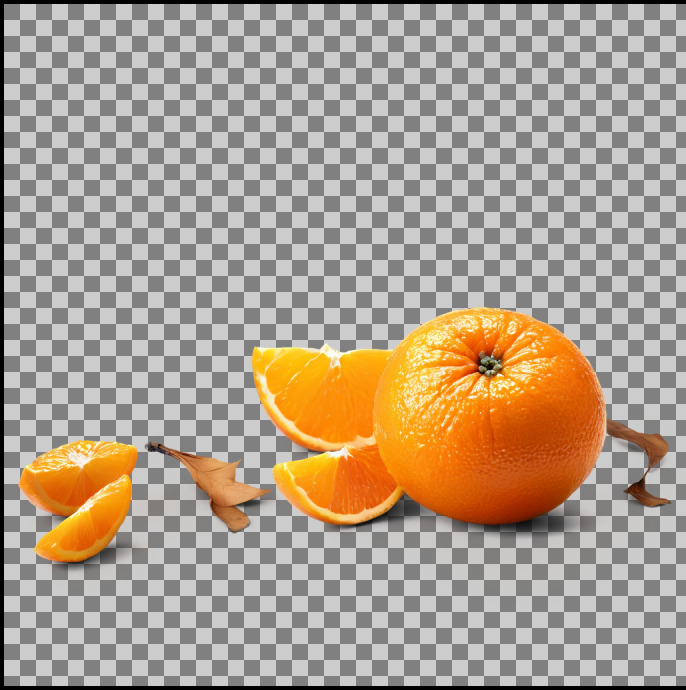}} &
    \parbox[c]{\LayerDiffIssueBoxWidth}{\includegraphics[width=\LayerDiffIssueImageWidth]{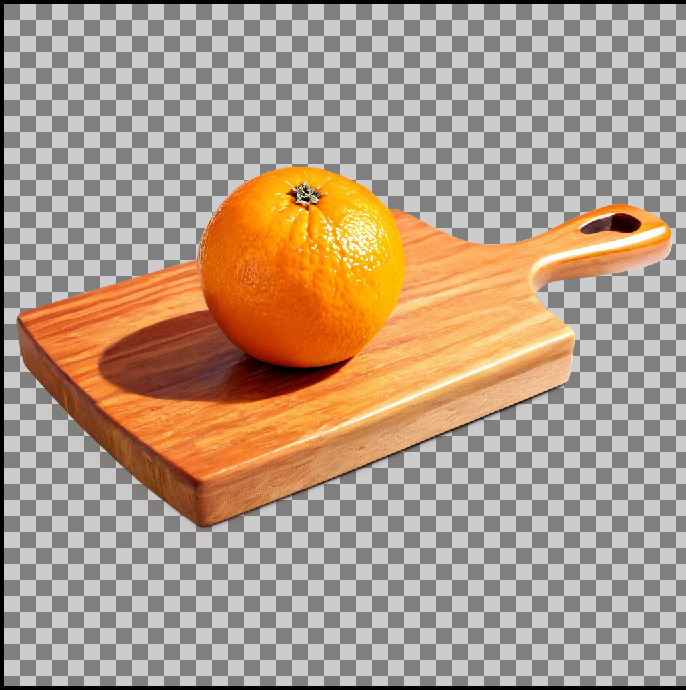}} \\
    w/o Object & Multiple Objects & Wrong Category & Incomplete & Multiple Conditions & Multiple Categories \\
    \end{tabular}
    }
    \caption{\textbf{Failure samples generated by Layer Diffusion~\cite{zhang2024layerdiffusion} with \emph{optimized} prompt.}}
    \label{fig:opt_prompt_issue}
\end{figure*}

\section{Introduction}
\label{sec:introduction}

High-quality data plays a crucial role in computer vision tasks, such as instance segmentation and object detection. However, these tasks suffer from insufficient data. This issue causes poor performance and class imbalance, resulting in low Average Precision (AP) in rare classes. The root of insufficient data is the cost of constructing the dataset. Building large-scale datasets is extremely time-consuming and labor-intensive, and the quality of labels could be unreliable due to human errors, such as missing labels, wrong labels, and inaccurate masks or bounding boxes.

The naive way to synthesize data could be copy-paste~\cite{dvornik2018copypaste}, which means copying foreground objects from images and pasting them onto random backgrounds, or generative models, such as generative adversarial networks (GAN)~\cite{zhang2021datasetgan} or diffusion models~\cite{baranchuk2022labelefficient}. Because of recent advances in text-to-image diffusion models~\cite{rombach2022sd}, X-Paste~\cite{zhao2023xpaste} and MosaicFusion~\cite{xie2024mosaicfusion} use diffusion models to generate images.
However, compared to image generation, obtaining labeled masks or bounding boxes for synthetic data is more challenging.
X-Paste uses an off-the-shelf segmentation model to get the instance mask of an object.
MosaicFusion generates images and masks simultaneously by using a cross-attention map. 
While MosaicFusion can produce a broad vocabulary of instances, several challenges remain in creating high-quality synthetic data for instance segmentation, causing MosaicFusion to discard $\approx50\%$ of the generated images and masks. Moreover, as Figure~\ref{fig:mosaic_issue} shows, another $\approx50\%$ of the filtered generated images and masks still suffer from several issues. First, it may generate multiple objects within a single mask. For example, it generates two snowmen and marks them with a single mask (red region). Second, it could produce masks that do not accurately capture individual instances. For example, not all objects are properly marked. Third, it could occasionally generate an incorrect category. For example, the yellow-masked area should be ``apple \textit{sauce}'', but it generates ``apples''. 

To tackle these challenges, in this work, we introduce Gen-n-Val, a novel agentic data generation and validation framework. Gen-n-Val leverages Layer Diffusion~(LD)~\cite{zhang2024layerdiffusion} and large language model (LLM)/vision large-language models (VLLM) agents~\cite{dubey2024llama} to generate high-quality synthetic data.
We leverage LD to generate foreground images and masks without using the off-the-shelf segmentation model or cross-attention. However, we observe that $\approx44\%$ of the synthetic data generated by LD with the standard prompt (\eg ``\texttt{a photo of a single <object>, <description>}'') is invalid, as shown in Figure~\ref{fig:std_prompt_issue}. 
These images and masks generated by standard prompts not only have low diversity but also have multiple objects due to monotonous and ambiguous descriptions.
Recently, LLMs and VLLMs as agents have gained significant attention. LLMs/VLLMs provide a natural language interface for interacting with other components.
Therefore, we leverage an LLM as the LD prompt agent and employ TextGrad~\cite{yuksekgonul2024textgrad} to refine the system prompts of the LLM, enabling the LLM agent to produce optimized prompts that guide LD to generate high-quality and diverse single-object image instances along with their corresponding segmentation masks.

However, the data generated by these optimized prompts still contains approximately $7\%$ invalid samples (\eg samples without the target object, with multiple objects, or with incorrect categories), as shown in Figure~\ref{fig:opt_prompt_issue}.
To filter out these failed cases, we use the VLLM as the data validation agent. The data validation agent's system prompt is also optimized by TextGrad. 
Finally, we composite validated foreground instances onto background images to produce diverse, high-quality synthetic data for downstream tasks.

As demonstrated in Figure~\ref{fig:teaser}, Gen-n-Val allows us to generate high-quality data for instance segmentation and object detection models, effectively augmenting rare classes, systematically balancing the category distribution, and consistently improving overall model performance.

Our main contributions are summarized as follows.

\begin{itemize}

    \item We propose Gen-n-Val, an agentic framework for generating diverse high-quality data for instance segmentation and object detection with two agents, the LD prompt agent and the data validation agent. 

    \item We use TextGrad to refine the system prompts of the LD prompt agent to generate a high-quality image generation prompt for LD, allowing LD to generate a single foreground instance with a precise mask. We also use a VLLM as the data validation agent to validate generated data, improving the quality of synthetic data. 

    \item The dataset synthesized by Gen-n-Val significantly outperforms previous data synthesis methods in improving model performance. Moreover, Gen-n-Val also has scalability in model capacity and dataset size.

\end{itemize}

%% file: sec/2_RelatedWork.tex
\section{Related Work}
\label{sec:relatedwork}

\subsection{Copy-Paste Augmentation}
To automatically identify realistic locations for object pasting, Dvornik~\etal~\cite{dvornik2018copypaste} introduce a context model that enables objects to be seamlessly pasted into new scenes, significantly improving object detection accuracy. In contrast, Fang~\etal~\cite{fang2019instaboost} jitter objects that already exist within the image rather than directly pasting them from other images. Additionally, Dwibedi~\etal~\cite{dwibedi2017cutpastelearn} propose a method that cuts and blends instances onto random backgrounds to avoid artifacts that can arise from direct pasting. In Simple-Copy-Paste~\cite{ghiasi2021simplecopypaste}, Ghiasi~\etal do not model the surrounding visual context before placing copied instances, but demonstrate that random object placement leads to notable improvements compared to previous approaches. 
While these methods can enhance performance, they may fall short in providing the diverse and high-quality masks and images.

\subsection{Generative-Based Augmentation}
\label{ssec:gen-aug}
Leveraging generative adversarial networks (GANs), \cite{zhang2021datasetgan} utilize a small set of labeled data to train a simple MLP classifier for classifying pixel-wise feature vectors produced by StyleGAN~\cite{karras2019stylegan}. This classifier then serves as a label-generating branch within the StyleGAN architecture. Consequently, data can be generated by sampling latent codes and passing them through the StyleGAN. Following DatasetGAN, \cite{li2022bigdatasetgan} extend BigGAN~\cite{brock2018biggan} and VQGAN~\cite{esser2021vqgan} with a segmentation branch, scaling DatasetGAN to the ImageNet~\cite{russakovsky2015imagenet} level. Since DatasetGAN is trained on synthetic images, it can only sample objects with limited diversity and a less natural appearance. With the aid of powerful diffusion models, Baranchuk~\etal~\cite{baranchuk2022labelefficient} propose a method trained on labeled real images, exploring intermediate activations in pre-trained diffusion models. 
These activations effectively capture semantic information from input images, making them useful for segmentation tasks. Although the quality of synthetic data is promising, previous methods can generate only a limited range of object categories. 

Recent advances in text-to-image models~\cite{rombach2022sd, podell2024sdxl, ramesh2022dalle2} can generate diverse objects in images with natural language prompts.
Therefore, Zhao~\etal\cite{zhao2023xpaste} introduce X-Paste, which synthesizes images with Stable Diffusion and segments these images with an off-the-shelf segmentation model. However, generating data with X-Paste is time-consuming, requiring 4.3 times more GPU hours than MosaicFusion~\cite{xie2024mosaicfusion}.
Gen2Det~\cite{suri2023gen2det} uses an off-the-shelf box-label-conditioned inpainting diffusion model and bounding-box masks to generate data for object detection. 
Xie~\etal~\cite{xie2024mosaicfusion} introduce a training-free, diffusion-based data augmentation pipeline capable of simultaneously producing image and mask pairs by leveraging off-the-shelf text-to-image diffusion models~\cite{rombach2022sd, podell2024sdxl}. 

\begin{figure*}[t]
    \centering

    \begin{subfigure}{\textwidth}
        \centering
        \includegraphics[width=\linewidth]{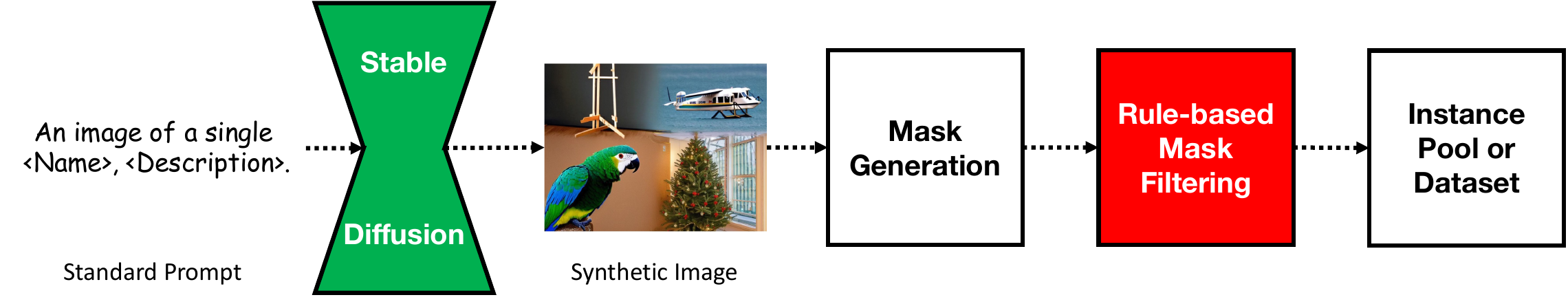}
        \caption{\textbf{Previous generative-based augmentation methods~\cite{zhao2023xpaste, xie2024mosaicfusion}.}}
        \label{fig:pipelineprevious}
    \end{subfigure}

    \vspace{2mm}  % 調整上下間距

    \begin{subfigure}{\textwidth}
        \centering
        \includegraphics[width=\linewidth]{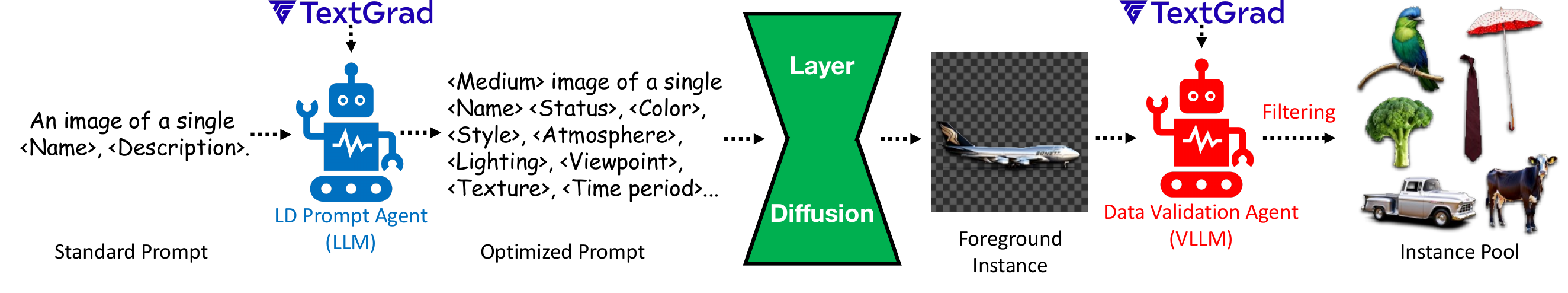}
        \caption{\textbf{Gen-n-Val (Ours).}}
        \label{fig:pipelineours}
    \end{subfigure}
    \vspace{-5mm}
    \caption{\textbf{Comparison of previous augmentation pipelines~\cite{zhao2023xpaste, xie2024mosaicfusion} and our Gen-n-Val pipeline.} (a) {Pipeline of previous generative-based augmentation methods~\cite{zhao2023xpaste, xie2024mosaicfusion}.} The process involves four steps: First, applying Stable Diffusion to generate images with a standard prompt. Next, using augmentation models~\cite{zhao2023xpaste} or the relationship between visual and textual embeddings in cross-attention~\cite{xie2024mosaicfusion} to obtain segmentation masks. Finally, filtering out flawed masks with hand-crafted rules. (b) {Pipeline of Gen-n-Val.} The process begins with optimizing system prompts of the Layer Diffusion~\cite{zhang2024layerdiffusion} prompt agent (LLM) using TextGrad. The optimized prompts generate Layer Diffusion~\cite{zhang2024layerdiffusion} prompts, and Layer Diffusion~\cite{zhang2024layerdiffusion} produces transparent instances and background images. A VLLM-based validation agent filters flawed images, and multiple valid instances are pasted into other images. See Section~\ref{sec:method} for more details.}
    \label{fig:pipeline}
\end{figure*}

As shown in Figure~\ref{fig:pipelineprevious}, the pipeline of these methods involves four steps: 
(1) Generate images using Stable Diffusion with prompts filled by category name; (2) Obtain segmentation masks using either augmentation models or cross-attention maps; (3) Refine masks through edge detection and filter out flawed masks; (4) Composite the instance and background to create final images.
However, these methods have several issues. First, the generated images are not guaranteed to contain only one instance. Second, the segmentation masks generated by cross-attention are of poor quality. Lastly, the generated images are filtered by hand-crafted rules, such as thresholding, leading to a large number of unqualified images, such as missing corresponding masks and incorrect object labels. In contrast, our method can generate data with precise masks, correct category, and the correct number of object instances, enhancing the performance of downstream tasks.

\subsection{Generative Models and Agents}
Layer Diffusion~\cite{zhang2024layerdiffusion} is an approach for generating transparent images. This method encodes the transparent alpha channel into the latent distribution of Stable Diffusion~\cite{rombach2022sd, podell2024sdxl} to generate transparent images. 

OpenAI released the Large Language Model~(LLM), GPT-4~\cite{achiam2023gpt}, and the Vision Large Language Model~(VLLM), GPT-4V~\cite{openai2023gpt4v}. 
Inspired by GPT-4V, Liu~\etal~\cite{liu2023llava} introduce Large Language-and-Vision Assistant~(LLaVA), an open-source VLLM based on the CLIP image encoder~\cite{radford2021clip} and the open-source LLM Vicuna~\cite{vicuna2023}.
Besides, Meta released Llama~\cite{touvron2023llama} and Llama 3~\cite{dubey2024llama}, powerful language models that perform exceptionally well across a wide range of language understanding tasks, comparable to leading models such as GPT-4. 
Moreover, Meta released Llama 3.2~\cite{meta2024llama32}, integrating vision capability into Llama and allowing it to process visual inputs. 
To enhance the feedback provided by LLM, Yuksekgonul~\etal~\cite{yuksekgonul2024textgrad} propose TextGrad, a method to refine system prompts by backpropagating textual information from outputs, encouraging precise answers. 

With the development of LLMs, using them to decide and solve various tasks has become a growing trend. A common approach is to use LLMs to generate textual ``actions" or ``decisions" for tasks.
VISPROG~\cite{gupta2023visual} uses GPT3~\cite{brown2020gpt3} to automatically generate programs that solve complex visual tasks.
ToolFormer~\cite{schick2023toolformer} also uses GPT3 to decide which APIs to call for solving tasks.
HuggingGPT~\cite{shen2023hugginggpt} uses ChatGPT to choose models on HuggingFace for solving complicated AI tasks. 
ReAct~\cite{yao2023react} combines reasoning and action steps, allowing LLM to make better decisions and provide more reliable answers.
Reflexion~\cite{shinn2023reflexion} enhances LLM agents’ decision-making by using reflective linguistic feedback stored in an episodic memory.
Zheng~\etal~\cite{zheng2024gpt4vagent} use GPT-4V as a web agent.
Generative Agents~\cite{Park2023GenerativeAgents} introduce a system of generative agents that simulate realistic human behaviors through memory, planning, and reflection, enabling interactive digital environments.
In summary, these methods demonstrate the growing power of LLMs to integrate decision-making and execution, allowing more robust and adaptable AI solutions across diverse domains.

%% file: sec/3_Method.tex
\section{Method}
\label{sec:method}

To tackle the issues we discuss in Section~\ref{ssec:gen-aug}, we leverage a Large Language Model (LLM) as the LD prompt agent, a Vision Large Language Model (VLLM) as the data validation agent, and Layer Diffusion (LD) to generate diverse and high-quality data, as shown in Figure~\ref{fig:pipelineours}.
First, in the \textbf{Open Vocabulary Prompt Generation} stage, we initiate the process by employing TextGrad~\cite{yuksekgonul2024textgrad} optimization techniques to refine the quality and effectiveness of the LD prompt agent's system prompt, allowing the LD prompt agent to generate diverse and high-quality prompts for LD. 
The details are shown in Section~\ref{ssec:OpenVocabularyPromptGeneration}.
Second, during the \textbf{Foreground Image Generation} stage, we utilize the LD to generate transparent instance images.
This step creates isolated instance images, allowing us to generate segmentation masks without manual annotation or additional segmentation algorithms. 
The details are shown in Section~\ref{ssec:ForegroundImageGeneration}.
Third, to ensure the quality of the generated images, we filter out flawed samples during the \textbf{Image Filtering} stage using the VLLM as the data validation agent, which is further optimized with TextGrad. 
The details are shown in Section~\ref{ssec:imagefiltering}.
Finally, we randomly paste multiple instances into background images. 
This step creates diverse scenes with multiple instances, further augmenting the dataset.

\subsection{Open Vocabulary Prompt Generation}
\label{ssec:OpenVocabularyPromptGeneration}

Previous generative-based augmentation methods
relied on adding the word ``single'' to the prompt during the stage of image generation to instruct the Stable Diffusion to generate a single-object instance. This approach not only struggles to keep the diffusion model focused on a single instance but also accidentally allows other types of objects to appear in the background due to the ambiguous concept of ``single''. The ideal Stable Diffusion prompts should be detailed and specific with several components, including the object class, action, environment setting, and other relevant details~\cite{stewart2024promptguide}. For example, Figure~\ref{fig:pipelineours} demonstrates both the standard and optimized prompts for image generation. 
For more examples, please refer to our appendix.

To generate optimized LD prompts, we need a high-quality system prompt $p_\text{sys}$ for the LD prompt agent $A_{p_\text{LD}}$ (a LLM). 
Therefore, we leverage two other LLMs, a prompt evaluator $E_{\text{prompt}}$ and a prompt validator $V_{\text{prompt}}$, and TextGrad~\cite{yuksekgonul2024textgrad} optimization techniques to optimize the system prompt $p_\text{sys}$. 
The system prompt generation process is shown in Algorithm~\ref{alg:gen_open_vocab_prompt}. 
First, we give the LD prompt agent $A_{p_\text{LD}}$ an initial system prompt $p_\text{sys}$ that instructs it to generate a prompt $p_\text{LD}$ for the LD. 
Second, we evaluate the $p_\text{LD}$ using a prompt evaluator $E_{\text{prompt}}$ to determine the quality of the prompt for LD. The prompt evaluator $E_{\text{prompt}}$ is a second LLM which is asked to evaluate the quality of the prompt generated by the LD prompt agent $A_{p_\text{LD}}$ and provide a criticism in the form of a loss $L$.
Third, with the loss $L$, we optimize the initial system prompt using Text Gradient Descent (TGD)~\cite{yuksekgonul2024textgrad} to generate a new optimized system prompt $p^*_\text{sys}$.
Fourth, we use the optimized system prompt $p^*_\text{sys}$ to generate an optimized prompt $p^*_\text{LD}$ for the LD model. 
Finally, we validate the $p^*_\text{LD}$ to compare the quality with the $p_\text{LD}$. The prompt validator $V_{\text{prompt}}$ is a third LLM that is asked to evaluate the quality of the optimized prompt and provide a decision $d$ of whether replacing the initial prompt with the optimized prompt at the next iteration is beneficial.

\begin{algorithm}[t]
\caption{Optimize System Prompt $p_{\text{sys}}$ of $A_{p_{\text{LD}}}$}
\label{alg:gen_open_vocab_prompt}
\begin{algorithmic}[1]
\State \textbf{Input:} initial system prompt $p_{\text{sys}}$, max iteration $I$
\State \textbf{Initialize:} decision $d \gets \text{False}$, iteration $i \gets 0$
\While{not $d$ and $i < I$}
    \State $p_{\text{LD}} \gets A_{p_{\text{LD}}}(p_{\text{sys}})$ \Comment{Generate LD prompt}
    \State $L \gets E_{\text{prompt}}(p_{\text{LD}})$ \Comment{Evaluate prompt quality}
    \State $p_{\text{sys}}^{*} \gets \mathrm{TGD}(L)$ \Comment{Update system prompt}
    \State $p_{\text{LD}}^{*} \gets A_{p_{\text{LD}}}(p_{\text{sys}}^{*})$ \Comment{Generate refined LD prompt}
    \State $d \gets V_{\text{prompt}}(p_{\text{LD}}^{*})$ \Comment{Check if prompt is satisfactory}
    \If{$d$}
        \State \textbf{break} \Comment{Stop if validator accepts}
    \Else
        \State $p_{\text{sys}} \gets p_{\text{sys}}^{*}$ \Comment{Adopt refined system prompt}
    \EndIf
    \State $i \gets i + 1$
\EndWhile
\end{algorithmic}
\end{algorithm}

%%%%%%%%%%%%%%%%%%%%%%%%%%%%%%%%%%%%%%%%%%%%%%%%%%%%%%%%%%%%%%%%%%%%%%%%%%%%%%%%%%%%%%%%%%%%%%%%%%%%%%%%%%%%%%%%%%%%%%%%%%%%%%%%%%%%%%%%%%%%%%%%%%%%%%%%%%%%%%%%%%%%%%%%%%%%%%%%

\subsection{Foreground Image Generation}
\label{ssec:ForegroundImageGeneration}

We propose a method that utilizes the LD to generate transparent images of foreground instances along with precise alpha masks. By employing the optimized prompts from the Open Vocabulary Prompt Generation stage, we empower the LD model to produce high-quality, single-object instances that strictly adhere to the specified prompts.
The optimized prompts are input into the LD model to generate transparent images where each pixel contains both RGB values and an alpha transparency channel. This transparency channel inherently provides an accurate segmentation mask for the foreground object. Unlike previous methods, our approach eliminates the need for additional segmentation algorithms, such as SAM~\cite{kirillov2023segany}, or manual annotation, as the alpha channel directly corresponds to the object's silhouette, resulting in precise and clean masks aligned perfectly with the generated images.

The detailed and specific prompts generated by the LD prompt agent $A_{p_\text{LD}}$ include various attributes, such as object class, style, color, texture, lighting, atmosphere, viewpoint, and time period. By incorporating these attributes, we guide the LD to produce a wide range of object appearances, increasing the diversity of the generated instances. Prompt conditioning and negative prompts are also used to guide the model toward generating images that closely match the desired characteristics while avoiding undesired elements.

While the LD generates high-quality transparent instances, minor background noise may still be present in the alpha channel due to imperfections in the diffusion process. To mitigate this, we apply a median filter to the alpha channel of the images. The median filter effectively removes isolated pixels and smooths the mask edges without significantly altering the overall shape of the object. This post-processing step ensures that the segmentation masks are clean and accurately represent the foreground instances.

%%%%%%%%%%%%%%%%%%%%%%%%%%%%%%%%%%%%%%%%%%%%%%%%%%%%%%%%%%%%%%%%%%%%%%%%%%%%%%%%%%%%%%%%

%%%%%%%%%%%%%%%%%%%%%%%%%%%%%%%%%%%%%%%%%%%%%%%%%%%%%%%%%%%%%%%%%%%%%%%%%%%%%%%%%%%%%%%%%%%%%%%%%%%%%%%%%%%%%%%%%%%%%%%%%%%%%%%%%%%%%%%%%%%%%%%%%%%%%%%%%%%%%%%%%%%%%%%%%%%%%%%%
\subsection{Image Filtering}
\label{ssec:imagefiltering}

Despite the improvements achieved through optimized prompts and the use of the LD, some generated images may still contain flaws, such as missing the target object, featuring multiple instances when only one is desired, incorrect object categories, or poor visual quality, as shown in Figure~\ref{fig:opt_prompt_issue}. To further enhance the quality of our synthetic dataset, we employ the VLLM as a data validation agent to automatically filter out these flawed images.

Similar to our approach in optimizing prompts for the LD, we apply the TextGrad optimization technique to refine the system prompts of the data validation agent. By optimizing the data validation agent's system prompts, we enhance its ability to 
% accurately 
evaluate and identify flaws in the generated images. The optimized prompts guide the data validation agent to focus on specific criteria essential for high-quality instance segmentation data, such as the presence of a single, correctly categorized object, and the cleanliness of the transparent background in the foreground images.

The data validation agent is tasked with analyzing generated images to assess their suitability for inclusion in the dataset. Here, $c$ represents the object category. The key criteria encoded into the data validation agent through the optimized system prompt include:
\begin{itemize}
    \item Single $c$: The image should contain only one $c$.
    \item Single View: The $c$ should be shown from a single angle or perspective.
    \item Intact $c$: The $c$ should be intact and fully visible.
    \item Plain Background: The background should be empty or plain, without distracting elements.
\end{itemize}

%%%%%%%%%%%%%%%%%%%%%%%%%%%%%%%%%%%%%%%%%%%%%%%%%%%%%%%%%%%%%%%%%%%%%%%%%%%%%%%%%%%%%%%%%%%%%%%%%%%%%%%%%%%%%%%%%%%%%%%%%%%%%%%%%%%%%%%%%%%%%%%%%%%%%%%%%%%%%%%%%%%%%%%%%%%%%%%%

%% file: sec/4_Experiments.tex
\section{Experiments}
\label{sec:experiments}
To demonstrate the enhancement of detection models after applying our methods, we evaluate on two challenging benchmarks, COCO~\cite{lin2014microsoft} and LVIS~\cite{gupta2019lvis}. 
COCO dataset is a large-scale dataset designed for object detection, segmentation, and captioning, containing 330K images with 1.5M object instances labeled across 80 categories. The LVIS dataset uses the same images as COCO but provides more detailed and partitioned annotations across 1,203 categories, offering finer classification.
To quantify this improvement, we compute mean Average Precision (mAP) as our evaluation metric. 
We use Meta-LLaMA-3.1-8B-Instruct as the LD prompt agent and Meta-LLaMA-3.2-11B-Vision-Instruct as the data validation agent. 
We apply Copy-Paste~\cite{dvornik2018copypaste}  in all YOLO experiments except for the baseline.
For more details, please refer to our appendix.

\begin{table}
    \centering
    % \rowcolors[]{8}{cvprblue!30}{white}
    \caption{\textbf{LVIS~\cite{gupta2019lvis} instance segmentation and object detection benchmark.} We borrow numbers of Mask R-CNN~\cite{he2017maskrcnn} and MosaicFusion~\cite{xie2024mosaicfusion} from MosaicFusion~\cite{xie2024mosaicfusion}.}
    \label{tab:lvis}
    \vspace{-3mm}
    \resizebox{\columnwidth}{!}{
    \begin{tabular}{l|cccc}
      \toprule % <-- Toprule here
      \textbf{Method} & $\textbf{mAP}^\textbf{box}$ & $\textbf{mAP}^\textbf{box}_\textbf{rare}$ & $\textbf{mAP}^\textbf{mask}$ & $\textbf{mAP}^\textbf{mask}_\textbf{rare}$\\

      \midrule % <-- Midrule here
      Mask R-CNN (baseline) & 22.5 & 9.1 & 21.7 & 9.6 \\% official 52.4 42.2
      MosaicFusion~\cite{xie2024mosaicfusion} & 24.0 & 14.8 & 23.1 & 15.2 \\
      Gen2Det~\cite{suri2023gen2det} & 24.4 & 15.4 &  23.6 &  15.3 \\
      Gen-n-Val (Ours) & \textbf{26.8} &\textbf{16.1} & \textbf{25.6} & \textbf{17.2} \\
      \textit{versus baseline}  & \textcolor{OliveGreen}{+4.3} & \textcolor{OliveGreen}{+7.0} & \textcolor{OliveGreen}{+3.9} & \textcolor{OliveGreen}{+7.6} \\
      
      \midrule
      \midrule
      YOLO11m (baseline) & 12.9 & 8.3 & 10.3 & 6.5 \\
      Copy-Paste~\cite{dvornik2018copypaste} & 12.9 & 8.2 & 10.4 & 6.7 \\
      Gen-n-Val (Ours) & \textbf{17.2} & \textbf{13.0} & \textbf{14.5} & \textbf{10.1} \\
      \textit{versus baseline}  & \textcolor{OliveGreen}{+4.3} & \textcolor{OliveGreen}{+4.7} & \textcolor{OliveGreen}{+4.2} & \textcolor{OliveGreen}{+3.6}\\

      \midrule
      \midrule
      CenterNet2 (baseline) & 47.5 & 41.4 & 42.3 & 36.8 \\
      XPaste~\cite{zhao2023xpaste} & 50.1 & 48.2 & 44.4 & 43.3 \\
      DiverGen~\cite{fan2024divergen} & 51.2 & 50.1 & 45.5 & 45.8 \\
      Gen-n-Val (Ours) & \textbf{51.5} & \textbf{55.6} &\textbf{45.9} & \textbf{48.3} \\
      \textit{versus baseline}  & \textcolor{OliveGreen}{+4.0} & \textcolor{OliveGreen}{+14.2} & \textcolor{OliveGreen}{+3.6} & \textcolor{OliveGreen}{+11.5} \\
      
      \bottomrule % <-- Bottomrule here
    \end{tabular}
    }
    
\end{table}

\begin{table}
    \centering
    \caption{\textbf{Effect of flattening the LVIS~\cite{gupta2019lvis} class distribution.}
By augmenting images with additional instances, we enforce a minimum of 10–1000 images per class, which substantially boosts performance on rare categories and improves overall accuracy. \textbf{Min. img./cls.}: the minimum number of images per class. \textbf{Add. inst.}: the number of added instances. }
    \label{tab:lvis_flat}
    \vspace{-3mm}
    \resizebox{\columnwidth}{!}{
    \begin{tabular}{l|c|cccc}
      \toprule % <-- Toprule here
      \textbf{Min. img./cls.} & \textbf{Add. inst.} & $\textbf{mAP}^\textbf{box}$ & $\textbf{mAP}^\textbf{box}_\textbf{rare}$ & $\textbf{mAP}^\textbf{mask}$ & $\textbf{mAP}^\textbf{mask}_\textbf{rare}$\\

      \midrule % <-- Midrule here
      Original & 0 & 22.5 & 9.1 & 21.7 & 9.6 \\% official 52.4 42.2
      10 & 1,874 & 23.1 & 9.8 & 22.5 & 10.5 \\
      100 & 60,306 & 24.2 & 13.5 & 23.4 & 14.7 \\
      1000 & 727,393& \textbf{26.8} & \textbf{16.1} & \textbf{25.6} & \textbf{17.2} \\
      
      \bottomrule % <-- Bottomrule here
    \end{tabular}
    }
    
\end{table}

%%%%%%%%%%%%%%%%%%%%%%%%%%%%%%%%%%%%%%%%%%%%%%%%%%%%%%%%%%%%%%%%%%%%%%%%%%%%%%%%%%%%%%%%
\subsection{LVIS Benchmark}

We train Mask R-CNN~\cite{he2017maskrcnn}, YOLO11m~\cite{khanam2024yolov11}, and CenterNet2~\cite{zhou2021probablistic} on the LVIS training set. As shown in the right part of Figure~\ref{fig:teaser}, the LVIS training set exhibits a highly long-tailed category distribution, where most categories appear in only a few images. To balance this distribution, we generate 727,393 instances and paste them onto LVIS images until all 1,203 categories reach at least 1,000 images. For rare category analysis, we adopt the category frequency annotations provided by LVIS. We evaluate on the LVIS validation set. The results are shown in Table~\ref{tab:lvis}. Compared to the baseline, our method improves the mAP by 4.3\% for box prediction and 3.9\% for mask prediction on Mask R-CNN, by 4.0\% for box prediction and 3.6\% for mask prediction on CenterNet2, and by 4.3\% for box prediction and 4.2\% for mask prediction on YOLO11m. 
Rare-category mAP is also significantly improved over the baselines.

\subsection{Distribution Balancing of LVIS}

To investigate Gen-n-Val's ability to balance the long-tailed distribution of LVIS, as shown in Figure~\ref{fig:teaser}, we progressively ``fill'' the tail of the distribution by generating additional synthetic instances for under-represented classes. Concretely, starting from the original LVIS distribution, we construct three variants of the training set by enforcing different lower bounds on the number of images per class: (1) a mild setting where we up-sample rare categories so that every class appears in at least 10 images, (2) a stronger setting that further densifies the common categories until each class has at least 100 images, and (3) an aggressive setting where we continue to generate instances until all 1,203 categories reach at least 1,000 images. These three stages correspond to injecting 1.8K, 60.3K, and 727K additional instances, respectively, and balance the long-tailed distribution as shown in Figure~\ref{fig:teaser}. As demonstrated in Table~\ref{tab:lvis_flat}, Gen-n-Val leads to consistent performance gains for Mask R-CNN: mask mAP improves from 21.7 to 22.5, 23.4, and 25.6, while rare-category mask mAP increases from 9.6 to 10.5, 14.7, and 17.2. Overall, enforcing 10$\sim$1,000 images per class translates into +0.8$\sim$+3.9 mAP on all categories and +0.9$\sim$+7.6 mAP on rare categories, demonstrating that Gen-n-Val effectively balances LVIS and benefits tail classes.

\begin{table}
    \centering
    \small
    % \rowcolors[]{8}{cvprblue!30}{white}
    \caption{\textbf{COCO~\cite{lin2014microsoft} instance segmentation and object detection benchmark with YOLO9c~\cite{wang2024yolov9} and YOLO11m~\cite{khanam2024yolov11}.}}
    \label{tab:coco}
    \vspace{-3mm}
    \resizebox{\columnwidth}{!}{
    \begin{tabular}{l|cccc}
      \toprule % <-- Toprule here
      \textbf{Method} & $\textbf{mAP}^\textbf{box}$ & $\textbf{mAP}^\textbf{box}_\textbf{rare}$ & $\textbf{mAP}^\textbf{mask}$ & $\textbf{mAP}^\textbf{mask}_\textbf{rare}$\\

      \midrule % <-- Midrule here
      YOLOv9c (baseline)  & 50.1 & 52.8 & 41.3 & 45.1\\
      Copy-Paste & 50.9 & 53.1 & 42.1 & 46.8\\
      MosaicFusion & 51.4 & 53.6 & 42.7 & 47.9\\
      Gen-n-Val & \textbf{51.9} & \textbf{54.4} & \textbf{43.4} & \textbf{48.7}\\
      \textit{versus baseline}  & \textcolor{OliveGreen}{+1.8} & \textcolor{OliveGreen}{+1.6} & \textcolor{OliveGreen}{+2.1} & \textcolor{OliveGreen}{+3.6}\\

      \midrule % <-- Midrule here
      \midrule
      YOLO11m (baseline) & 49.6 & 51.9 & 39.8 & 45.4\\
      Copy-Paste & 50.0 & 52.1 & 41.5 & 47.1\\
      MosaicFusion & 50.6 & 54.3 & 42.0 & 48.1\\
      Gen-n-Val & \textbf{51.7} & \textbf{55.4} & \textbf{42.9} & \textbf{49.0}\\

      \textit{versus baseline}  & \textcolor{OliveGreen}{+2.1} & \textcolor{OliveGreen}{+3.5} & \textcolor{OliveGreen}{+3.1} & \textcolor{OliveGreen}{+3.6}\\

      \bottomrule % <-- Bottomrule here
    \end{tabular}
    }
    \vspace{-1em}
    
\end{table}

\subsection{COCO Benchmark}
We train YOLOv9c~\cite{wang2024yolov9} and YOLO11m~\cite{khanam2024yolov11} using the COCO dataset and our 16K synthetic dataset, and evaluate on the validation set. We choose the 10 least frequent categories as rare categories and the remaining 70 categories as common categories. 
We use a synthetic background image and paste instances with image harmonization~\cite{wang2023harmonization}.  
We compare our method with the baseline YOLOv9c, YOLO11m, and Copy-Paste. 
As shown in Table~\ref{tab:coco}, Gen-n-Val outperforms the baseline and other methods, achieving the highest mAP scores for both box and mask predictions. Compared to the baseline, our method improves the mAP by 1.8\% for box prediction and 2.1\% for mask prediction on YOLOv9c, and by 2.1\% for box prediction and 3.1\% for mask prediction on YOLO11m. Furthermore, rare-category mask mAP improves by 3.6\% on YOLOv9c and by 3.6\% on YOLO11m, demonstrating the effectiveness of our method. 
These results demonstrate the effectiveness of our synthetic data generation approach.

\begin{table}
  \begin{center}
  \small
  \caption{\textbf{COCO~\cite{lin2014microsoft} open-vocabulary object detection.}}
    \label{tab:openvocabulary}
    \vspace{-3mm}
    \begin{tabular}{l|cc}
      \toprule % <-- Toprule here
      \textbf{Method} & $\textbf{mAP}^\textbf{box}$ & $\textbf{mAP}^\textbf{box}_\textbf{novel}$\\
      \midrule % <-- Midrule here
      YOLO-Worldv2-M           & 42.7 & 20.6\\
      YOLO11m w/ Gen-n-Val       & \textbf{49.8} & \textbf{25.5}\\
      \bottomrule % <-- Bottomrule here
    \end{tabular}
  \end{center}
\end{table}

\newcommand\QualitativeBoxWidth{.33\textwidth}
\newcommand\QualitativeImageWidth{.33\textwidth}
\begin{figure*}[t]

  \centering
  \begin{tabular}{@{}c@{}c@{}c@{}}

    \parbox[c]{\QualitativeBoxWidth}{
    \includegraphics[width=\QualitativeImageWidth]{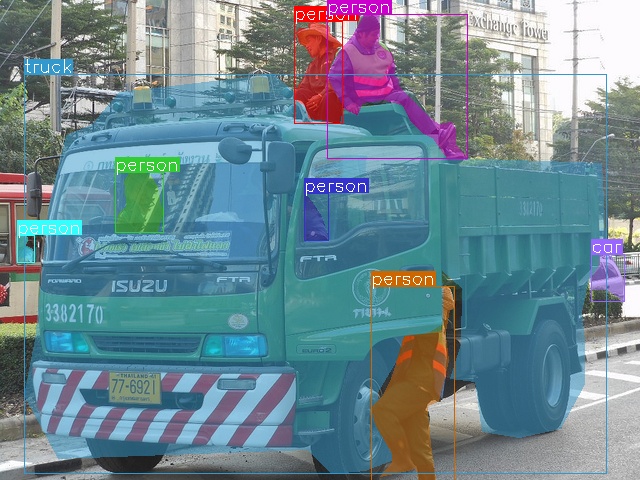} 
    } & 
    \parbox[c]{\QualitativeBoxWidth}{
    \includegraphics[width=\QualitativeImageWidth]{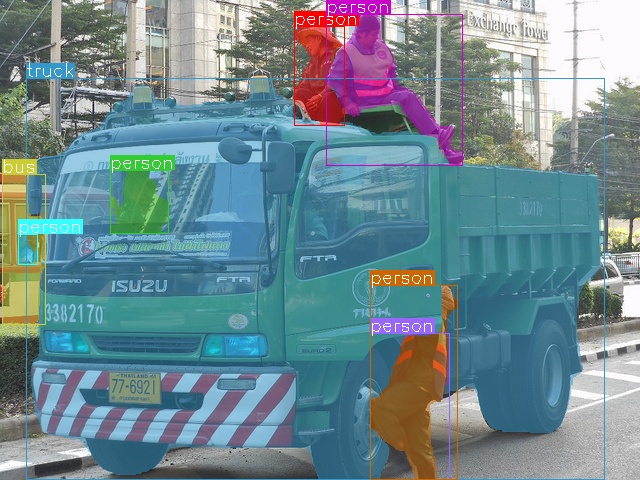} 
    } & 
    \parbox[c]{\QualitativeBoxWidth}{
    \includegraphics[width=\QualitativeImageWidth]{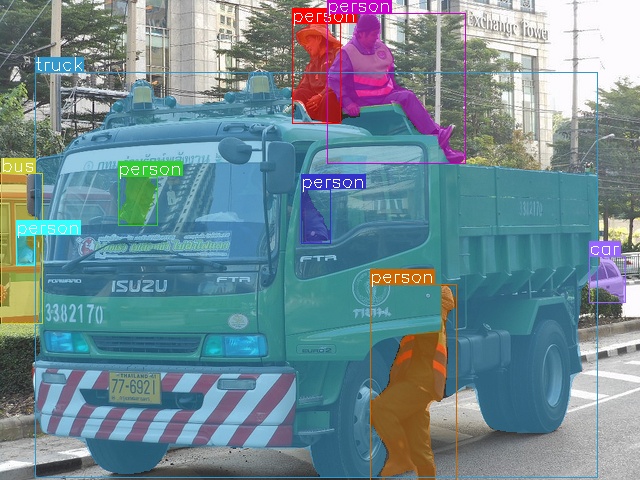} 
    } \\

    Ground Truth & Baseline & Gen-n-Val \\
  \end{tabular}
  \caption{\textbf{Qualitative comparison of ground truth, baseline model, and model trained on Gen-n-Val.}}
  \label{fig:qualitative}
\end{figure*}

\subsection{Open-Vocabulary Object Detection}
\label{sec:openvocabulary}

We train YOLO11m using the COCO dataset and our 16K synthetic dataset, which includes 80 categories from the COCO dataset and 10 additional categories from the LVIS dataset, and evaluate on the 5K-image COCO/LVIS validation set for open-vocabulary object detection. Ten LVIS categories are selected as novel categories in the validation set, and the whole COCO dataset is used as the training set. We compare our method with the baseline YOLO-Worldv2-M~\cite{cheng2024yoloworld}, which is an advanced real-time object detection model that enhances the YOLO framework with open-vocabulary capabilities. As shown in Table~\ref{tab:openvocabulary}, compared to the baseline, Gen-n-Val improves mAP by 7.1\% for box prediction and by 4.9\% on novel categories with YOLO11m, demonstrating the effectiveness of our method on open-vocabulary object detection.

\begin{table}
  \begin{center}
  \small
  \caption{
  \textbf{Ablation results on YOLO11m~\cite{khanam2024yolov11} trained with synthetic data on COCO~\cite{lin2014microsoft}.}
  We evaluate the effects of Layer Diffusion~\cite{zhang2024layerdiffusion} prompt optimization ($p^*_\text{LD}$), VLLM filtering (VLLM), median filtering (Med.), and Copy-Paste~\cite{dvornik2018copypaste} (C-P).}
    \label{tab:ablation}
    \vspace{-3mm}

    \begin{tabular}{cccc|cc}
      \toprule % <-- Toprule here
       $p^*_\text{LD}$ & VLLM & Med. & C-P & $\textbf{mAP}^\textbf{box}$ & $\textbf{mAP}^\textbf{mask}$\\
      \midrule % <-- Midrule here
        & \checkmark & \checkmark & \checkmark & 48.8 & 39.2\\
       \checkmark &   & \checkmark & \checkmark & 51.5 & 42.0\\
       \checkmark & \checkmark & & \checkmark& 51.5 & 42.2\\
        \checkmark  & \checkmark & \checkmark &  & 51.2 & 42.7 \\
       
        \checkmark  & \checkmark & \checkmark & \checkmark & \textbf{51.7} & \textbf{42.9}\\
      \bottomrule % <-- Bottomrule here
    \end{tabular}

  \end{center}
\end{table}

\subsection{Ablation Studies}

We conduct an ablation study on COCO by comparing YOLO11m with and without prompt optimization, VLLM filtering, median filtering, and Copy-Paste. As shown in Table~\ref{tab:ablation}, prompt optimization provides the largest gain, improving mAP by 2.9\% for object detection and 3.7\% for instance segmentation. On top of that, VLLM filtering improves mAP by 0.2\% and 0.9\%, respectively. Median filtering yields an additional 0.2\% and 0.7\% mAP improvement, while Copy-Paste brings another 0.5\% and 0.2\% mAP gain for object detection and instance segmentation.

\begin{figure}
    \centering
    \includegraphics[width=\columnwidth]{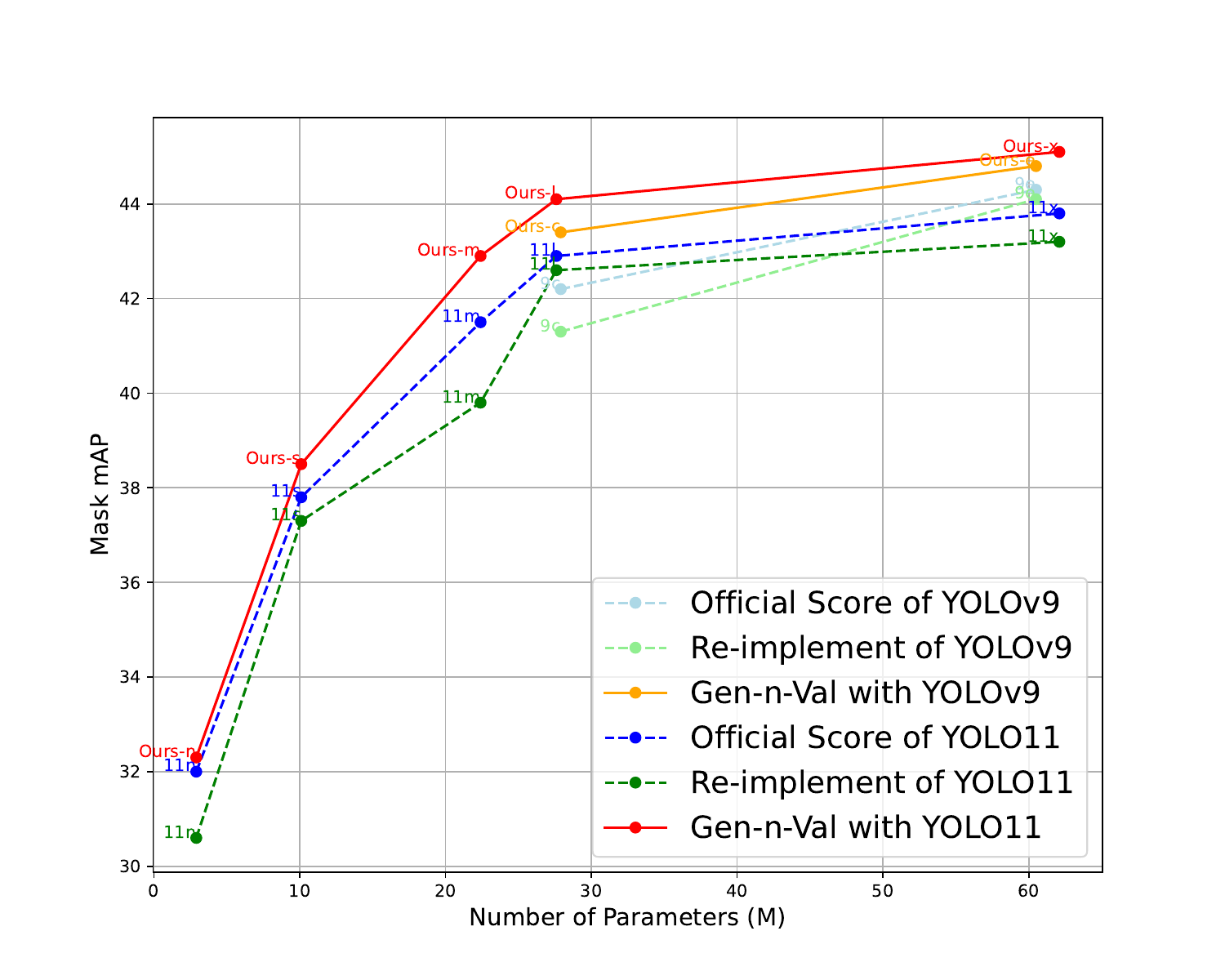}
  \vspace{-6mm}
    \caption{\textbf{The instance segmentation performance of YOLOv9~\cite{wang2024yolov9} and YOLO11~\cite{khanam2024yolov11} family models.} As the size of the model increases, performance improves. Gen-n-Val is effective with all family models. }

    \label{fig:coco_yolo_seg}
\end{figure}

\subsection{Model Scalability}
To comprehensively evaluate our method across different model scales, we evaluate Gen-n-Val on the COCO dataset using the YOLOv9 and YOLO11 family models, which are state-of-the-art in the YOLO series. In the instance segmentation task, as shown in Figure~\ref{fig:coco_yolo_seg}, Gen-n-Val yields significant performance improvements over the baseline models, with the most notable improvement observed being a $+3.1\%$ mask mAP gain on YOLO11m. 

\begin{table}[t]
    \centering
    \small
    \caption{\textbf{The scalability of synthetic data.} As the size of the dataset increases, performance improves.}
    \label{tab:scalability}
  \vspace{-3mm}
\begin{tabular}{l|cccc}
      \toprule % <-- Toprule here
    Size & \textbf{mAP}$^\textbf{box}$ & \textbf{mAP}$^\textbf{box}_\textbf{rare}$ & \textbf{mAP}$^\textbf{mask}$ & \textbf{mAP}$^\textbf{mask}_\textbf{rare}$ \\
    \midrule
    4K  & 50.8 & 54.6 & 42.2 & 48.3 \\
    8K  & 51.1 & 55.0 & 42.5 & 48.6 \\
    16K & 51.7 & 55.4 & 42.9 & 49.0 \\
    20K & \textbf{52.0} & \textbf{55.6} & \textbf{43.0} & \textbf{49.2} \\
    \bottomrule
\end{tabular} 
\end{table}

\subsection{Data Scalability}
To comprehensively evaluate our method across different data scales, we trained YOLO11m on the COCO dataset and varying sizes of synthetic data, generating the same 80 classes as in the COCO dataset (4K, 8K, 16K, and 20K), and evaluated it on the COCO. 
The results in Table~\ref{tab:scalability} show our model's scalability with increasing dataset size.

\subsection{Qualitative Results}
We present qualitative comparisons between the ground truth, baseline model outputs, and Gen-n-Val outputs, evaluated using the YOLO11m. As shown in Figure~\ref{fig:qualitative}, the baseline model fails to segment the person in the driver’s seat of the truck and the car behind the truck. In contrast, Gen-n-Val accurately segments both the person and the car. This result highlights the effectiveness of Gen-n-Val. 
For more results, please refer to our supplementary material. 

%% file: sec/5_Conclusion.tex
\section{Conclusion}
\label{sec:conclusion}
In this work, we introduce a novel agentic framework, Gen-n-Val, for generating synthetic data, allowing us to create high-quality, diverse, and precisely annotated synthetic datasets. By leveraging Layer Diffusion (LD), Large Language Model (LLM), and Vision Large Language Model (VLLM), Gen-n-Val significantly enhances the quality and usability of synthetic data. The framework consists of two key agents, the LD prompt agent (a LLM) and the data validation agent (a VLLM). The LD prompt agent optimizes the prompt for LD to generate high-quality images and segmentation masks. The data validation agent filters out failure cases of image instances. Our experiments demonstrate that Gen-n-Val outperforms previous data synthesis approaches in instance segmentation and object detection, and scales with dataset size. On LVIS, Gen-n-Val further serves as an effective mechanism for addressing long-tailed category imbalance; it substantially flattens the class distribution. This work highlights the potential of leveraging LLM- and VLLM-based agents not only to improve synthetic data quality, but also to mitigate data imbalance in large-vocabulary computer vision datasets.

%% file: sec/6_Acknowledgement.tex
\section*{Acknowledgment}
This project was supported in part by the National Science and Technology Council (NSTC), Taiwan, under Grants 114-2221-E-001-016, 114-2221-E-001-004, 113-2634-F-002-008, 114-2634-F-001-001-MBK, 111-2628-E-001-002-MY3, 114-2221-E-001-017-MY2, and 114-2634-F-002-004, and by Academia Sinica under Grant AS-IAIA-114-M10 and AS-KPQ-112-NETZ-10-A. We thank the National Center for High-performance Computing (NCHC) of the National Institutes of Applied Research (NIAR) in Taiwan for providing computational and storage resources.

%% file: sec/X_suppl.tex
\setcounter{page}{1}
\maketitlesupplementary

\section*{Overview}
This supplementary material extends the main manuscript with implementation details, expanded analyses, and qualitative examples: (i)
additional results and analyses
(Sec.~\ref{sec:add_results});
(ii) discussions of limitations (Sec.~\ref{sec:limitation}) and societal impacts (Sec.~\ref{sec:societalimpacts}); (iii) full implementation details of the Gen-n-Val pipeline (prompt generation, instance generation, filtering, optional background synthesis, augmentation, and training setup) (Sec.~\ref{sec:implementation_detail}); (iv) qualitative comparisons of baseline and Gen-n-Val outputs (Sec.~\ref{sec:qualitative}); (v) prompt optimization comparisons of initial and optimized prompts (Sec.~\ref{sec:promptoptimization}); (vi) foreground image filtering examples (Sec.~\ref{sec:visionllmfiltering}); and (vii) synthetic data examples for COCO and LVIS (Sec.~\ref{sec:moreexamples}).

%%%%%%%%%%%%%%%%%%%%%%%%%%%%%%%%%%%%%%%%%%%%%%%%%%%%%%%%%%%%%%%%%%%%%%%%%%%%%%%%%%%%%%%%%

\section{Additional Results and Analyses}
\label{sec:add_results}

%%%%%%%%%%%%%%%%%%%%%%%%%%%%%%%%%%%%%%%%%%%%%%%%%%%%%%%%%%%%%%%%%%%%%%%%%%%%%%%%%%%%%%%%%

\begin{table}[h]
    \caption{\textbf{Ablation study of Mask R-CNN~\cite{he2017maskrcnn} on the LVIS~\cite{gupta2019lvis} benchmark.} Comparison of performance under different numbers of pasted instances.}
  \begin{center}

    \resizebox{\columnwidth}{!}{
    \begin{tabular}{c|cccc}
      \toprule % <-- Toprule here
      \# of Pasted Instances & $\textbf{mAP}^\textbf{box}$ & $\textbf{mAP}_\textbf{r}^\textbf{box}$ & $\textbf{mAP}^\textbf{mask}$& $\textbf{mAP}_\textbf{r}^\textbf{mask}$\\
      \midrule % <-- Midrule here
      1 & 24.7 & 13.0 & 23.9 & 14.7\\
      3 & 25.4 & 13.3 & 24.1 & 13.6\\
       5 & \textbf{25.5} & \textbf{14.9} & \textbf{26.6}& \textbf{15.5}\\
      \bottomrule % <-- Bottomrule here
    \end{tabular}
    }
    \label{tab:ablation_lvis}
  \end{center}
\end{table}

\subsection{Number of Pasted Instances}
\label{subsec:num_of_paste}
We evaluate the performance of Mask R-CNN~\cite{he2017maskrcnn} on the LVIS benchmark~\cite{gupta2019lvis} using synthetic training data generated with different numbers of pasted instances per image. As shown in Table~\ref{tab:ablation_lvis}, pasting 5 instances per image yields the best performance.

\subsection{Synthetic-Only and Instance-Count-Matched}
\label{subsec:real_vs_syn_control}

To further analyze whether the gains come from the quality of synthetic data rather than simply seeing more training instances, we conduct a controlled study on LVIS rare categories using Mask R-CNN~\cite{he2017maskrcnn}. Specifically, we compare the same number of real and synthetic training instances, and also evaluate an instance-count-matched setting.

\begin{table}[h]
    \caption{\textbf{Controlled study on LVIS rare categories with Mask R-CNN~\cite{he2017maskrcnn}.} Comparison between real and Gen-n-Val synthetic training instances.}
    \centering
    \resizebox{\columnwidth}{!}{
    \begin{tabular}{l|cc}
      \toprule
      \textbf{Training Data} & $\textbf{mAP}^{\textbf{box}}$ & $\textbf{mAP}^{\textbf{mask}}$\\
      \midrule
      4k Real (LVIS) & 1.78 & 1.32\\
      4k Synthetic (Gen-n-Val) & \textbf{1.82} & \textbf{1.78}\\
      \midrule
      8k Real & 5.06 & 5.52\\
      4k Real + 4k Synthetic (Gen-n-Val) & \textbf{6.28} & \textbf{6.64}\\
      \bottomrule
    \end{tabular}
    }
    \label{tab:lvis_real_vs_syn_supp}
\end{table}

As shown in Table~\ref{tab:lvis_real_vs_syn_supp}, with 4k training instances, Gen-n-Val synthetic data already performs comparably to or better than 4k real LVIS instances. Moreover, under the same total number of instances, \textit{4k real + 4k synthetic} outperforms \textit{8k real}, suggesting that the gains are not merely due to seeing more data, but also due to the usefulness of the generated instances for long-tail categories.

In addition, all LVIS experiments in our paper are iteration-based and use the same number of training iterations for the baseline (real-only) and Gen-n-Val (real+synthetic). Therefore, Gen-n-Val does not receive extra optimization steps in the LVIS setting.

\begin{figure}[t]
    \centering
    \includegraphics[width=\columnwidth]{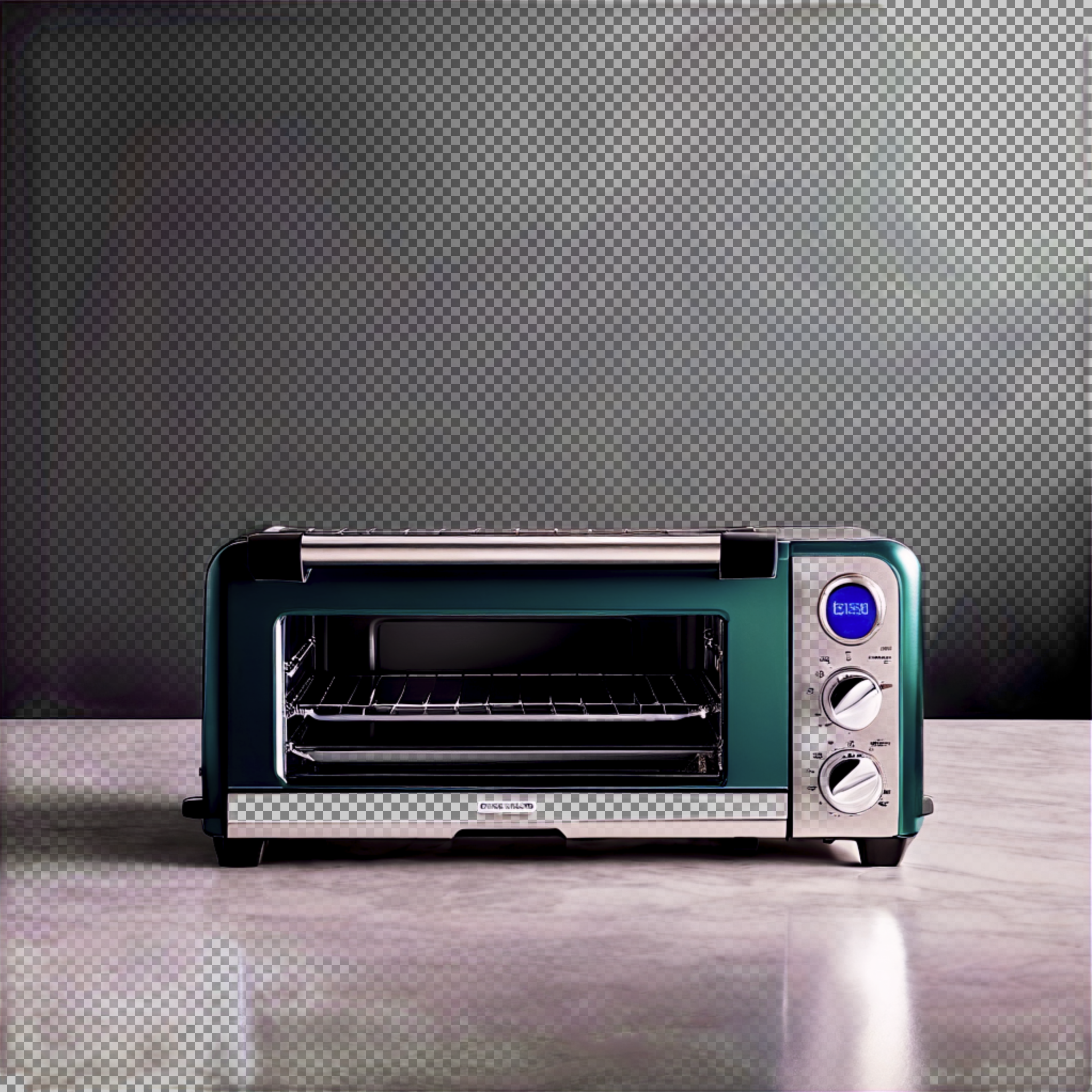}
    \caption{\textbf{Semi-transparent background residue.} A representative false-acceptance case caused by imperfect layer separation.}
    \label{fig:failure_case_supp}
\end{figure}

\subsection{Validation Reliability and Failure Analysis}
\label{subsec:validator_reliability}

We manually audited 300 accepted and 300 rejected foreground instances from the validation agent. We observe 5.7\% (17/300) false acceptances and 32.0\% (96/300) false rejections. The dominant false-acceptance failure mode is imperfect layer separation, where a small amount of semi-transparent background residue remains around the generated foreground. An example is shown in Fig.~\ref{fig:failure_case_supp}.

\begin{figure}[t]
    \centering
    \includegraphics[width=\columnwidth]{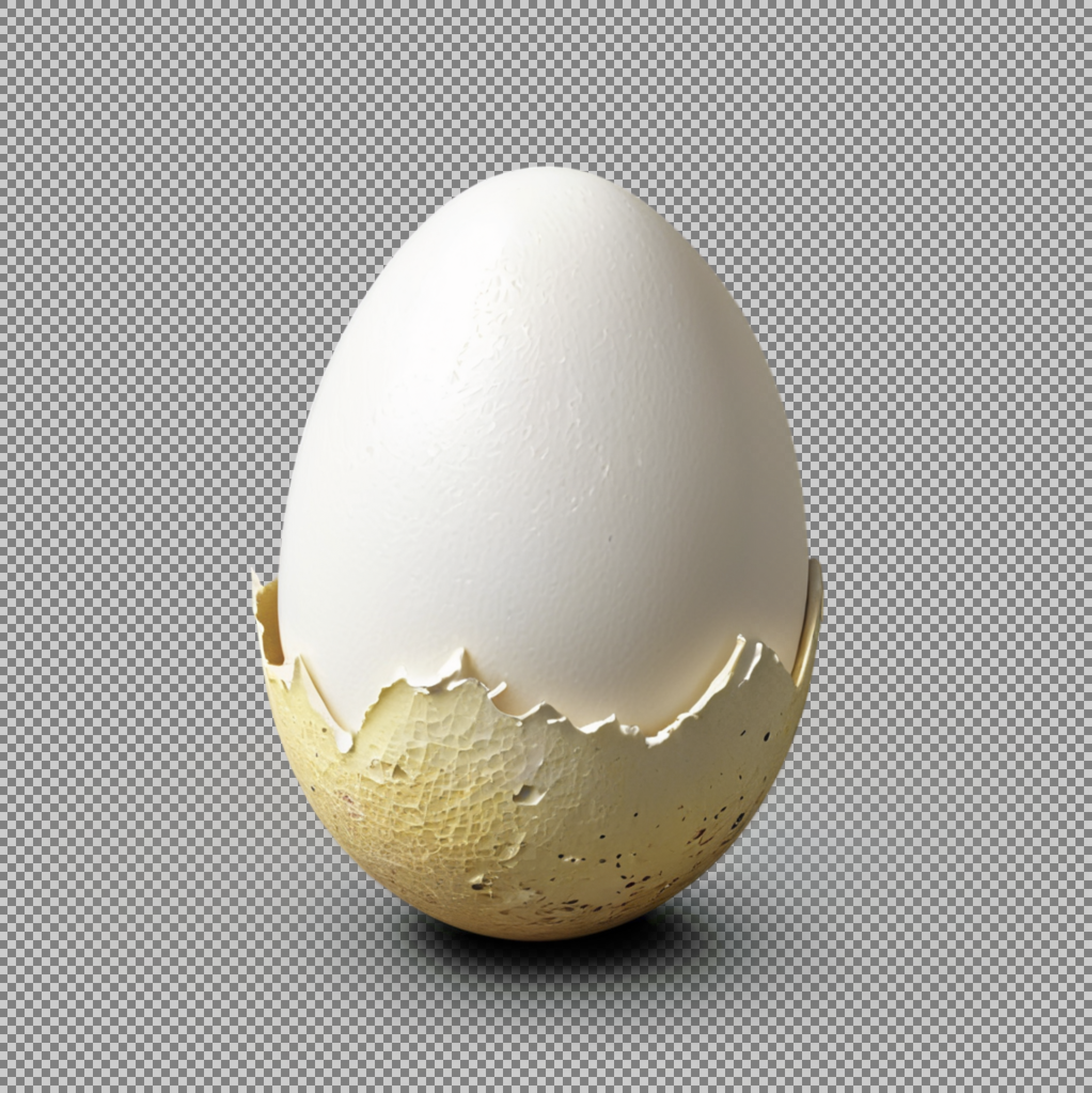}
    \caption{\textbf{Conservative validator rejection example.} The VLLM rejects this egg instance because the cracked shell violates the ``intact object'' constraint, illustrating that many false rejections come from over-strict filtering.}
    \label{fig:false_rejection_supp}
\end{figure}

Most false rejections are caused by overly strict validation rather than unusable samples. For example, in Fig.~\ref{fig:false_rejection_supp}, the validator rejects an egg instance because it considers the cracked shell as violating the ``intact object'' constraint, even though the instance is still usable for training. We intentionally keep the validator conservative, since false acceptance is typically more harmful than false rejection when constructing a synthetic training set.

\begin{figure}[t]
    \centering
    \includegraphics[width=\columnwidth]{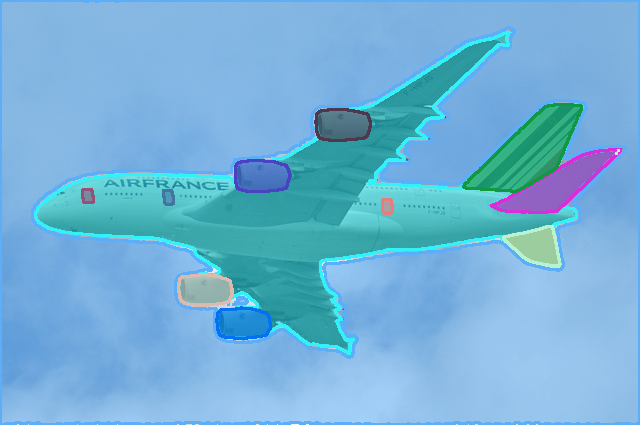}
    \caption{\textbf{SAM may split a single object into multiple masks.} This fragmentation makes the extracted foreground less suitable for direct synthetic instance generation.}
    \label{fig:sam_split_supp}
\end{figure}

\subsection{Alternative Foreground Extraction Pipeline}
\label{subsec:alternative_pipeline}

We also analyze an alternative pipeline based on SAM~\cite{kirillov2023segany}. Although SAM is a strong segmenter, it often produces fragmented masks, where a single object is split into multiple disconnected regions, as shown in Fig.~\ref{fig:sam_split_supp}. Such outputs require additional merging and cleanup before they can be used as instance masks. In contrast, our pipeline directly generates transparent foregrounds with alpha-aligned masks, which are more suitable for copy-paste style instance synthesis.

\subsection{Compute Comparison}
\label{subsec:compute_comparison}

We compare the practical generation cost of Gen-n-Val and X-Paste~\cite{zhao2023xpaste}. In our reproduction, generating 150k synthetic instances with X-Paste required about \textbf{20 days} on \textbf{4 NVIDIA RTX A6000 GPUs}, whereas Gen-n-Val required about \textbf{5 days} under the same hardware setting. This result suggests that Gen-n-Val is substantially more efficient for large-scale synthetic instance generation.

\begin{figure}[h]
  \centering
  \includegraphics[width=\columnwidth]{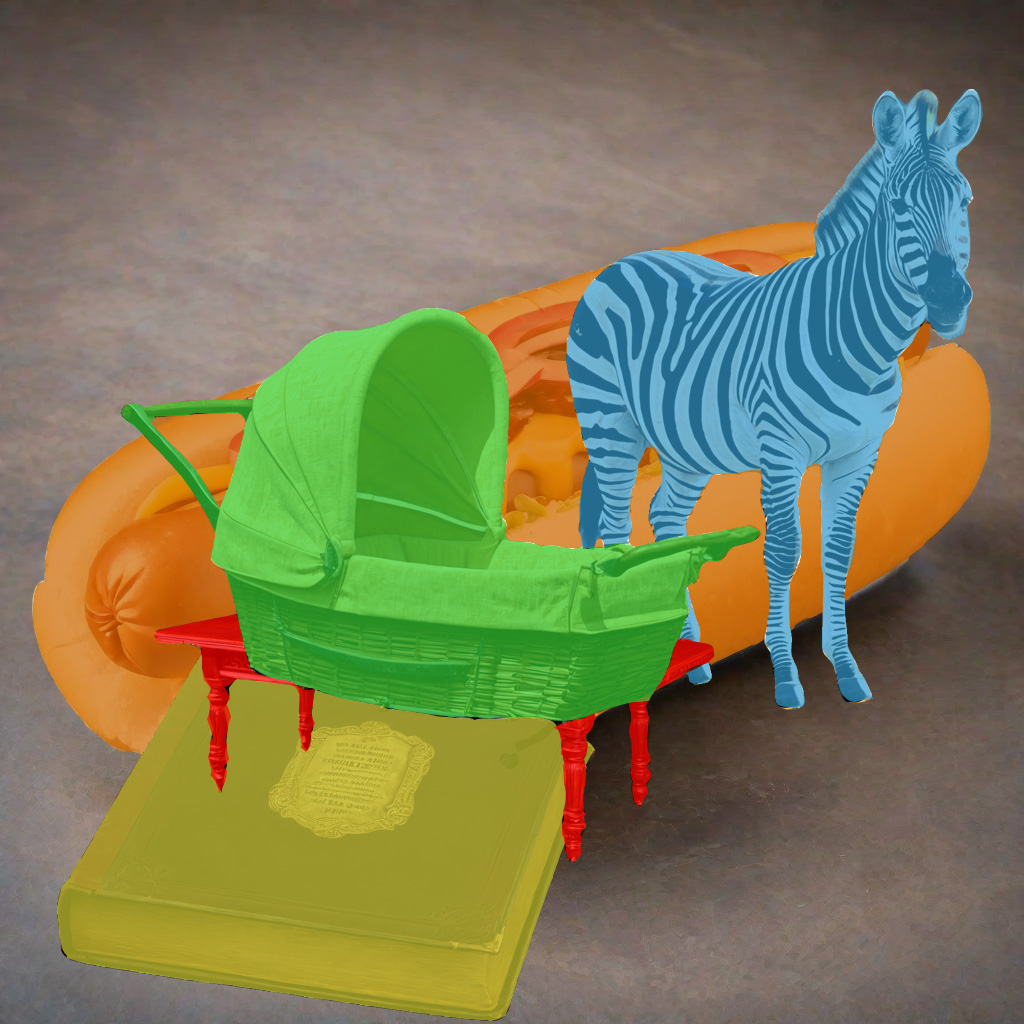}
  \caption{\textbf{Example of contextual coherence in instance placement.} The placement of the zebra standing on the huge hot dog is semantically incoherent, which may lead to unrealistic or nonsensical images.}
  \label{fig:limitationinstanceplacement}

\end{figure}

\section{Limitations}
\label{sec:limitation}
\textbf{Contextual Coherence in Instance Placement.}
In this work, the main focus is to generate high-quality and diverse training instances. As with other data augmentation methods, although the proposed method utilizes image harmonization to integrate multiple instances into background scenes, the instance placement process does not inherently consider the semantic or contextual relationships between the objects and their environment. This randomness in placement can result in synthesized images that lack logical coherence or realism, potentially introducing noise into the training process. For example, objects may appear in physically implausible positions or in contexts where their presence is incongruous, which could hinder model generalization when applied to real-world scenarios. As shown in Figure~\ref{fig:limitationinstanceplacement}, the placement of the zebra standing on the huge hot dog is semantically incoherent, which may lead to unrealistic or nonsensical images. Future work could explore incorporating contextual constraints or relationships between objects to improve the coherence and realism of the generated data.

\section{Societal Impacts}
\label{sec:societalimpacts}

Our research focuses on developing a novel synthetic data generation pipeline, which we believe does not pose significant negative societal impacts. The methodology is designed for general-purpose instance segmentation tasks and does not directly facilitate applications with harmful implications, such as surveillance, privacy violations, or discrimination. The synthetic data generated does not involve any human-derived data or personal information, thereby eliminating concerns related to privacy or human rights. Furthermore, our approach emphasizes efficiency and scalability without incentivizing environmental harm. We remain committed to responsible research practices and transparency, ensuring our work contributes positively to the advancement of computer vision.

\section{Implementation Details}
\label{sec:implementation_detail}

\subsection{Open Vocabulary Prompt Generation}
With the initial system prompt, 37\% sampled LD prompts were invalid (e.g., extra system-style replies or additional objects). 
Using TextGrad~\cite{yuksekgonul2024textgrad}, we optimize the system prompts of the Meta-LLaMA-3.1-8B-Instruct~\cite{dubey2024llama} as the Layer Diffusion (LD) prompt agent with default parameters (configured with temperature of 0.7, top-$p$ of 0.9, and max-new-tokens set to 256) five times, enabling it to generate detailed and precise prompts for Layer Diffusion~\cite{zhang2024layerdiffusion}. Examples of the initial and optimized system prompts are provided in Table~\ref{tab:llmsystempromptcomparison} of Section~\ref{sec:promptoptimization}. Examples of the optimized LD prompts produced from the optimized system prompts are provided in Figures~\ref{fig:ldpromptcomparisonairplane}, ~\ref{fig:ldpromptcomparisonorange}, ~\ref{fig:ldpromptcomparisoncar}, and ~\ref{fig:ldpromptcomparisonperson} of Section~\ref{sec:promptoptimization}.

\newcommand\QualitativeSuppBoxWidth{.33\textwidth}
\newcommand\QualitativeSuppImageWidth{.33\textwidth}
\begin{figure*}[t!]
  \centering
  \begin{tabular}{@{}c@{}c@{}c@{}}

    Ground Truth & Baseline & Gen-n-Val \\

    \parbox[c]{\QualitativeSuppBoxWidth}{
    \includegraphics[width=\QualitativeSuppImageWidth]{figure/supplementary/qualitative/000000018837_val.jpg} 
    } & 
    \parbox[c]{\QualitativeSuppBoxWidth}{
    \includegraphics[width=\QualitativeSuppImageWidth]{figure/supplementary/qualitative/000000018837_base.jpg} 
    } & 
    \parbox[c]{\QualitativeSuppBoxWidth}{
    \includegraphics[width=\QualitativeSuppImageWidth]{figure/supplementary/qualitative/000000018837_gen.jpg} 
    } \\ 

    \parbox[c]{\QualitativeSuppBoxWidth}{
    \includegraphics[width=\QualitativeSuppImageWidth]{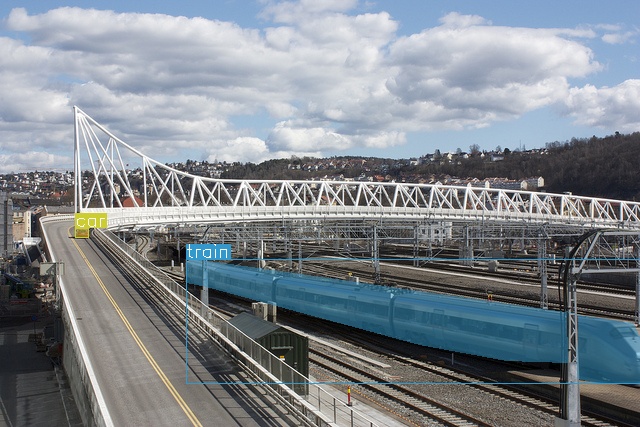} 
    } & 
    \parbox[c]{\QualitativeSuppBoxWidth}{
    \includegraphics[width=\QualitativeSuppImageWidth]{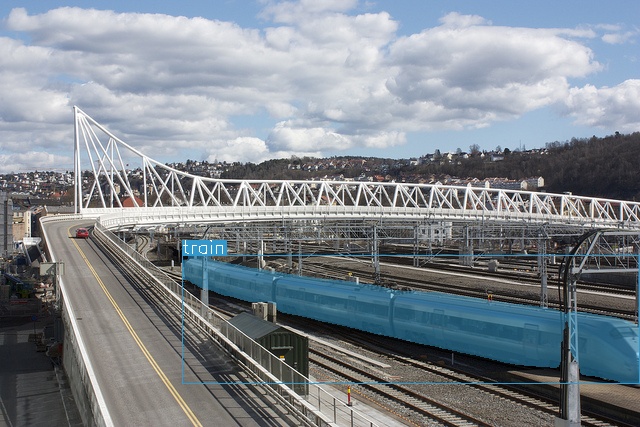} 
    } & 
    \parbox[c]{\QualitativeSuppBoxWidth}{
    \includegraphics[width=\QualitativeSuppImageWidth]{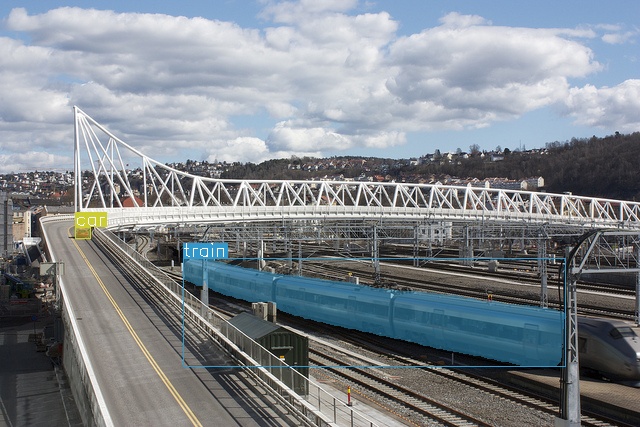} 
    } \\ 

    \parbox[c]{\QualitativeSuppBoxWidth}{
    \includegraphics[width=\QualitativeSuppImageWidth]{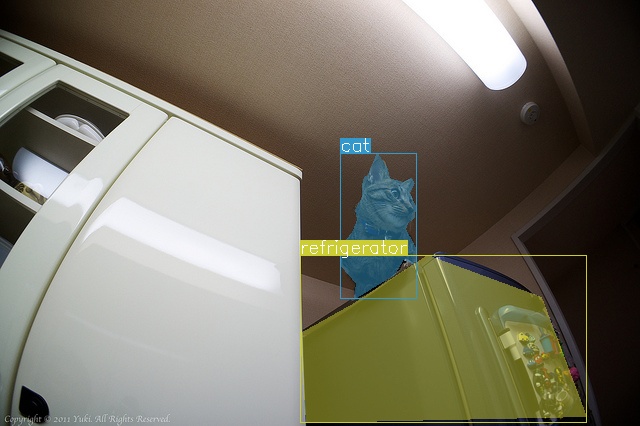} 
    } & 
    \parbox[c]{\QualitativeSuppBoxWidth}{
    \includegraphics[width=\QualitativeSuppImageWidth]{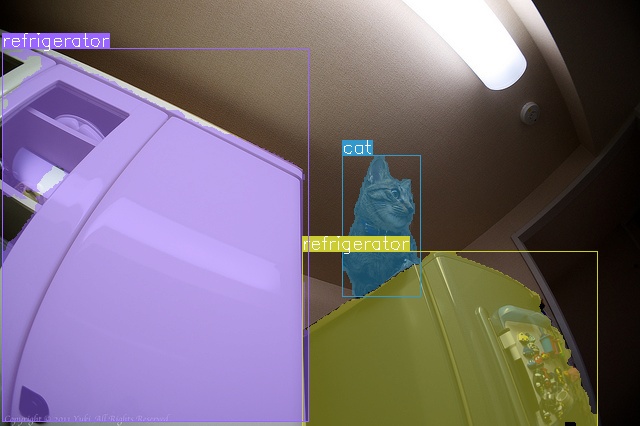} 
    } & 
    \parbox[c]{\QualitativeSuppBoxWidth}{
    \includegraphics[width=\QualitativeSuppImageWidth]{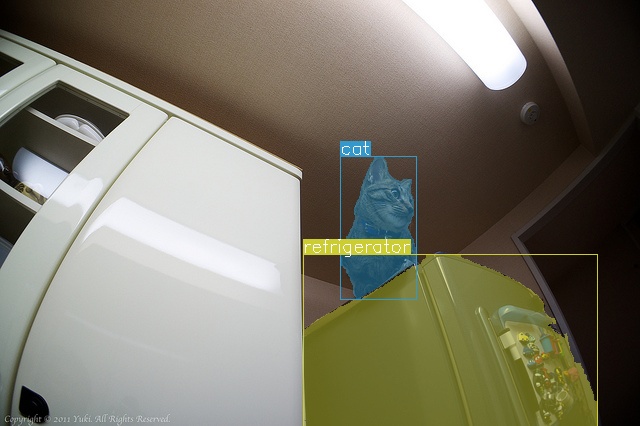} 
    } \\ 
    
    \parbox[c]{\QualitativeSuppBoxWidth}{
    \includegraphics[width=\QualitativeSuppImageWidth]{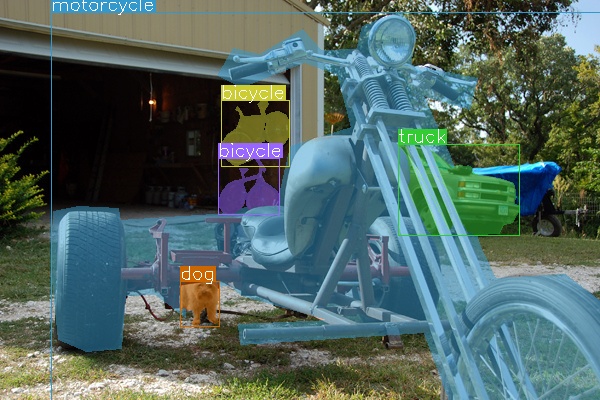} 
    } & 
    \parbox[c]{\QualitativeSuppBoxWidth}{
    \includegraphics[width=\QualitativeSuppImageWidth]{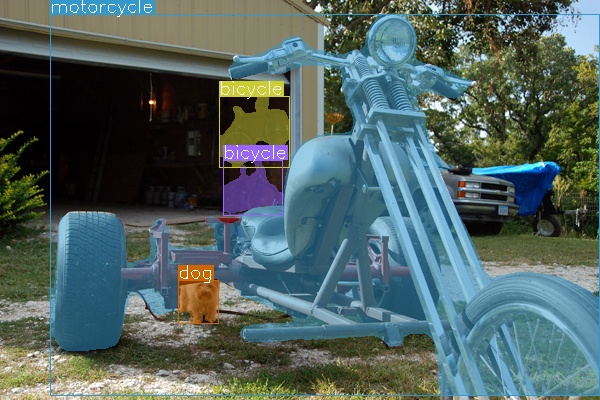} 
    } & 
    \parbox[c]{\QualitativeSuppBoxWidth}{
    \includegraphics[width=\QualitativeSuppImageWidth]{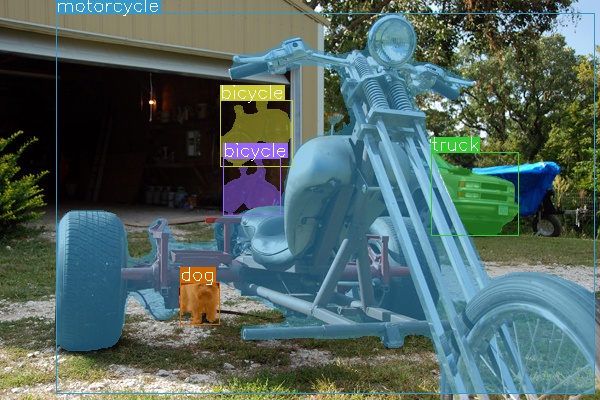} 
    } \\ 

    \parbox[c]{\QualitativeSuppBoxWidth}{
    \includegraphics[width=\QualitativeSuppImageWidth]{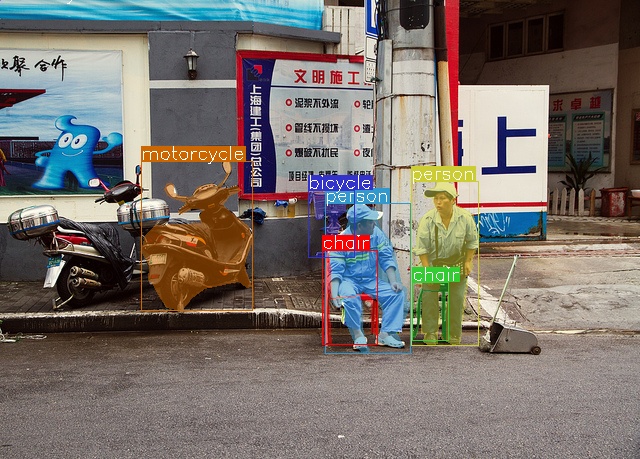} 
    } & 
    \parbox[c]{\QualitativeSuppBoxWidth}{
    \includegraphics[width=\QualitativeSuppImageWidth]{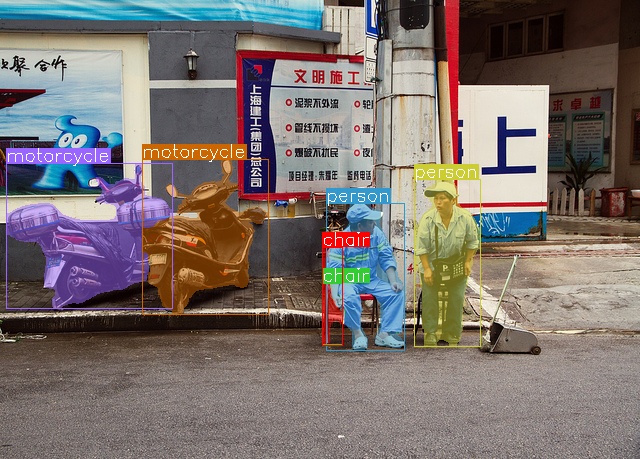} 
    } & 
    \parbox[c]{\QualitativeSuppBoxWidth}{
    \includegraphics[width=\QualitativeSuppImageWidth]{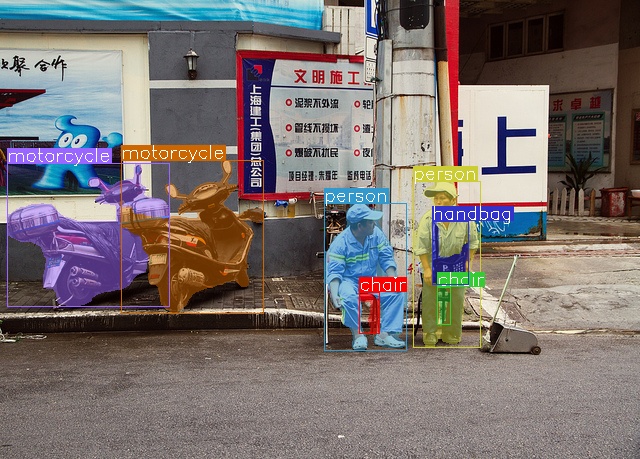} 
    } \\ 

    Ground Truth & Baseline & Gen-n-Val \\
    
  \end{tabular}
  \caption{\textbf{Qualitative comparison of ground truth, baseline, and Gen-n-Val.}}
  \label{fig:qualitative_supp}
\end{figure*}

\subsection{Foreground Image Generation}
These optimized LD prompts are then fed into Layer Diffusion with default parameters (configured with strength of 1, num-inference-steps of 25, and a guidance scale of 7) to sample the desired images. Examples of foreground instances generated using the standard and optimized LD prompts are provided in Figures~\ref{fig:personstitchedstandard} and~\ref{fig:personstitchedoptimized} of Section~\ref{sec:promptoptimization}.

\subsection{Foreground Image Filtering}
Using TextGrad~\cite{yuksekgonul2024textgrad}, we optimize the system prompts of the Meta-LLaMA-3.2-11B-Vision-Instruct~\cite{meta2024llama32} as the data validation agent with default parameters (configured with temperature of 0.7, top-$p$ of 0.9, and max-new-tokens set to 256) five times, enabling it to effectively evaluate and filter the generated images based on specific criteria. Examples of the initial and optimized system prompts are provided in Tables~\ref{tab:vllmsystempromptinitial} and~\ref{tab:vllmsystempromptoptimized} of Section~\ref{sec:promptoptimization}. After obtaining an intact foreground instance, we apply a median filter with a kernel size of 15 to denoise the output image and obtain a precise segmentation mask. Examples of the data validation agent's filtering process for different subjects are provided in Figures~\ref{fig:vllmfilteringorange}, ~\ref{fig:vllmfilteringalarmclock}, ~\ref{fig:vllmfilteringbirthdaycard}, and ~\ref{fig:vllmfilteringpancake} of Section~\ref{sec:visionllmfiltering}.

\subsection{Background Image Generation (Optional)}
To synthesize background images, we use simple prompts (\eg ``\texttt{A <object> in an empty <indoor or outdoor> background}'') along with the generated foreground instances as input to Layer Diffusion with default parameters (configured with strength of 1, num-inference-steps of 25, and a guidance scale of 7). We use this optional setting for our COCO~\cite{lin2014microsoft} experiments. To reduce the computational cost on LVIS, we remove the extra synthetic background generation step.

\subsection{Instance Augmentation}
We synthesize augmented training images by pasting multiple foreground instances produced by our Gen-n-Val pipeline onto original images or synthetic backgrounds (for COCO experiments) generated in the Background Image Generation phase. The procedure preserves all original LVIS/COCO annotations and adds annotations only for newly pasted synthetic instances.

\noindent\textbf{Foreground selection and count.} 
We use two configurations for choosing synthetic foreground instances to paste. (i) \emph{Standard configuration}: paste five instances per image; Table~\ref{tab:ablation_lvis} shows this count provides the strongest empirical performance. (ii) \emph{Long-tail configuration}: perform targeted augmentation for under-represented categories by adding extra instances of categories whose image-level frequency is below LVIS thresholds (rare: fewer than 10 images; common: fewer than 100 images).

\noindent\textbf{Scaling and positioning.} Let the original synthetic width/height be \((w_s, h_s)\) and the image size \((W, H)\). In this work we use purely random scaling and placement. The scale factor \(s\) is drawn from a log-normal distribution \(s \sim \text{LogNormal}(\mu=\log(m), \sigma=0.4)\) where \(m\) is the mean scale factor (set to 0.5 in our experiments); if the proposed size would exceed image bounds we clip it to \(0.9 \times \min(W/w_s, H/h_s)\). The (x,y) paste location is sampled uniformly over all valid upper-left coordinates so the resized instance lies fully inside the image.

\noindent\textbf{Occlusion trimming and retries.} We decode all existing annotation masks plus masks of previously pasted synthetic instances. Overlapping pixels are removed from the new synthetic mask. We enforce a minimum retained area ratio: if the visible area \(A_{\text{vis}}\) after trimming satisfies \(A_{\text{vis}} < 0.1\, A_{\text{orig}}\) or is fully occluded, we shrink the tentative mean size (multiplying by 0.8) and resample scale/position, retrying up to 10 attempts. This prevents adding negligible fragments and avoids duplicate coverage. Original annotations are never modified.
This strategy yields: (1) diverse multi-instance compositions without corrupting existing ground truth, (2) occlusion-aware clean masks, and (3) selective boosting of under-represented categories for improved long-tail performance.

\subsection{Model Training}

\noindent\textbf{LVIS experiments.} We use the Mask R-CNN~\cite{he2017maskrcnn} model with a ResNet-50-FPN backbone to evaluate the effectiveness of Gen-n-Val on the LVIS benchmark~\cite{gupta2019lvis}. We follow the standard LVIS setup and the common ``1\texttimes{}'' training schedule: 90k iterations with a global batch size of 16 (2 images per GPU $\times$ 8 GPUs equivalent). Optimization uses SGD with momentum 0.9 and weight decay 0.0001. The initial learning rate is 0.02 and is reduced by a factor of 10 at 60k and 80k iterations. Repeat factor sampling is enabled with an oversample threshold of $10^{-3}$. Data augmentation consists of horizontal flipping and multi-scale resizing where the shorter image side is randomly sampled in the range [640, 800] while ensuring the longer side does not exceed 1333 pixels.

\noindent\textbf{COCO experiments.} For COCO we evaluate Gen-n-Val using YOLO11m and YOLOv9c as additional baselines. YOLO11m is trained for 100 epochs with a batch size of 200, while YOLOv9c is trained for 100 epochs with a batch size of 100. All images are resized to 640$\times$640 pixels. We use SGD with momentum 0.9 (learning rate 0.01).

%%%%%%%%%%%%%%%%%%%%%%%%%%%%%%%%%%%%%%%%%%%%%%%%%%%%%%%%%%%%%%%%%%%%%%%%%%%%%%%%%%%%%%%%%

%%%%%%%%%%%%%%%%%%%%%%%%%%%%%%%%%%%%%%%%%%%%%%%%%%%%%%%%%%%%%%%%%%%%%%%%%%%%%%%%%%%%%%%%

%%%%%%%%%%%%%%%%%%%%%%%%%%%%%%%%%%%%%%%%%%%%%%%%%%%%%%%%%%%%%%%%%%%%%%%%%%%%%%%%%%%%%%%%

\section{Qualitative Results}
\label{sec:qualitative}

In Figure~\ref{fig:qualitative_supp}, we present qualitative comparisons between the ground truth, baseline model outputs, and Gen-n-Val outputs, evaluated using the YOLO11m~\cite{khanam2024yolov11} model. These results demonstrate the robustness of Gen-n-Val in addressing various challenges in instance segmentation.

\begin{itemize}

  \item \textbf{First Row:} The baseline model overlooks the car on the highway entirely, while Gen-n-Val successfully segments the car, demonstrating its ability to detect small and distant objects.

  \item \textbf{Second Row:} The baseline model mistakenly segments the cabinet as a refrigerator, highlighting confusion in object classification. Gen-n-Val correctly segments the cabinet, avoiding this misclassification.

  \item \textbf{Third Row:} The baseline model fails to segment the truck on the right-hand side of the image. Gen-n-Val, however,   segments the truck successfully, illustrating its superior handling of challenging complex scenes.

  \item \textbf{Fourth Row:} The baseline model fails to segment the chair under the right-hand-side person and mistakenly identifies the chair under the left-hand-side person as two separate chairs. Gen-n-Val segments the right chair correctly and accurately identifies the left chair as a single object. Additionally, Gen-n-Val successfully segments a handbag carried by the right-hand-side person, which is present in the image but missing in the ground truth annotation.
\end{itemize}

This highlights Gen-n-Val’s ability to capture fine details and segment unannotated objects, demonstrating its potential to improve object detection and segmentation in scenarios with incomplete ground truth labels.

%%%%%%%%%%%%%%%%%%%%%%%%%%%%%%%%%%%%%%%%%%%%%%%%%%%%%%%%%%%%%%%%%%%%%%%%%%%%%%%%%%%%%%%%

\section{Prompt Optimization Results}
\label{sec:promptoptimization}

In Table~\ref{tab:llmsystempromptcomparison}, we provide a comparison between the LD prompt agent's initial system prompt and the optimized system prompt for generating detailed positive prompts for the Layer Diffusion model. This optimized system prompt is designed to guide the LD prompt agent in generating high-quality prompts that focus solely on the main subject, ensuring that the generated images are detailed, realistic, and visually appealing. The guidelines provided in the optimized system prompt help the LD prompt agent to create diverse and specific prompts that adhere to the requirements of the task, resulting in high-quality image generation.

In Tables~\ref{tab:vllmsystempromptinitial} and~\ref{tab:vllmsystempromptoptimized}, we present a comparison between the data validation agent's initial system prompt and the optimized system prompt for analyzing images based on specific criteria. This optimized system prompt provides clear instructions for describing images, evaluating them against specific criteria, and deciding whether to keep or filter out images based on the evaluation results, which ensures that the data validation agent analyzes images accurately and consistently, leading to improved performance in evaluating image suitability.

\begin{figure}[t!]
  \centering
  \includegraphics[width=\columnwidth]{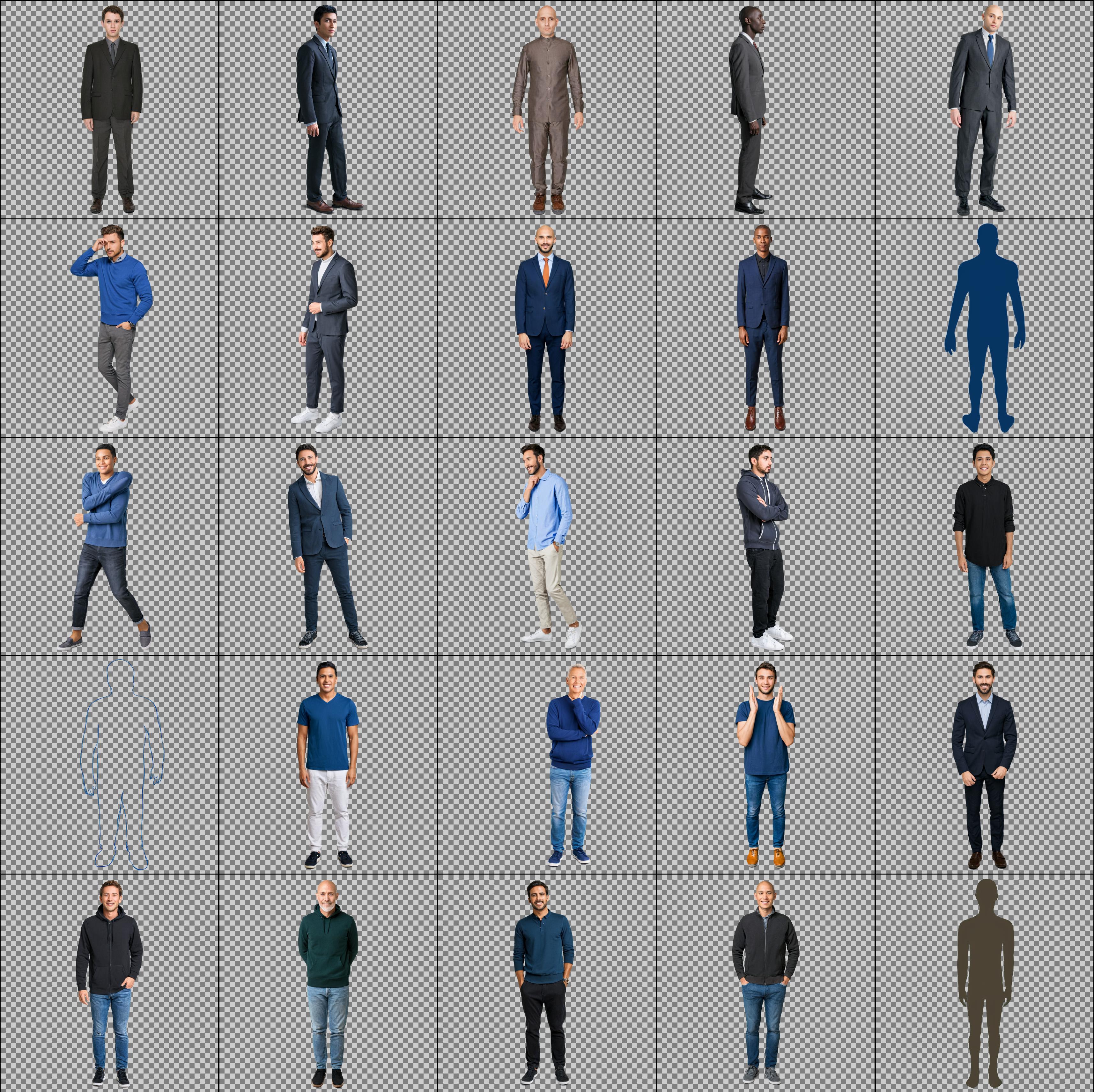}
  \caption{\textbf{Example of the person foreground instance generated using the standard LD prompt.}}
  \label{fig:personstitchedstandard}
\end{figure}
\begin{figure}[h!]
  \centering
  \includegraphics[width=\columnwidth]{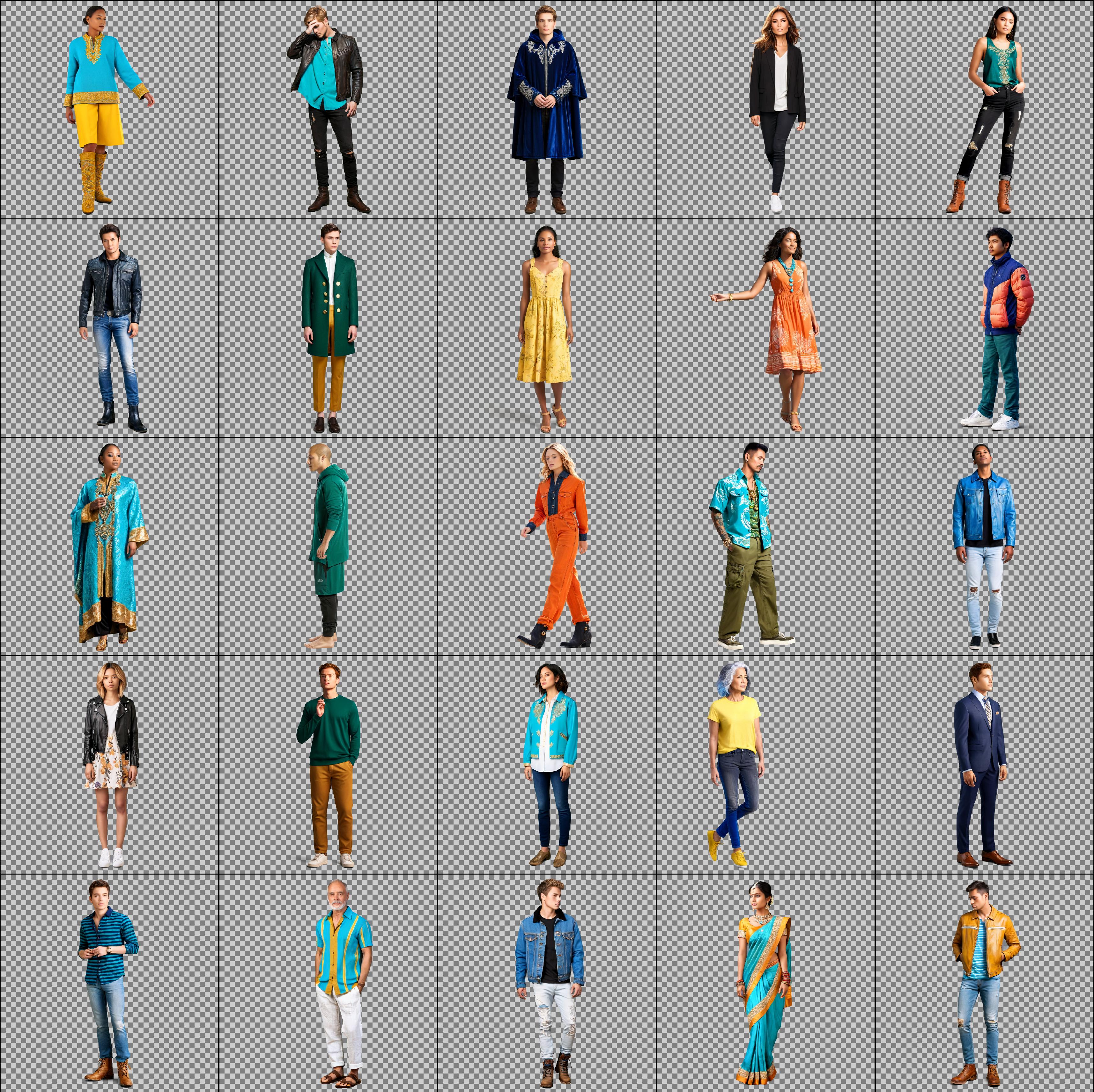}
  \caption{\textbf{Example of the person foreground instance generated using the optimized LD prompt.}}
  \label{fig:personstitchedoptimized}
\end{figure}

In Figures~\ref{fig:personstitchedstandard} and~\ref{fig:personstitchedoptimized}, we present examples of person foreground instances generated using the standard LD prompt and the optimized LD prompt, respectively. The optimized LD prompt demonstrates its ability to produce a broader range of images with enhanced diversity. Unlike the standard prompt, which often results in generic and less varied outputs, the optimized prompt generates images that vary significantly in terms of style, color, texture, lighting, and perspective. This diversity ensures the inclusion of individuals with distinct appearances, clothing styles, and postures, thereby enriching the dataset and improving the generalization capability of downstream models trained on these synthetic examples.

In Figures~\ref{fig:ldpromptcomparisonairplane}, \ref{fig:ldpromptcomparisonorange}, \ref{fig:ldpromptcomparisoncar}, and \ref{fig:ldpromptcomparisonperson}, we compare the standard LD prompt with the optimized LD prompts for the subjects airplane, orange, car, and person, respectively. The optimized prompts are designed to provide detailed information about the subject's status, color, style, mood/atmosphere, lighting, perspective/viewpoint, textures/material, time period, and medium, ensuring that the generated images are highly realistic and visually appealing. The optimized prompts include trigger words like ``high-resolution'' and ``highly realistic'' to emphasize the quality of the generated images. The examples demonstrate how the optimized prompts lead to the generation of diverse images that focus solely on the main subject, enhancing the quality and realism of the generated images.

\section{Foreground Image Filtering Results}
\label{sec:visionllmfiltering}

In Figures~\ref{fig:vllmfilteringorange}, ~\ref{fig:vllmfilteringalarmclock}, ~\ref{fig:vllmfilteringbirthdaycard}, and ~\ref{fig:vllmfilteringpancake}, we provide examples of the data validation agent's filtering process for different subjects. The data validation agent evaluates each generated image based on specific criteria, including the presence of a single subject, a single view, an intact subject, and a plain background. It analyzes the image and provides a detailed description of the content, highlighting the presence or absence of the specified criteria. Based on the evaluation, the data validation agent determines whether the image meets all the criteria and should be retained or fails to meet the criteria and should be filtered out. This filtering process ensures that only high-quality images that adhere to the task requirements are retained for further processing, enhancing the overall quality of the generated dataset.

\section{More Gen-n-Val Synthetic Data}
\label{sec:moreexamples}

In Figure~\ref{fig:moredata1}, ~\ref{fig:moredata2}, ~\ref{fig:moredata3}, ~\ref{fig:moredata4}, ~\ref{fig:moredata5}, and ~\ref{fig:moredata6}, we provide additional examples of Gen-n-Val synthetic data generated for COCO~\cite{lin2014microsoft}. Moreover, in Figure~\ref{fig:lvisdata_1}, ~\ref{fig:lvisdata_2}, and ~\ref{fig:lvisdata_3}, we provide additional examples of Gen-n-Val synthetic data generated for LVIS~\cite{gupta2019lvis}.

%% file: sec/Y_suppl_large.tex
\begin{table*}[ht!]
  \centering
  \renewcommand{\arraystretch}{1.5} % Adjust row height
    \caption{\textbf{Comparison of the initial and optimized system prompts of LD prompt agent.}}
  \begin{tabular}{|p{0.96\linewidth}|}
    \hline
    \textbf{The LD Prompt Agent's Initial System Prompt} \\ \hline
    Generate detailed positive prompts for the Stable Diffusion Juggernaut-XL-v6 model to create images focusing solely on the main subject. Each prompt must be specific and cover aspects such as the subject’s status, color, style, mood/atmosphere, lighting, perspective/viewpoint, textures/material, time period, and medium. Prompts should emphasize the use of trigger words like ``high-resolution'' and ``highly realistic'' to ensure quality. Prompts should be concise, limited to under 75 tokens, and must not include disallowed or sensitive content. Background descriptions should be absent, avoiding the inclusion of additional objects. \\ \hline

    \textbf{The LD Prompt Agent's Optimized System Prompt} \\ \hline
    You are an AI assistant designed to generate detailed and realistic prompts for the **Stable Diffusion XL model**, focusing only on a single subject. The background and environment should be omitted in the prompts. Your prompts should be specific, descriptive, diverse, and follow the provided guidelines to ensure high-quality image generation. \\[5pt]

    **Guidelines for Prompt Creation:** \\ 
    1. **Subject:** The only single object in the image. Ensure a wide variety of subjects, ranging from everyday items to unique or uncommon objects. \\ 
    2. **Status:** The current state or condition of the subject. \\ 
    3. **Color:** Dominant colors of the subject. Include specific shades and variations to enhance visual detail. \\ 
    4. **Style:** Artistic style or rendering method. Incorporate a range of styles (e.g., photorealistic, hyper-realistic) to promote diversity. \\ 
    5. **Mood/Atmosphere:** Emotional quality related to the subject. Convey realistic emotions or states that align with the subject. \\ 
    6. **Lighting:** Specific lighting on the subject. Describe natural or artificial lighting conditions that highlight the subject's features. \\ 
    7. **Perspective/Viewpoint:** Angle or perspective of the subject. Use varied viewpoints (e.g., top-down, eye-level, close-up) to add depth. \\ 
    8. **Texture/Material:** Textures or materials of the subject. Detail the tactile qualities to enhance realism. \\ 
    9. **Time Period:** Specific era. When relevant, specify a realistic time period to provide context. \\ 
    10. **Medium:** Artistic medium or level of detail. \\[5pt]
    
    - **Key Trigger Words:** Include terms like `high-resolution', 'highly realistic'. \\ 
    - **Length:** Keep the prompt under 75 tokens. \\ 
    - **Avoid:** Do not include any additional subjects in the prompt. Do not include any descriptions about the background. \\ \hline
  \end{tabular}
  \label{tab:llmsystempromptcomparison}
\end{table*}

\begin{table*}[h!]
  \centering
  \renewcommand{\arraystretch}{1.5} % Adjust row height
    \caption{\textbf{The initial system prompt of the data validation agent.} The category name is a placeholder for the specific object category.}
  \begin{tabular}{|p{0.96\linewidth}|}
    \hline
    \textbf{\normalsize The Data Validation Agent's Initial System Prompt} \\ \hline
    As an AI assistant, your role is to analyze images to determine their suitability based on specific criteria. First, provide a detailed description of the image. Second, evaluate the image against four criteria: 1. it should contain only one subject; 2. the subject should be shown from a single angle or perspective, without multiple views or angles within the same image; 3. the subject should be intact and fully visible; and 4. the background should be empty or plain, without distracting elements. Third, based on this evaluation, decide whether to filter out the image if it violates any of the criteria or keep it if it meets all of them. At last, conclude with a result stating "Keep" if the image meets all criteria or "Filter Out" if it violates any. Present your analysis in the specified output format, including the image description, detailed evaluations with explanations and results for each criterion, a conclusion, and the final result. \\

    Output Format: \\ 
    Image Description: \\ 

    Evaluation Criteria: \\ 

    1. Single [Category Name]: \\ 
    \hspace{1cm}- Explanation \\ 
    \hspace{1cm}- Result: Meet or Fail \\ 
    2. Single View: \\ 
    \hspace{1cm}- Explanation \\ 
    \hspace{1cm}- Result: Meet or Fail \\ 
    3. Intact [Category Name]: \\ 
    \hspace{1cm}- Explanation \\ 
    \hspace{1cm}- Result: Meet or Fail \\ 
    4. Plain Background: \\ 
    \hspace{1cm}- Explanation \\ 
    \hspace{1cm}- Result: Meet or Fail \\ 

    Conclusion: \\ 

    Result: Keep or Filter Out \\ \hline
  \end{tabular}  
  \label{tab:vllmsystempromptinitial}
\end{table*}

\begin{table*}[h!]
  \centering
  \renewcommand{\arraystretch}{1.5} % Adjust row height
    \caption{\textbf{The optimized system prompt of the data validation agent.} The category name is a placeholder for the specific object category.}
  \begin{tabular}{|p{0.96\linewidth}|}
    \hline
    \textbf{\normalsize The Data Validation Agent's Optimized System Prompt} \\ \hline
    You are an AI assistant that analyzes images to determine their suitability based on specific criteria. \\ 

    **Instructions:** \\ 

    1. **Describe the image in detail**. \\ 

    2. **Evaluate the image** against the following criteria: \\ 
    \hspace{1cm} - **Criteria 1 - Single subject**: The image should contain only one subject. \\ 
    \hspace{1cm} - **Criteria 2 - Single View**: The subject should be shown from a single angle or perspective. \\ 
    \hspace{1cm} - **Criteria 3 - Intact subject**: The subject should be intact and fully visible. \\ 
    \hspace{1cm} - **Criteria 4 - Plain Background**: The background should be empty or plain, without distracting elements. \\ 

    3. **Decide whether to filter out the image** based on these criteria. \\ 

    4. Conclude with **Result:** Keep if the image meets all criteria or **Result:** Filter Out if it violates any criteria. \\ 

    **Output Format:** \\ 
    **Image Description:** \\ 

    [Your detailed description here] \\ 

    **Evaluation Criteria:** \\ 

    1. **Single [Category Name]:** \\ 
    \hspace{1cm}* [Explanation] \\ 
    \hspace{1cm}* **Result:** [Meet/Fail] \\ 
    2. **Single View:** \\ 
    \hspace{1cm}* [Explanation] \\ 
    \hspace{1cm}* **Result:** [Meet/Fail] \\ 
    3. **Intact [Category Name]:** \\ 
    \hspace{1cm}* [Explanation] \\ 
    \hspace{1cm}* **Result:** [Meet/Fail] \\ 
    4. **Plain Background:** \\ 
    \hspace{1cm}* [Explanation] \\ 
    \hspace{1cm}* **Result:** [Meet/Fail] \\ 

    **Conclusion:** \\ 

    [Your conclusion here] \\ 

    **Result:** [Keep/Filter Out] \\ \hline
  \end{tabular}
  \label{tab:vllmsystempromptoptimized}
\end{table*}

\clearpage

\newcommand\LDTextBoxWidth{.7\textwidth}
\newcommand\LDImageBoxWidth{3.2cm}
\newcommand\LDImageWidth{3.2cm}
\begin{figure*}[ht!]
  \centering
  \renewcommand{\arraystretch}{1.5} % Adjust row height
  \begin{tabular}{|p{.7\linewidth}|c|}
    \hline
    \textbf{Standard Layer Diffusion Prompt} & \textbf{Foreground Instance Image} \\ \hline

    \parbox[c]{\LDTextBoxWidth}{
      An image of a single airplane, an aircraft that has a fixed wing and is powered by propellers or jets.
    } & 
    \parbox[c]{\LDImageBoxWidth}{
      \includegraphics[width=\LDImageWidth]{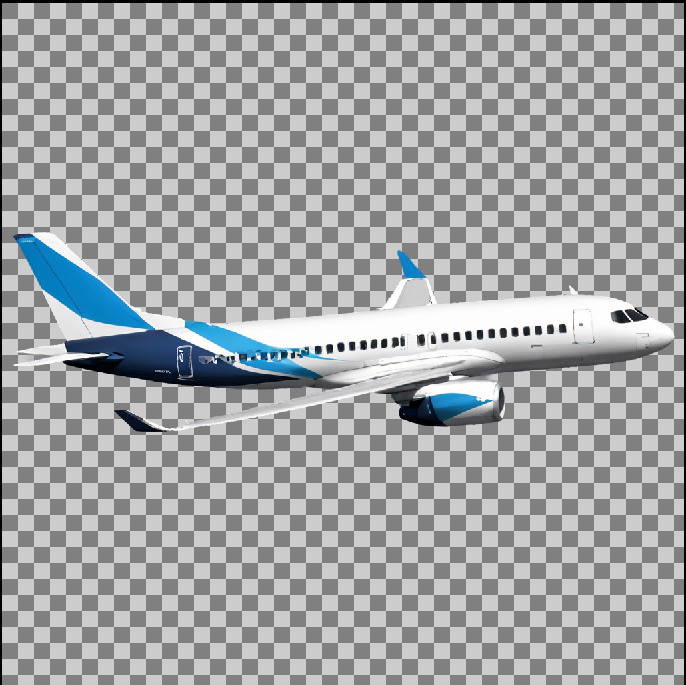}
    } \\ \hline \hline

    \textbf{Optimized Layer Diffusion Prompts} & \textbf{Foreground Instance Image} \\ \hline

    \parbox[c]{\LDTextBoxWidth}{
      High-resolution digital rendering of a single airplane, alone, with a diverse range of metallic shades from a gleaming silver to a weathered copper, ultra-realistic in style, conveying a sense of nostalgia and longing, under soft morning sunlight with warm shadows, viewed from a low-angle perspective that accentuates the aircraft's sleek lines, featuring intricate textures of polished aluminum and worn leather seats, set against a timeless backdrop of the early 20th century. 
    } & 
    \parbox[c]{\LDImageBoxWidth}{
    \includegraphics[width=\LDImageWidth]{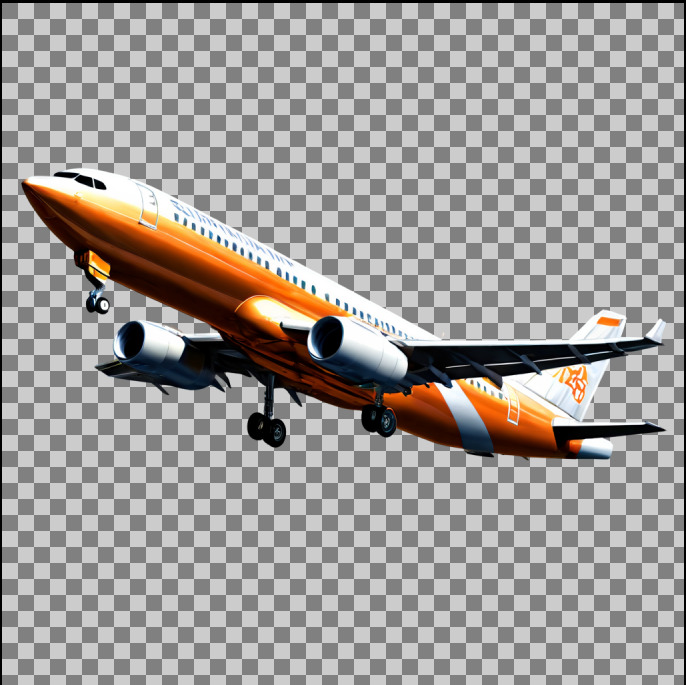} 
    } \\ \hline

    \parbox[c]{\LDTextBoxWidth}{
      High-resolution digital rendering of a single, sleek, silver, Boeing 747-400F airplane, alone, in a majestic, photorealistic style, evoking feelings of serenity and power, with soft, warm, golden sunlight casting long shadows across its metallic surface, from a low, 45-degree angle, showcasing the intricate details of its textured, riveted aluminum skin, during the golden hour of a late summer evening.
    } & 
    \parbox[c]{\LDImageBoxWidth}{
    \includegraphics[width=\LDImageWidth]{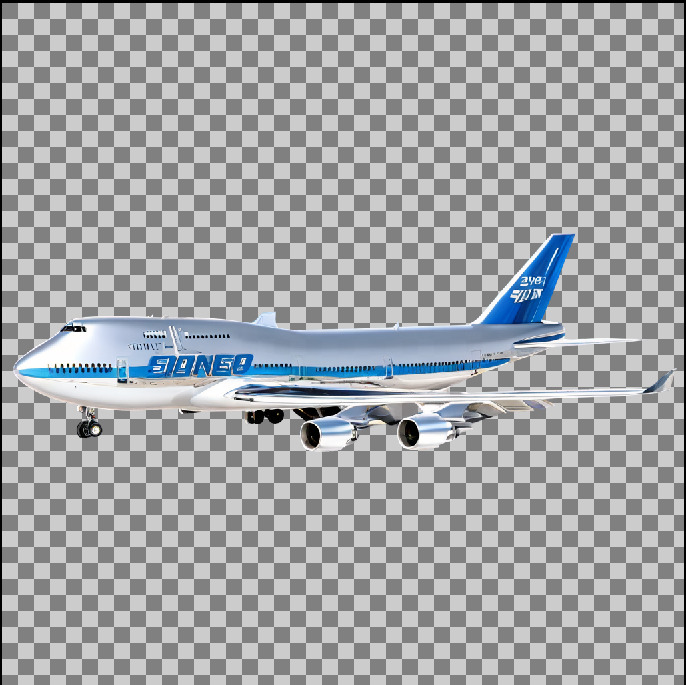}
    } \\ \hline

    \parbox[c]{\LDTextBoxWidth}{
      A high-resolution digital rendering of a single, vintage, metallic silver 1940s Boeing B-17 Flying Fortress aeroplane, alone, with a warm golden glow of sunlight illuminating its weathered aluminum skin, ultra-realistic details of its propellers and fuselage visible, a mix of nostalgic and eerie atmosphere, a dramatic low-angle shot from directly below, with intricate textures and rivets on its surface visible, and a subtle sense of motion as if in mid-flight.
      } & 
    \parbox[c]{\LDImageBoxWidth}{
    \includegraphics[width=\LDImageWidth]{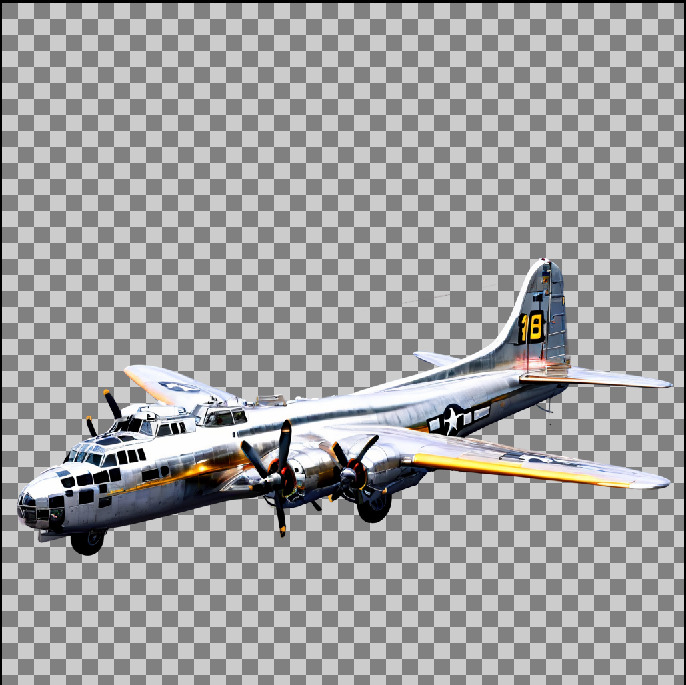}
    } \\ \hline

    \parbox[c]{\LDTextBoxWidth}{
      High-resolution digital rendering of a single, sleek, and aerodynamic airplane alone, with a metallic silver fuselage, a deep blue tail fin, and bright red propellers, rendered in an ultra-realistic style with a sense of nostalgia and excitement, under soft golden morning sunlight, from a low-angle, bird's-eye view, showcasing the intricate textures of the aircraft's aluminum skin and the worn, leather-like seats, set against a timeless era of classic aviation.
    } & 
    \parbox[c]{\LDImageBoxWidth}{
    \includegraphics[width=\LDImageWidth]{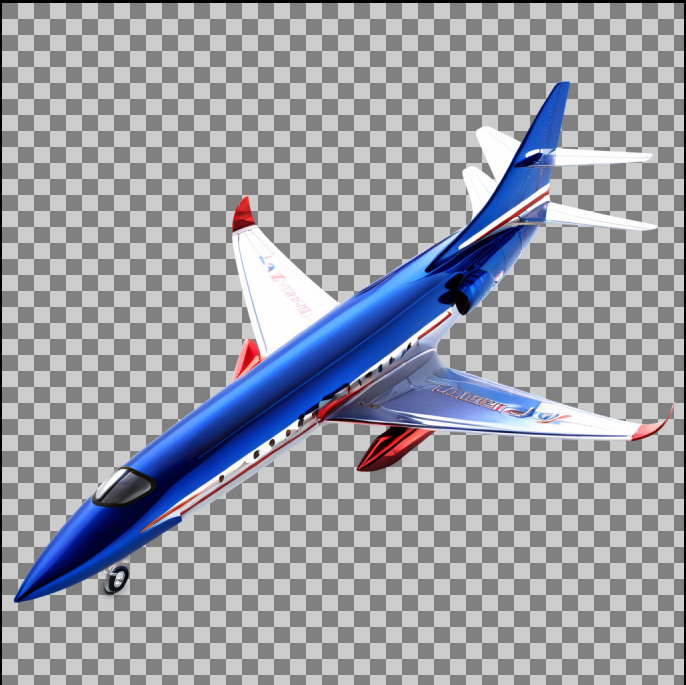}
    } \\ \hline

    \parbox[c]{\LDTextBoxWidth}{
      A highly realistic digital rendering of a single vintage airplane, an aeroplane, alone, with a metallic silver body, dark blue accents, and a red stripe, sitting on a runway, exuding a sense of nostalgia, under warm golden sunset lighting, from a low-angle perspective, showcasing its worn aluminum skin and rusty propellers, during the 1940s.
    } & 
    \parbox[c]{\LDImageBoxWidth}{
    \includegraphics[width=\LDImageWidth]{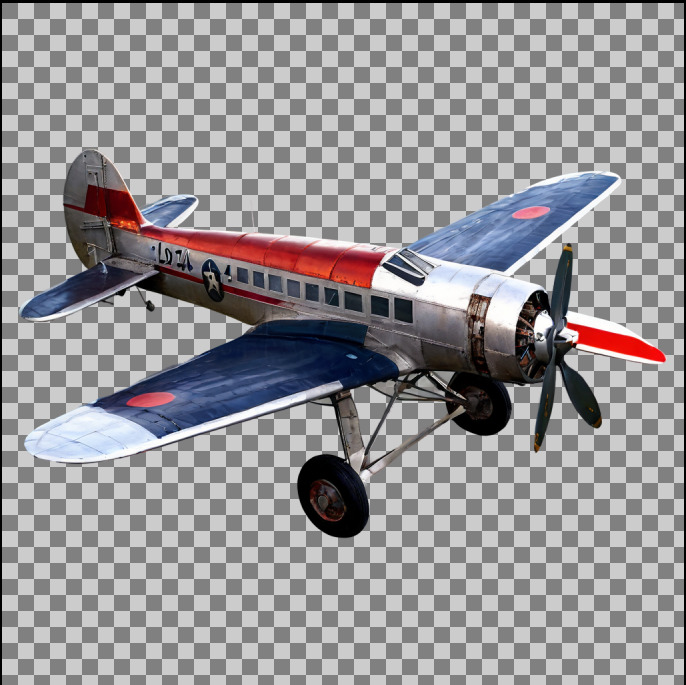}
    } \\ \hline

  \end{tabular}
  \caption{\textbf{Comparison of the standard and optimized LD prompts and their corresponding foreground instances for airplane.}}
  \label{fig:ldpromptcomparisonairplane}
\end{figure*}

\begin{figure*}[ht!]
  \centering
  \renewcommand{\arraystretch}{1.5} % Adjust row height
  \begin{tabular}{|p{.7\linewidth}|c|}
    \hline
    \textbf{Standard Layer Diffusion Prompt} & \textbf{Foreground Instance Image} \\ \hline

    \parbox[c]{\LDTextBoxWidth}{
      An image of a single orange (fruit), orange (FRUIT of an orange tree).
    } & 
    \parbox[c]{\LDImageBoxWidth}{
    \includegraphics[width=\LDImageWidth]{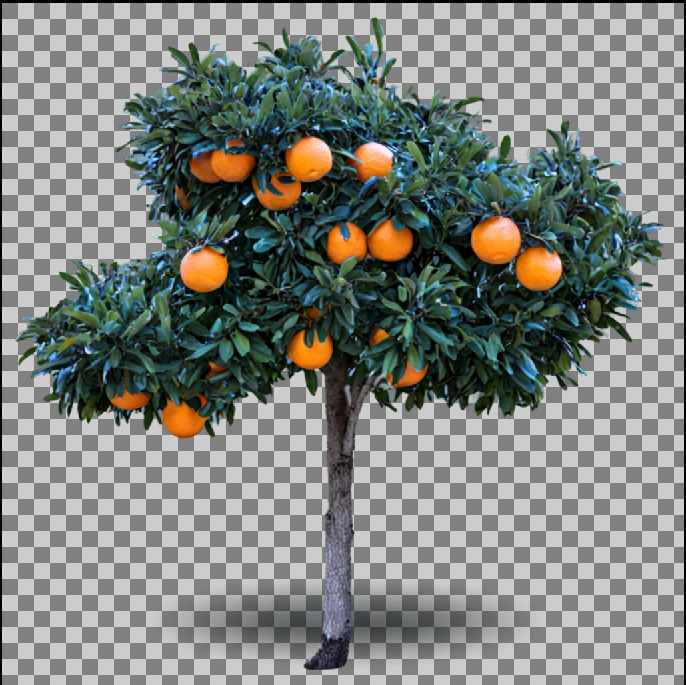}
    } \\ \hline \hline

    \textbf{Optimized Layer Diffusion Prompts} & \textbf{Foreground Instance Image} \\ \hline

    \parbox[c]{\LDTextBoxWidth}{
      High-resolution digital rendering of a single orange fruit alone, with vibrant orange hues ranging from deep burnt orange to bright juicy orange, ultra-realistic, conveying a sense of serenity and satisfaction, under soft warm natural light with subtle shadows, viewed from a 45-degree angle with a slight macro perspective, featuring a glossy skin with subtle ridges and a slightly dimpled texture, set in a timeless, nostalgic atmosphere evoking memories of summertime.
    } & 
    \parbox[c]{\LDImageBoxWidth}{
    \includegraphics[width=\LDImageWidth]{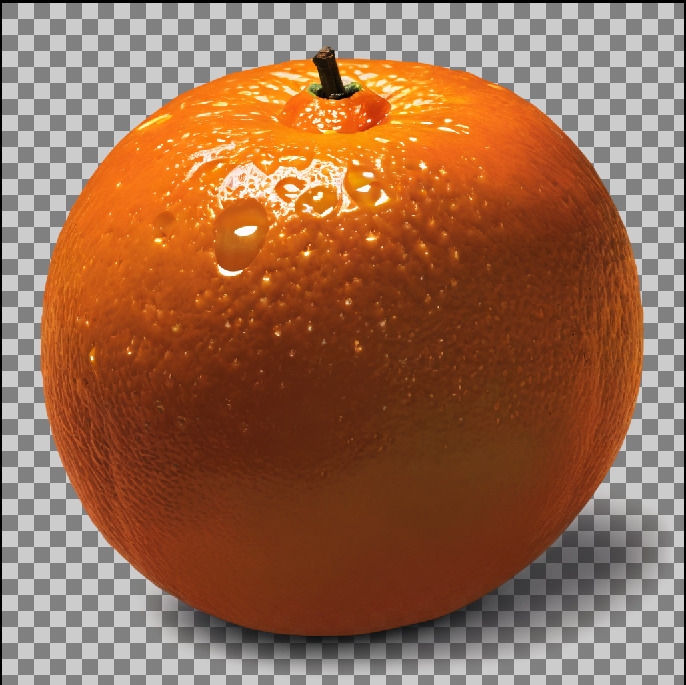}
    } \\ \hline

    \parbox[c]{\LDTextBoxWidth}{
      High-resolution digital rendering of a single, perfectly ripe, vibrant orange fruit alone, with a warm, inviting orange color gradating from a deep burnt orange shade at the stem to a bright, juicy orange hue near the peel, ultra-realistic in style, conveying a sense of nostalgia and warmth, under soft, golden natural lighting, from a 45-degree angle, with a subtle sheen and slight oiliness to the peel, as if freshly picked from an orange tree in a lush, Mediterranean orchard during the peak summer season.
    } & 
    \parbox[c]{\LDImageBoxWidth}{
    \includegraphics[width=\LDImageWidth]{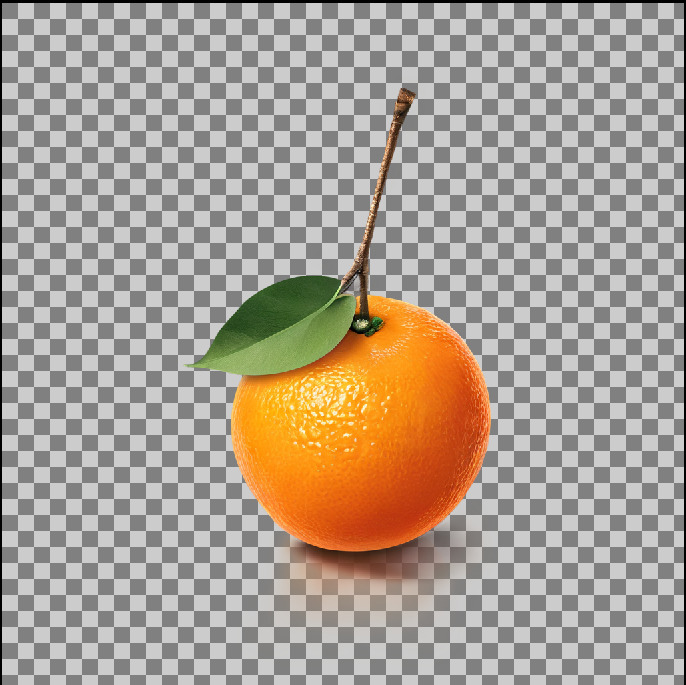}
    } \\ \hline

    \parbox[c]{\LDTextBoxWidth}{
      High-resolution digital rendering of a single orange fruit alone, with vibrant shades of orange, coral, and golden hues, ultra-realistic in style, conveying a sense of freshness, serenity, and ripeness, illuminated by soft, warm sunlight, from a 45-degree angle, showcasing the intricate texture of its slightly bumpy skin and the subtle sheen of its juicy pulp, in a timeless, modern setting.
    } & 
    \parbox[c]{\LDImageBoxWidth}{
    \includegraphics[width=\LDImageWidth]{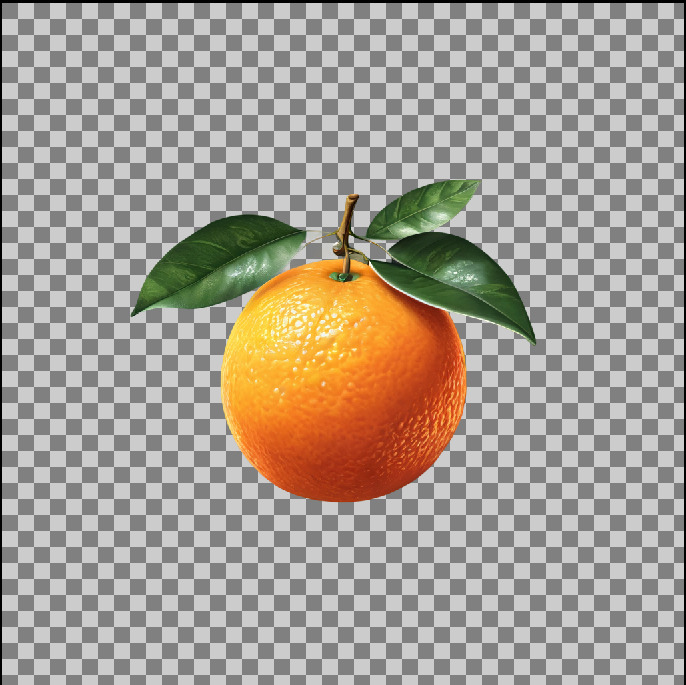}
    } \\ \hline

    \parbox[c]{\LDTextBoxWidth}{
      High-resolution digital rendering of a single, juicy orange fruit, alone, with vibrant orange hues, deep orange-red undertones, and subtle yellow-green highlights, in an ultra-realistic style, conveying a sense of ripeness, freshness, and satisfaction, under warm, soft, golden natural lighting, viewed from a 45-degree angle with a shallow depth of field, showcasing the intricate texture of the fruit's skin, which is slightly wrinkled and slightly sticky to the touch, as if plucked from a tree in the Mediterranean during the peak summer season.
    } & 
    \parbox[c]{\LDImageBoxWidth}{
    \includegraphics[width=\LDImageWidth]{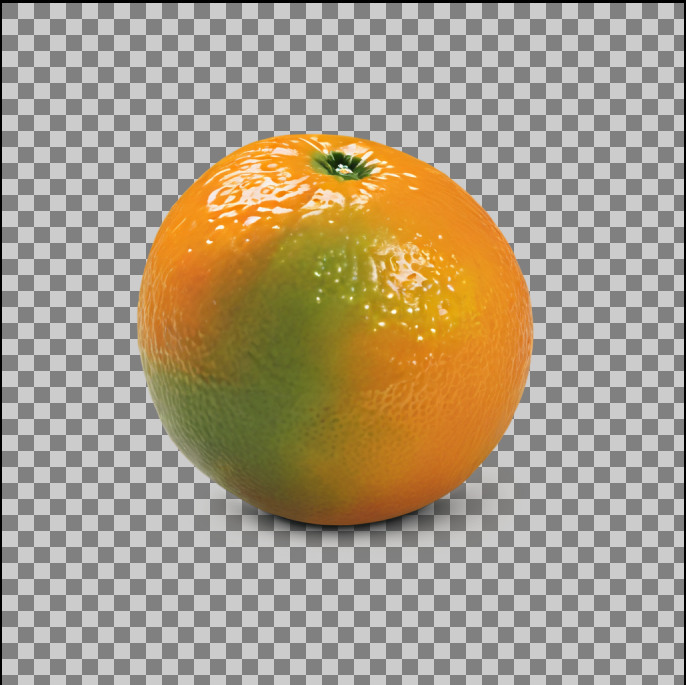}
    } \\ \hline

    \parbox[c]{\LDTextBoxWidth}{
      High-resolution digital rendering of a single vibrant orange fruit, alone on a surface, radiating warm golden hues with deep orange undertones, ultra-realistic in style, evoking feelings of nostalgia and abundance, bathed in soft warm sunlight with subtle shadows, captured from a 45-degree angle with the fruit slightly rotated, showcasing its intricate texture of fine oil glands and a slight sheen from a gentle mist, set in a timeless era of rustic simplicity.
    } & 
    \parbox[c]{\LDImageBoxWidth}{
    \includegraphics[width=\LDImageWidth]{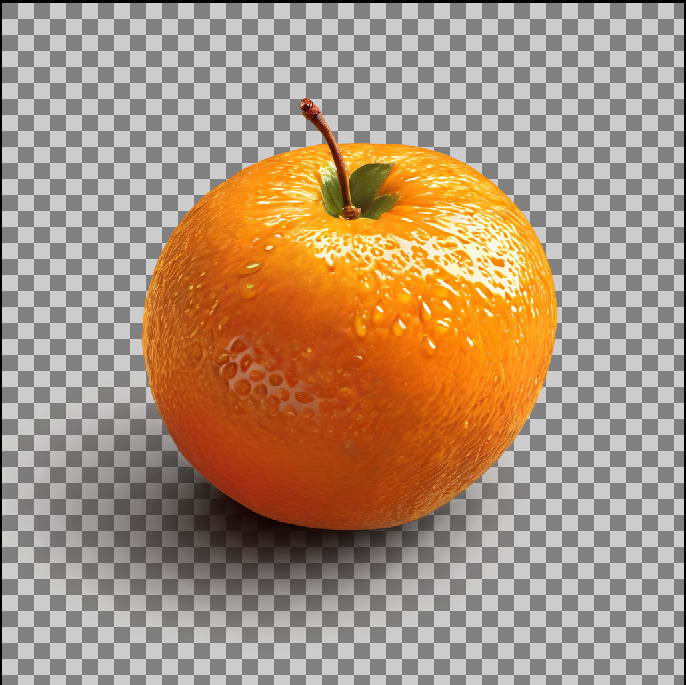}
    } \\ \hline

  \end{tabular}
  \caption{\textbf{Comparison of the standard and optimized LD prompts and their corresponding foreground instances for orange.}}
  \label{fig:ldpromptcomparisonorange}
\end{figure*}

\begin{figure*}[ht!]
  \centering
  \renewcommand{\arraystretch}{1.5} % Adjust row height
  \begin{tabular}{|p{.7\linewidth}|c|}
    \hline
    \textbf{Standard Layer Diffusion Prompt} & \textbf{Foreground Instance Image} \\ \hline

    \parbox[c]{\LDTextBoxWidth}{
      An image of a single car, a motor vehicle with four wheels.
    } & 
    \parbox[c]{\LDImageBoxWidth}{
    \includegraphics[width=\LDImageWidth]{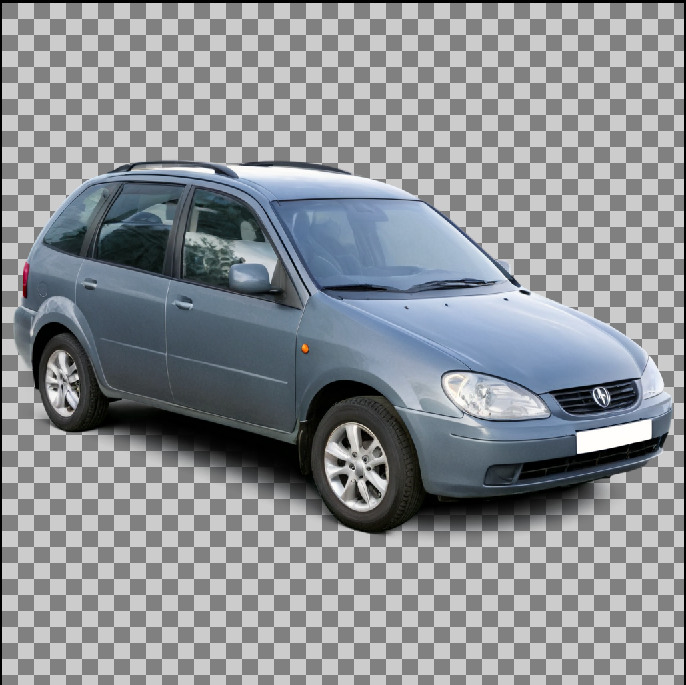}
    } \\ \hline \hline

    \textbf{Optimized Layer Diffusion Prompts} & \textbf{Foreground Instance Image} \\ \hline

    \parbox[c]{\LDTextBoxWidth}{
      High-resolution digital rendering of a single, ultra-realistic car standing alone on a dimly lit city street at sunset, its sleek metallic body glinting with a gradient of deep blues and rich silvers, its aerodynamic curves accentuated by warm golden light spilling from the setting sun, its tires appearing worn and weathered with a tactile texture, its surface reflecting a mesmerizing array of colors and shades, from the deep, rich tones of its metallic paint to the subtle, nuanced hues of its tinted windows, all captured from a dramatic low-angle perspective that emphasizes the car's powerful, aggressive stance.
    } & 
    \parbox[c]{\LDImageBoxWidth}{
    \includegraphics[width=\LDImageWidth]{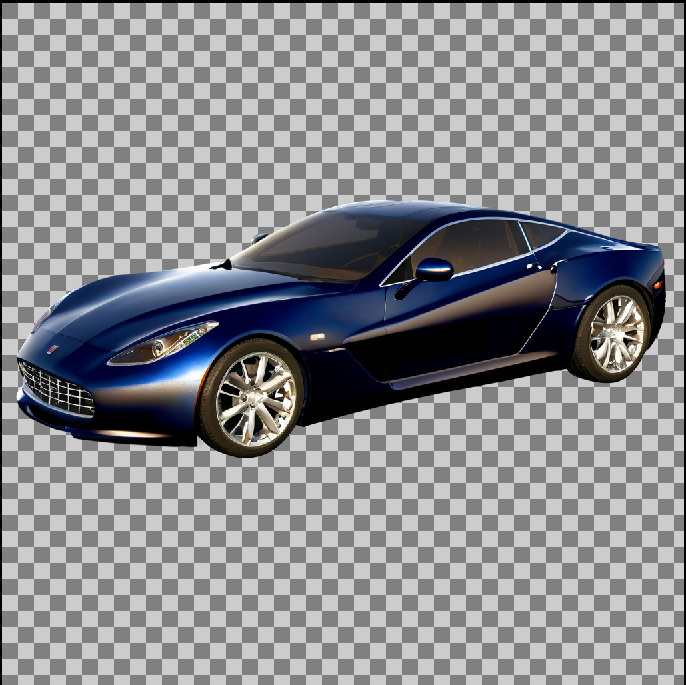}
    } \\ \hline

    \parbox[c]{\LDTextBoxWidth}{
      Highly realistic digital rendering of a single, sleek, 1969 Chevrolet Camaro SS alone, featuring a bold, metallic red paint job with a deep, glossy finish, a dark, matte black hood and roof, and a bright, chrome exhaust tip, captured in a moody, atmospheric scene with a warm, golden sunlight illuminating the car from a low, angled perspective, emphasizing the curved lines and aggressive stance of the vehicle, with a soft, velvety texture on the leather interior and a rough, industrial texture on the exposed engine components, set against a timeless, nostalgic backdrop.
    } & 
    \parbox[c]{\LDImageBoxWidth}{
    \includegraphics[width=\LDImageWidth]{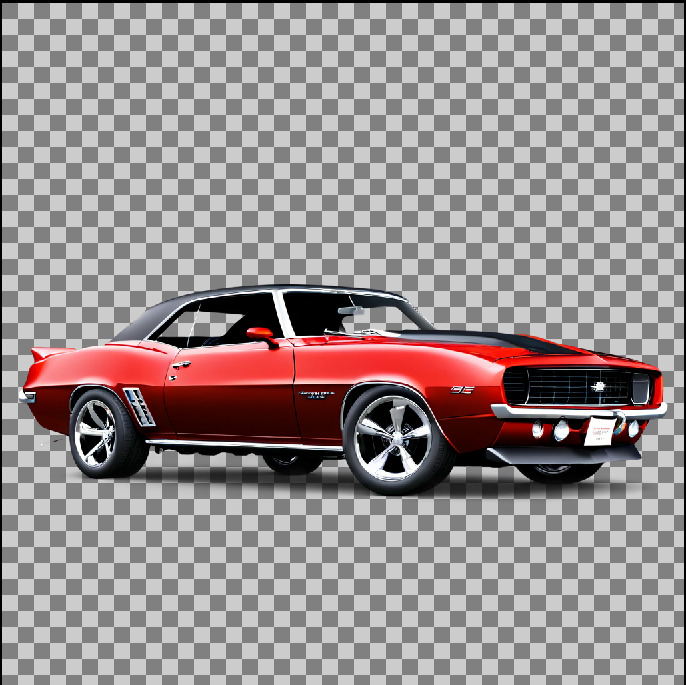}
    } \\ \hline

    \parbox[c]{\LDTextBoxWidth}{
      High-resolution digital rendering of a single sleek, high-performance sports car, alone, with a glossy metallic blue finish featuring hints of navy and turquoise, an ultra-realistic style, an eerie and mysterious mood, dramatic side lighting with deep shadows, a low-angle, dramatic perspective, a smooth and aerodynamic texture, and a contemporary, modern time period.
    } & 
    \parbox[c]{\LDImageBoxWidth}{
    \includegraphics[width=\LDImageWidth]{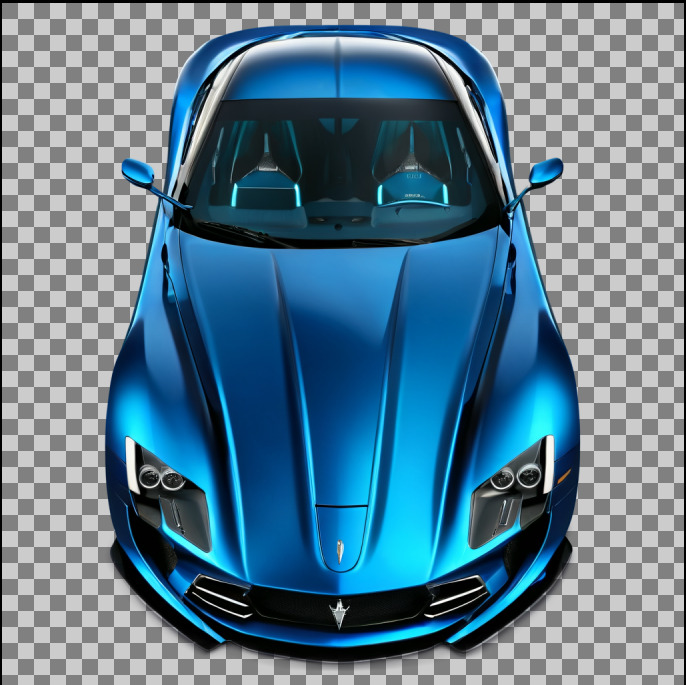}
    } \\ \hline

    \parbox[c]{\LDTextBoxWidth}{
      High-resolution digital rendering of a single, sleek, high-performance, 2023, Lamborghini Aventador, alone, with a predominantly glossy, matte black, and metallic silver body, ultra-realistic, conveying a sense of speed and power, with dramatic, golden hour lighting casting a warm glow on its chiseled lines, from a low, eye-level, 45-degree angle, showcasing the intricate, hand-stitched, black and silver leather interior, and the smooth, rubberized, textured steering wheel.
    } & 
    \parbox[c]{\LDImageBoxWidth}{
    \includegraphics[width=\LDImageWidth]{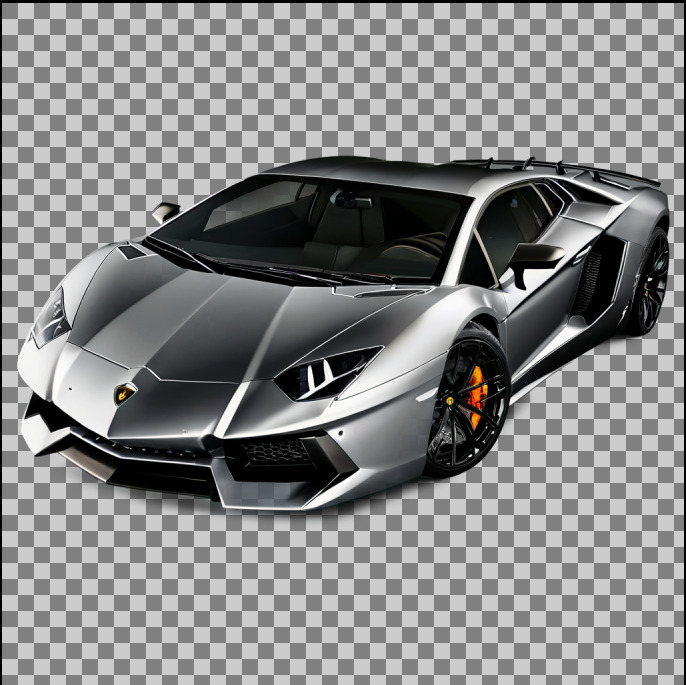}
    } \\ \hline

    \parbox[c]{\LDTextBoxWidth}{
      High-resolution digital rendering of a single, ultra-realistic, sleek, 1969 cherry-red Ferrari 250 GTO, alone, with a mix of glossy and matte black leather interior, racing stripes, and gleaming chrome accents, exuding a sense of speed and luxury, under a warm, golden sunlight, from a low, eye-level perspective, showcasing its textured, hand-stitched leather seats and intricate dashboard details, set in a nostalgic, vintage era.
    } & 
    \parbox[c]{\LDImageBoxWidth}{
    \includegraphics[width=\LDImageWidth]{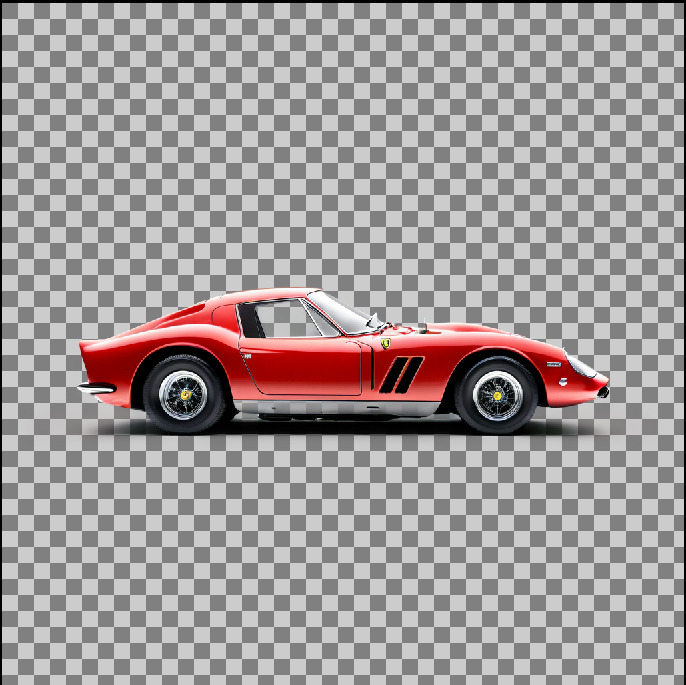}
    } \\ \hline

  \end{tabular}
  \caption{\textbf{Comparison of the standard and optimized LD prompts and their corresponding foreground instances for car.}}
  \label{fig:ldpromptcomparisoncar}
\end{figure*}

\begin{figure*}[ht!]
  \centering
  \renewcommand{\arraystretch}{1.5} % Adjust row height
  \begin{tabular}{|p{.7\linewidth}|c|}
    \hline
    \textbf{Standard Layer Diffusion Prompt} & \textbf{Foreground Instance Image} \\ \hline

    \parbox[c]{\LDTextBoxWidth}{
      An image of a single person, a human being.
    } & 
    \parbox[c]{\LDImageBoxWidth}{
    \includegraphics[width=\LDImageWidth]{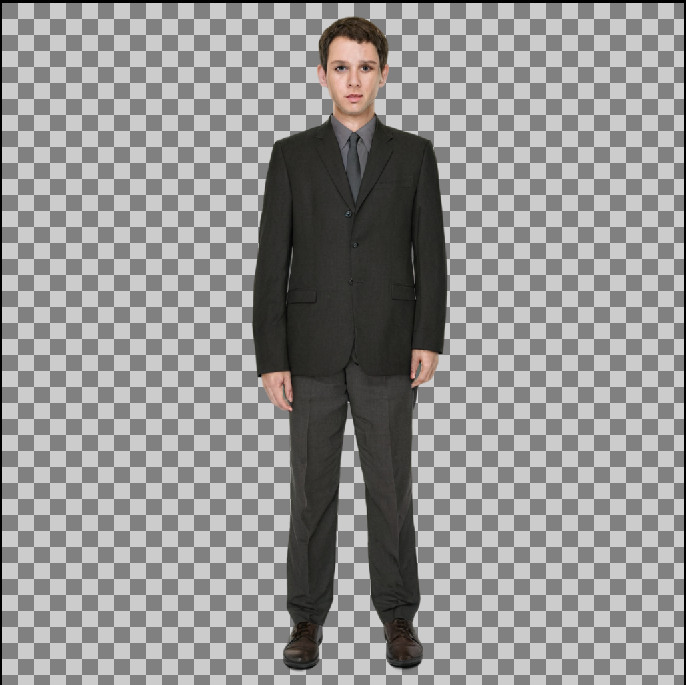}
    } \\ \hline \hline

    \textbf{Optimized Layer Diffusion Prompts} & \textbf{Foreground Instance Image} \\ \hline

    \parbox[c]{\LDTextBoxWidth}{
      High-resolution digital rendering of a single person alone, donning a vibrant turquoise shirt with a slight sheen, a pair of distressed brown jeans, and a worn black leather jacket, captured in an ultra-realistic style that conveys a sense of melancholic introspection under soft, warm golden hour sunlight, viewed from a dynamic low-angle perspective that accentuates the subject's angular features, showcasing a mix of smooth skin and subtle facial hair texture, set in a timeless era that blends modern and vintage elements.
    } & 
    \parbox[c]{\LDImageBoxWidth}{
    \includegraphics[width=\LDImageWidth]{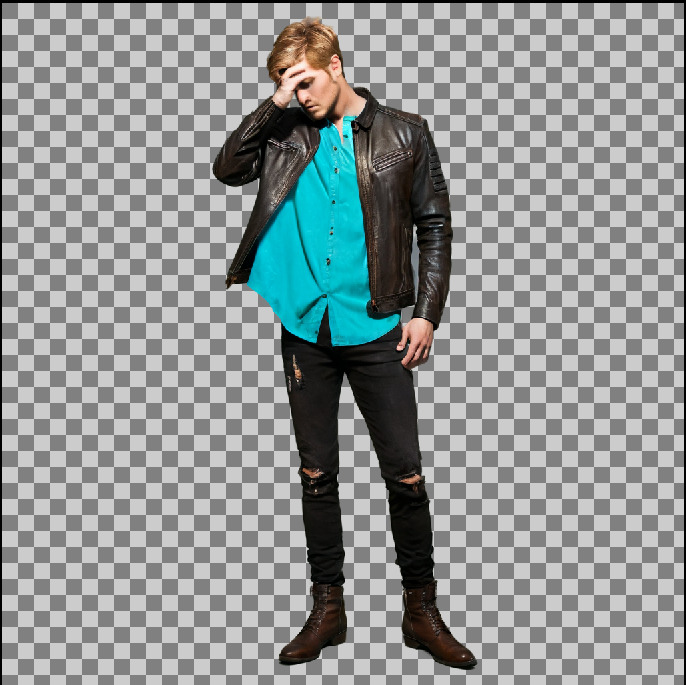}
    } \\ \hline

    \parbox[c]{\LDTextBoxWidth}{
      High-resolution digital rendering of a single person alone, dressed in a vibrant, high-collared, emerald-green coat with intricate, golden-brown buttons, paired with a crisp, snow-white shirt, rendered in ultra-realistic style, conveying a sense of serene contemplation, melancholic introspection, and quiet determination, under soft, warm, golden-hour sunlight that casts a gentle, diffused glow across their features, from a low, eye-level perspective that emphasizes their introspective expression, with a subtle, velvety texture to their skin and a luxurious, smooth sheen to their coat.
    } & 
    \parbox[c]{\LDImageBoxWidth}{
    \includegraphics[width=\LDImageWidth]{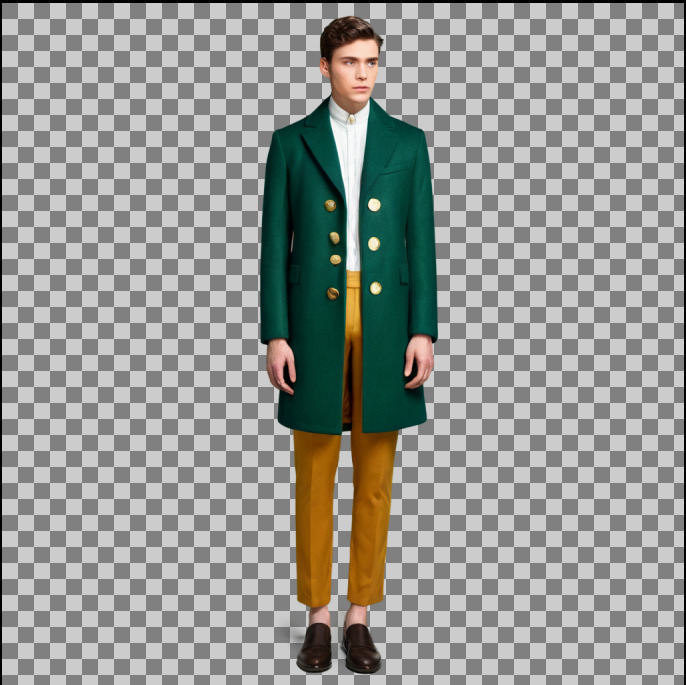}
    } \\ \hline

    \parbox[c]{\LDTextBoxWidth}{
      High-resolution digital rendering of a single person alone, wearing a bright yellow sundress with golden accents and a subtle floral pattern, standing in a quiet alleyway with soft warm sunlight filtering through the trees, cast in an ultra-realistic style with intricate details, conveying a mix of confidence and vulnerability, with the light dancing across their features and casting a warm glow on their skin, viewed from a low-angle perspective that accentuates their tall stature, with a smooth and silky texture to their dress and a subtle sheen to their hair, set in a modern contemporary era.
    } & 
    \parbox[c]{\LDImageBoxWidth}{
    \includegraphics[width=\LDImageWidth]{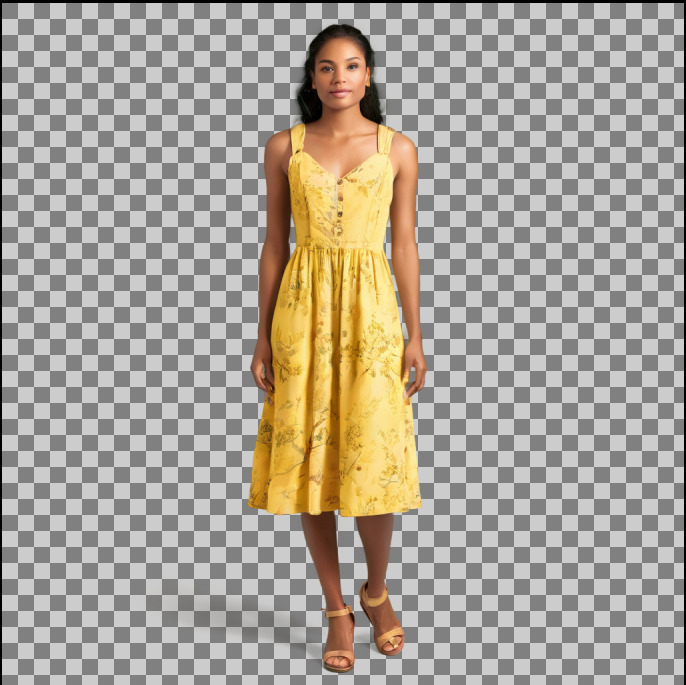}
    } \\ \hline

    \parbox[c]{\LDTextBoxWidth}{
      High-resolution digital rendering of a single person alone, dressed in a vibrant turquoise and golden outfit with intricate, hand-beaded patterns, ultra-realistic style, conveying a mix of serenity and confidence, illuminated by soft, warm sunlight and dramatic, moody shadows, captured from a low-angle, dynamic perspective, with a focus on the intricate texture of their ornate, beaded necklace and the soft, smooth skin of their face, set in a contemporary, modern time period.
    } & 
    \parbox[c]{\LDImageBoxWidth}{
    \includegraphics[width=\LDImageWidth]{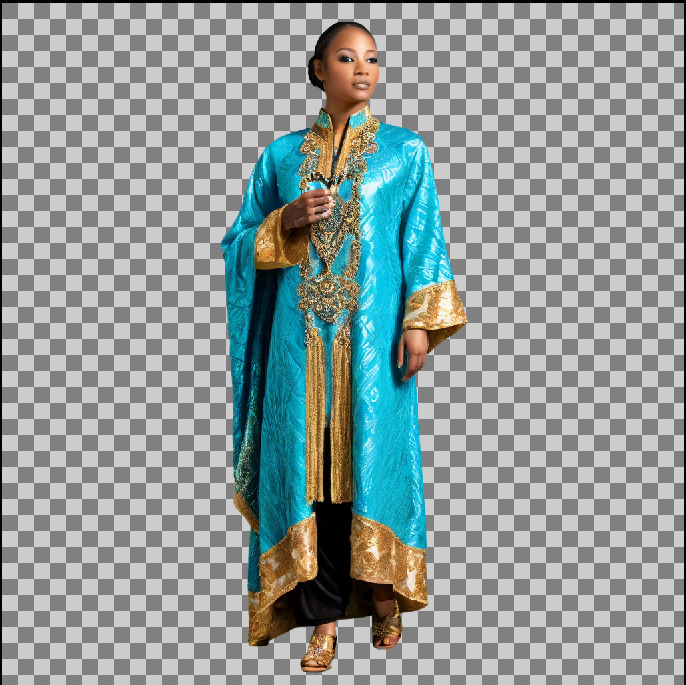}
    } \\ \hline

    \parbox[c]{\LDTextBoxWidth}{
      High-resolution digital rendering of a single person alone, dressed in a vibrant turquoise and golden embroidered traditional Indian outfit, ultra-realistic, conveying a mix of serenity and introspection, softly illuminated by warm morning sunlight, captured from a low-angle perspective, showcasing the intricate texture of their silk sari and the delicate pattern on their intricately crafted silver jewelry, set in a timeless and nostalgic era.
    } & 
    \parbox[c]{\LDImageBoxWidth}{
    \includegraphics[width=\LDImageWidth]{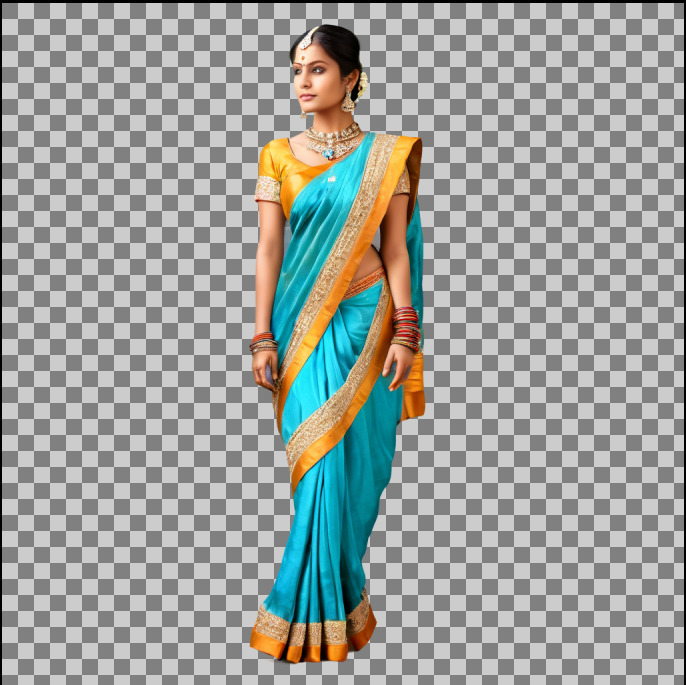}
    } \\ \hline

  \end{tabular}
  \caption{\textbf{Comparison of the standard and optimized LD prompts and their corresponding foreground instances for person.}}
  \label{fig:ldpromptcomparisonperson}
\end{figure*}

%%%%%%%%%%%%%%%%%%%%%%%%%%%%%%%%%%%%%%%%%%%%%%%%%%%%%%%%%%%%%%%%%%%%%%%%%%%%%%%%%%%%%%%%

\begin{figure*}[ht!]
  \centering
  \renewcommand{\arraystretch}{1.5} % Adjust row height

  \begin{tabular}{|p{.7\linewidth}|c|}
    \hline
    \textbf{Optimized Layer Diffusion Prompt} & \textbf{Foreground Instance Image} \\ \hline

    \parbox[c]{\LDTextBoxWidth}{
      A highly realistic, high-resolution digital rendering of a single, juicy, vibrant orange (fruit) of an orange tree, alone on a smooth, glossy surface, with warm golden undertones and deep reddish-orange hues, ultra-realistic in style, conveying feelings of abundance and gratitude, under soft, diffused morning sunlight, from a 45-degree angle, with intricate, slightly wrinkled skin and a few tiny, raised oil glands, evoking a sense of tactile experience, set in a timeless, modern era.
    } & 
    \parbox[c]{\LDImageBoxWidth}{
    \includegraphics[width=\LDImageWidth]{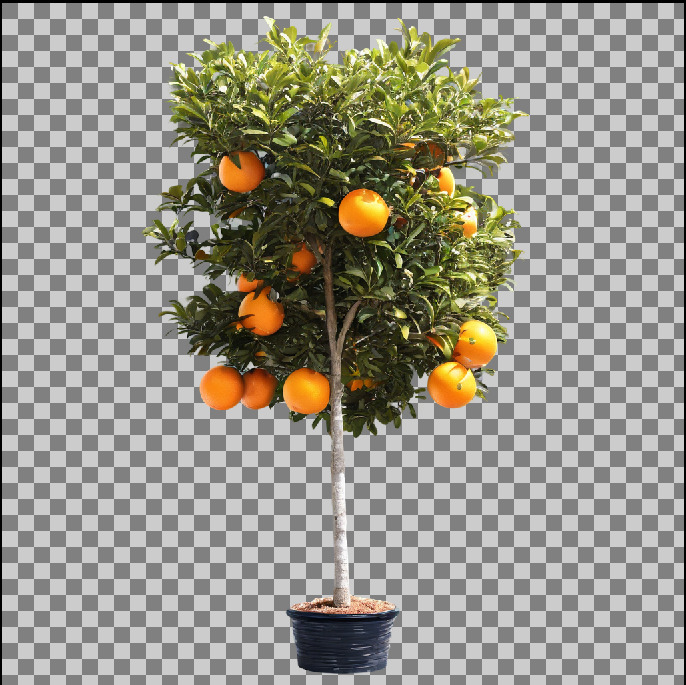}
    } \\ \hline
  \end{tabular}

  \begin{tabular}{|p{0.9625\linewidth}|}
    \hline
    \textbf{The Data Validation Agent's Response} \\ \hline
    **Image Description:** \\

    The image depicts a potted tree with green leaves and multiple oranges hanging from its branches. The tree is positioned centrally in the frame, with its trunk visible at the bottom. The oranges are a vibrant orange color, and the leaves are a deep green. The tree appears to be healthy and well-maintained, with no visible signs of damage or disease. \\
    
    **Evaluation Criteria:** \\
    
    1. **Single orange:** \\
    \hspace{1cm}* The image contains multiple oranges, not just one. \\
    \hspace{1cm}* **Result:** Fail \\
    2. **Single View:** \\
    \hspace{1cm}* The image shows the tree from a single angle, with no multiple views or perspectives. \\
    \hspace{1cm}* **Result:** Meet \\
    3. **Intact orange:** \\
    \hspace{1cm}* All the oranges in the image appear to be intact and fully visible. \\
    \hspace{1cm}* **Result:** Meet \\
    4. **Plain Background:** \\
    \hspace{1cm}* The background of the image is solid black, which is considered plain. \\
    \hspace{1cm}* **Result:** Meet \\
    
    **Conclusion:** \\ 
    
    Based on the evaluation criteria, the image fails to meet the first criterion, which requires only one orange. However, it meets the remaining three criteria, including showing the tree from a single view, having intact oranges, and having a plain background. Therefore, the image does not meet all the criteria and should be filtered out. \\ 
    
    **Result:** Filter Out \\

    \hline

  \end{tabular}
  \caption{\textbf{Example of the data validation agent filtering for orange.}}
  \label{fig:vllmfilteringorange}
\end{figure*}

\begin{figure*}[ht!]
  \centering
  \renewcommand{\arraystretch}{1.5} % Adjust row height

  \begin{tabular}{|p{.7\linewidth}|c|}
    \hline
    \textbf{Optimized Layer Diffusion Prompt} & \textbf{Foreground Instance Image} \\ \hline

    \parbox[c]{\LDTextBoxWidth}{
      High-resolution digital rendering of a single alarm clock alone, in a mesmerizing gradient of warm golden hues and soft pastel blues, ultra-realistic in style, conveying a sense of serene tranquility and anxious anticipation, illuminated by the soft glow of a bedside lamp and the faint morning light seeping through the curtains, captured from a low-angle perspective that emphasizes the clock's sleek, metallic texture and intricate mechanisms, with a subtle sheen of polished stainless steel and a delicate scratch on its surface, set against the timeless backdrop of a bygone era.
    } & 
    \parbox[c]{\LDImageBoxWidth}{
    \includegraphics[width=\LDImageWidth]{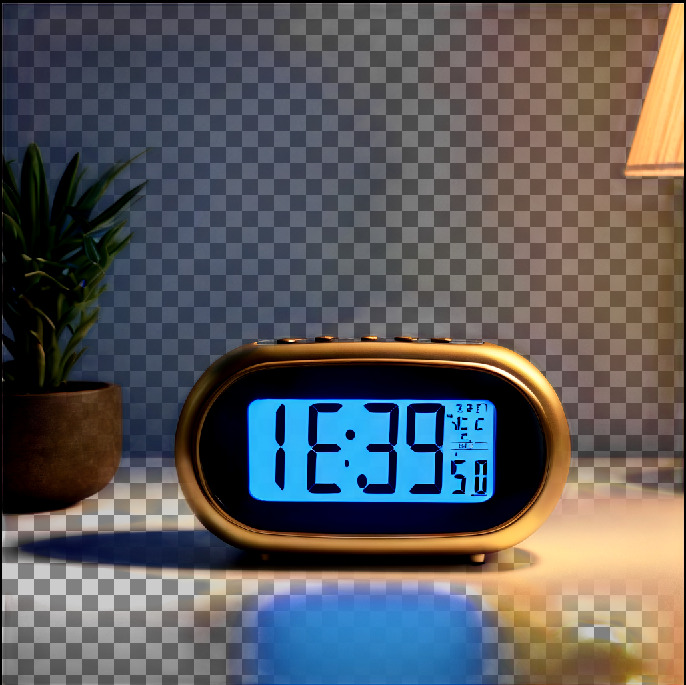}
    } \\ \hline
  \end{tabular}

  \begin{tabular}{|p{0.9625\linewidth}|}
    \hline
    \textbf{The Data Validation Agent's Response} \\ \hline
    **Image Description:** \\

    The image depicts a digital clock with a gold border and blue screen, displaying the time "16:39" in large black numbers. The clock is positioned on a table or desk, accompanied by a potted plant to its left and a lamp in the top-right corner. \\

    **Evaluation Criteria:** \\

    1. **Single clock:** \\
    \hspace{1cm}* The image contains only one clock. \\
    \hspace{1cm}* **Result:** Meet \\
    2. **Single View:** \\
    \hspace{1cm}* The clock is shown from a single angle or perspective. \\
    \hspace{1cm}* **Result:** Meet \\
    3. **Intact clock:** \\
    \hspace{1cm}* The clock is intact and fully visible. \\
    \hspace{1cm}* **Result:** Meet \\
    4. **Plain Background:** \\
    \hspace{1cm}* The background is not entirely plain, as there are other objects present (the plant and lamp). \\
    \hspace{1cm}* **Result:** Fail \\

    **Conclusion:** \\
    The image meets most of the evaluation criteria, but fails to have a plain background due to the presence of additional objects. Therefore, the final answer is: \\

    **Result:** Filter Out \\

    \hline

  \end{tabular}
  \caption{\textbf{Example of the data validation agent filtering for clock.}}
  \label{fig:vllmfilteringalarmclock}
\end{figure*}

\begin{figure*}[ht!]
  \centering
  \renewcommand{\arraystretch}{1.5} % Adjust row height

  \begin{tabular}{|p{.7\linewidth}|c|}
    \hline
    \textbf{Optimized Layer Diffusion Prompt} & \textbf{Foreground Instance Image} \\ \hline

    \parbox[c]{\LDTextBoxWidth}{
      High-resolution digital rendering of a single birthday card alone, adorned with vibrant shades of cerulean, golden, and crimson hues, ultra-realistically crafted in exquisite detail, evoking a mix of joy, nostalgia, and warmth, illuminated by soft, diffused natural light, and by the soft glow of a single candle, captured from a close-up, eye-level perspective, featuring a textured, embossed, and glossy surface, reminiscent of a vintage card from the early 20th century.
    } & 
    \parbox[c]{\LDImageBoxWidth}{
    \includegraphics[width=\LDImageWidth]{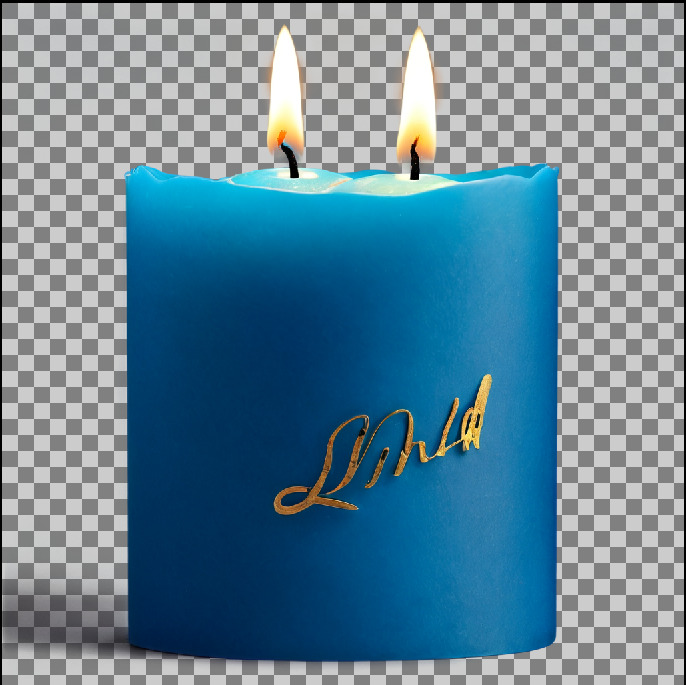}
    } \\ \hline
  \end{tabular}

  \begin{tabular}{|p{0.9625\linewidth}|}
    \hline
    \textbf{The Data Validation Agent's Response} \\ \hline

    **Image Description:** \\

    The image depicts a large blue candle with two lit wicks, set against a solid black background. The candle is cylindrical in shape and features a wavy top edge. It has a bright blue color that gradually darkens towards the bottom, where it meets the black background. The candle's surface appears smooth, with no visible imperfections or markings. \\

    **Evaluation Criteria:** \\

    1. **Single birthday card:** \\
    \hspace{1cm}* The image does not contain a birthday card; instead, it features a candle. \\
    \hspace{1cm}* **Result:** Fail \\
    2. **Single View:** \\
    \hspace{1cm}* The image shows the candle from a single angle, with no multiple views or perspectives. \\
    \hspace{1cm}* **Result:** Meet \\
    3. **Intact birthday card:** \\
    \hspace{1cm}* Since there is no birthday card in the image, this criterion is not applicable. \\
    \hspace{1cm}* **Result:** N/A \\
    4. **Plain Background:** \\
    \hspace{1cm}* The background of the image is solid black, with no distracting elements or patterns. \\
    \hspace{1cm}* **Result:** Meet \\

    **Conclusion:** \\

    Based on the evaluation criteria, the image fails to meet the first criterion, as it does not contain a birthday card. However, it meets the remaining criteria, showcasing the candle from a single view and having a plain black background. Therefore, the image is suitable for use as a birthday card. \\

    **Result:** Filter Out \\

    \hline

  \end{tabular}
  \caption{\textbf{Example of the data validation agent filtering for birthday card.}}
  \label{fig:vllmfilteringbirthdaycard}
\end{figure*}

\begin{figure*}[ht!]
  \centering
  \renewcommand{\arraystretch}{1.5} % Adjust row height

  \begin{tabular}{|p{.7\linewidth}|c|}
    \hline
    \textbf{Optimized Layer Diffusion Prompt} & \textbf{Foreground Instance Image} \\ \hline

    \parbox[c]{\LDTextBoxWidth}{
      High-resolution digital rendering of a single pancake alone, with a warm golden-brown color and a subtle sheen, ultra-realistic in style, evoking feelings of comfort and satisfaction, illuminated by soft morning light with a hint of shadow, viewed from a 45-degree angle to showcase its fluffy interior, featuring a delicate crispy texture on the outside and a soft, airy interior, set in a modern kitchen with a clean, minimalist aesthetic, circa present day.
    } & 
    \parbox[c]{\LDImageBoxWidth}{
    \includegraphics[width=\LDImageWidth]{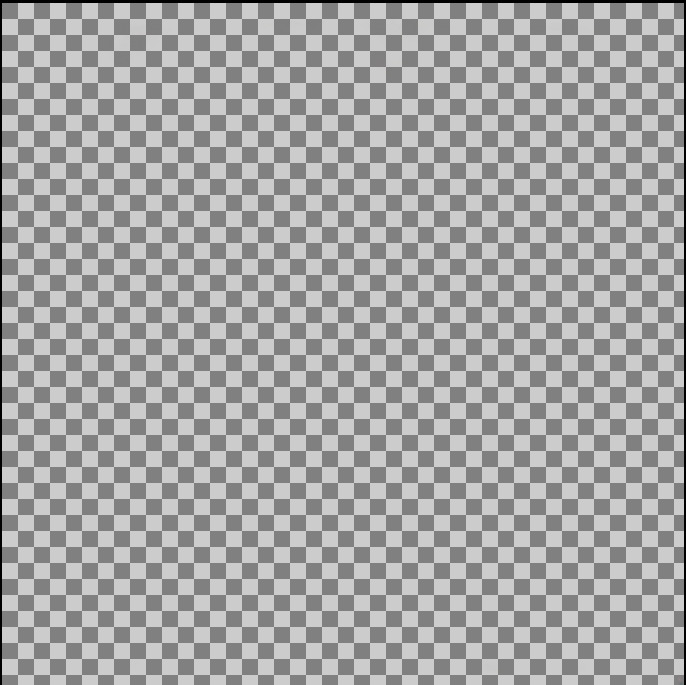}
    } \\ \hline
  \end{tabular}

  \begin{tabular}{|p{0.9625\linewidth}|}
    \hline
    \textbf{The Data Validation Agent's Response} \\ \hline
    **Image Description:** \\

    The image is a solid black square with no visible objects or features. \\

    **Evaluation Criteria:** \\

    1. **Single pancake:** \\
    \hspace{1cm}* The image does not contain any pancakes. \\
    \hspace{1cm}* **Result:** Fail \\
    2. **Single View:** \\
    \hspace{1cm}* There is no visible object to provide a view of. \\
    \hspace{1cm}* **Result:** N/A \\
    3. **Intact pancake:** \\
    \hspace{1cm}* There is no pancake to evaluate. \\
    \hspace{1cm}* **Result:** N/A \\
    4. **Plain Background:** \\
    \hspace{1cm}* The background is solid black, which can be considered plain. \\
    \hspace{1cm}* **Result:** Meet \\

    **Conclusion:** \\

    The image does not meet the criteria for containing a single pancake or providing a single view of one. However, it does meet the criterion for having a plain background. \\

    **Result:** Filter Out \\

    \hline

  \end{tabular}
  \caption{\textbf{Example of the data validation agent filtering for pancake.}}
  \label{fig:vllmfilteringpancake}
\end{figure*}

%%%%%%%%%%%%%%%%%%%%%%%%%%%%%%%%%%%%%%%%%%%%%%%%%%%%%%%%%%%%%%%%%%%%%%%%%%%%%%%%%%%%%%%%

\newcommand\MoredataBoxWidth{.4\textwidth}
\newcommand\MoredataImageWidth{.4\textwidth}
\begin{figure*}[ht!]
  \centering
  \begin{tabular}{@{}c@{}c@{}}

    \parbox[c]{\MoredataBoxWidth}{
    \includegraphics[width=\MoredataImageWidth]{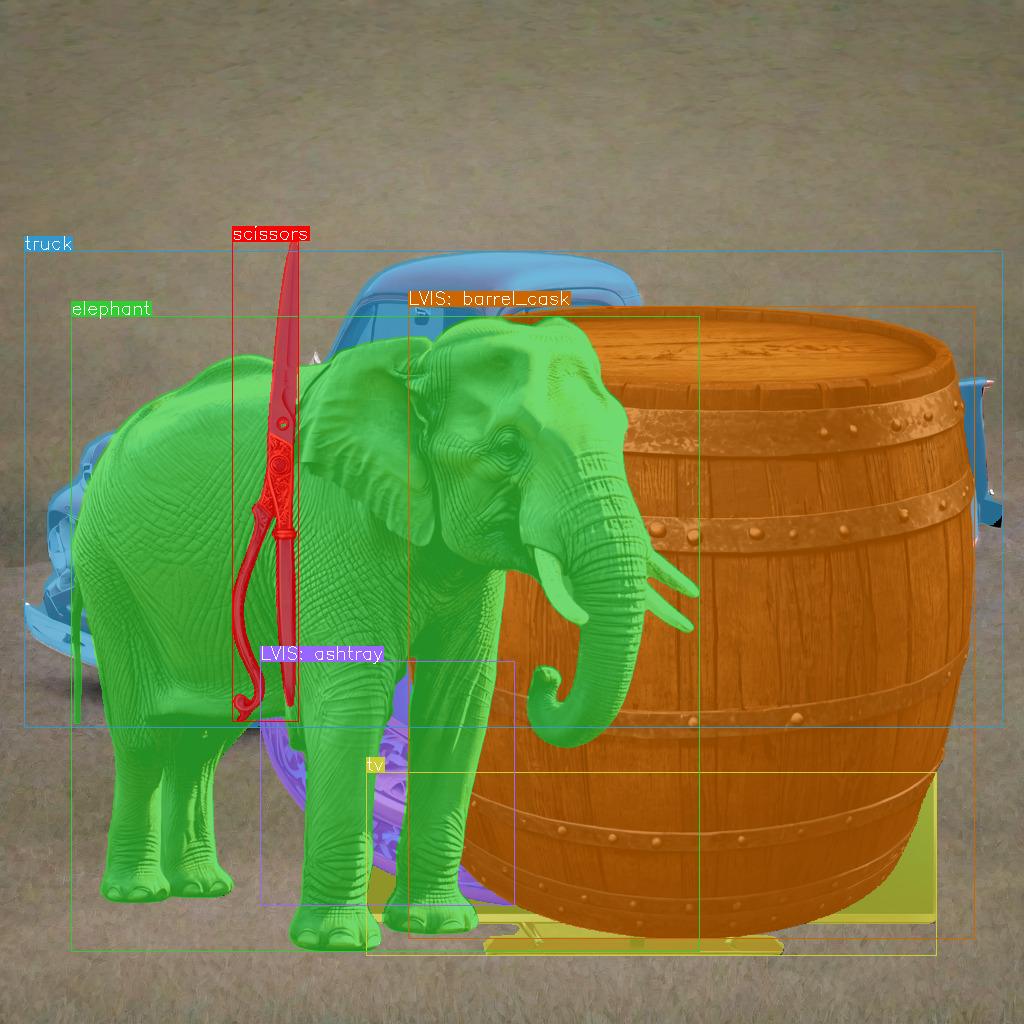} 
    } & 
    \parbox[c]{\MoredataBoxWidth}{
    \includegraphics[width=\MoredataImageWidth]{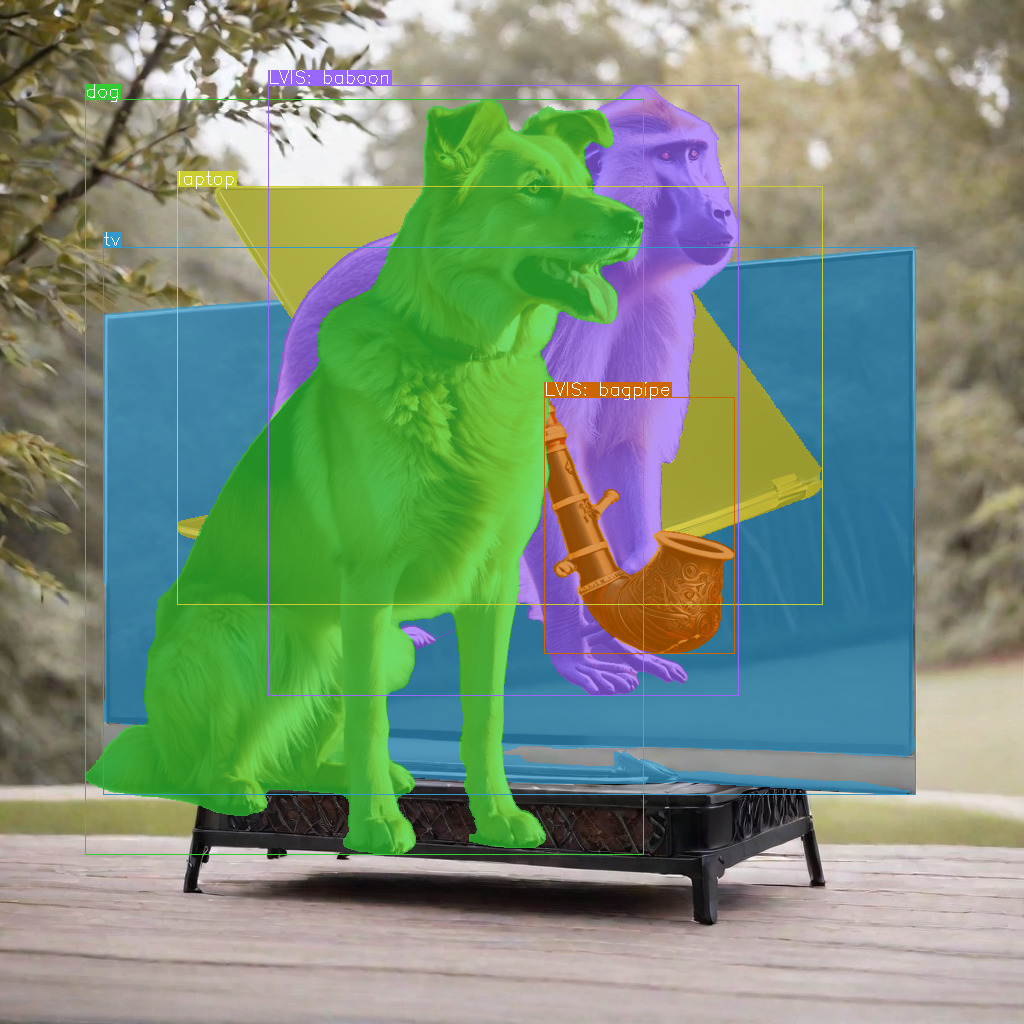} 
    } \\ 

    \parbox[c]{\MoredataBoxWidth}{
    \includegraphics[width=\MoredataImageWidth]{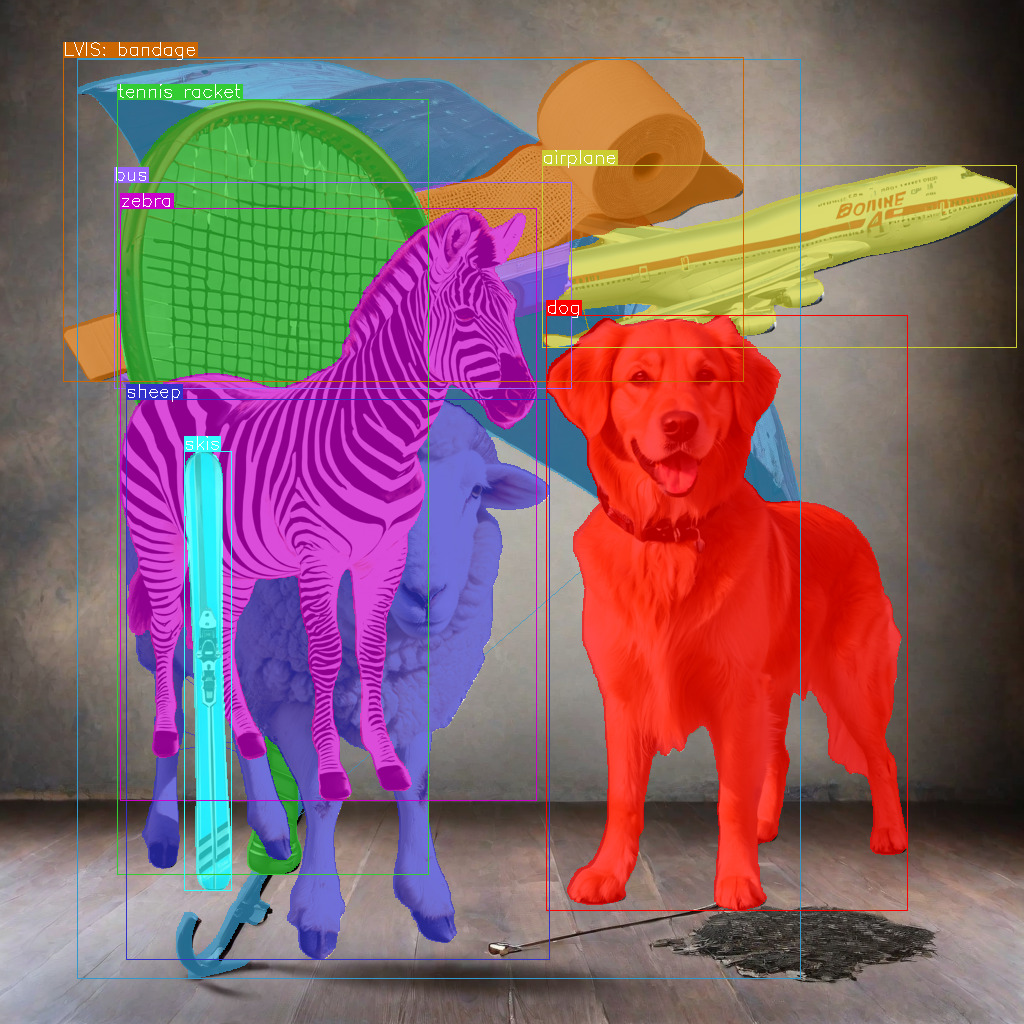} 
    } &
    \parbox[c]{\MoredataBoxWidth}{
    \includegraphics[width=\MoredataImageWidth]{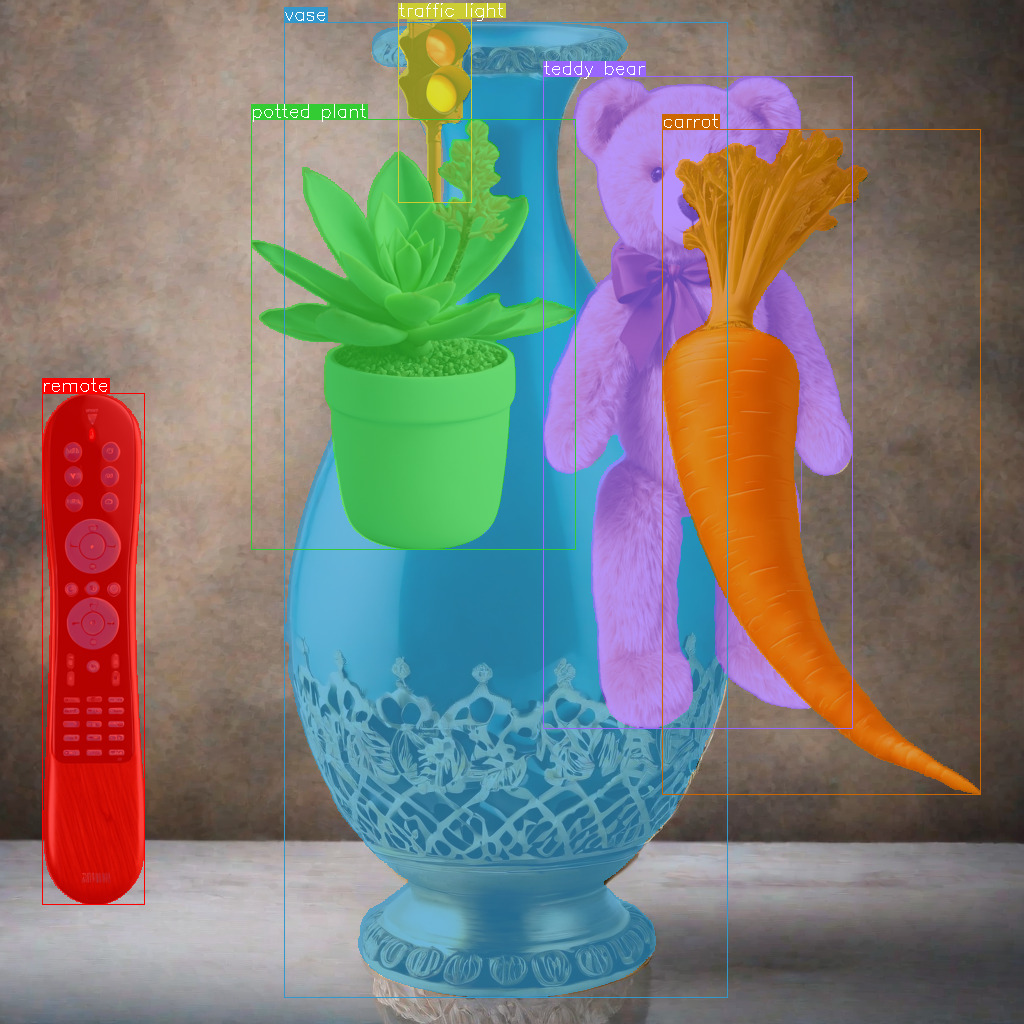} 
    } \\ 
    
    \parbox[c]{\MoredataBoxWidth}{
    \includegraphics[width=\MoredataImageWidth]{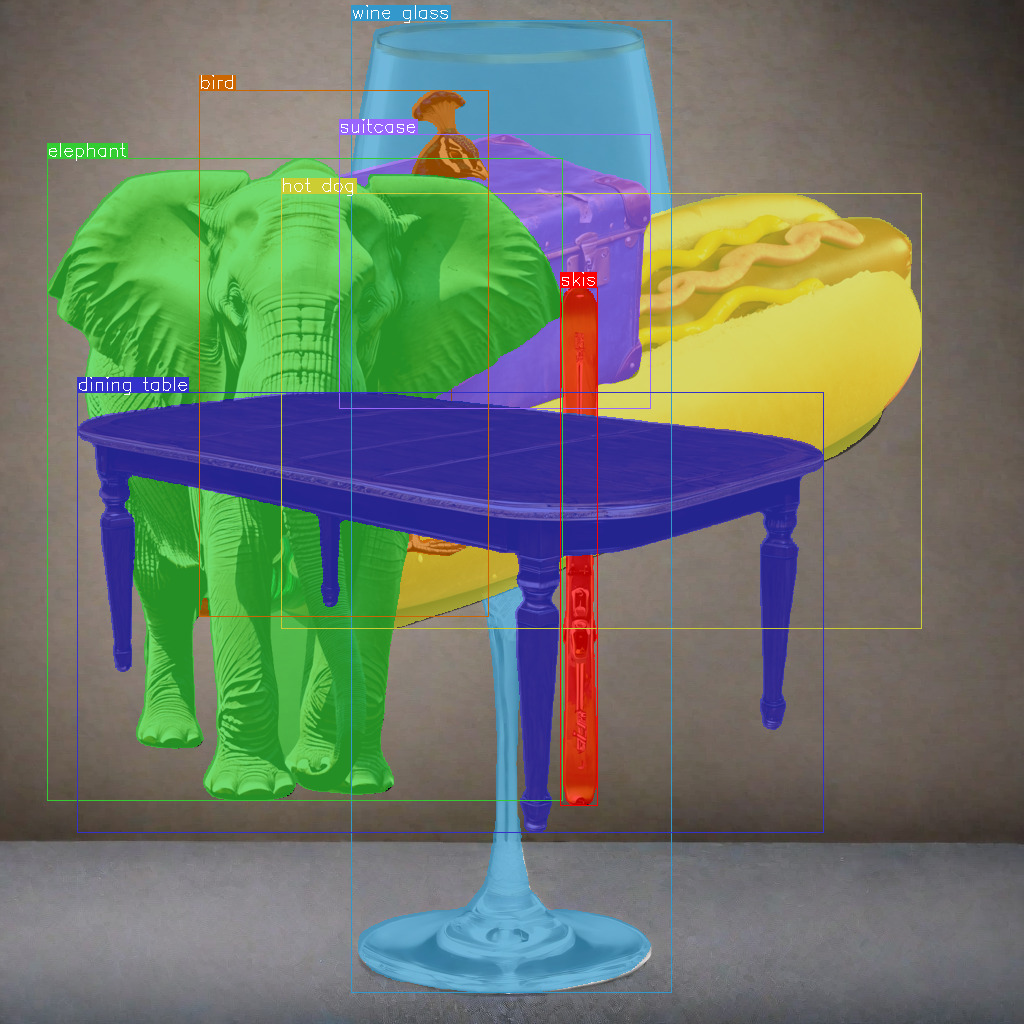} 
    } & 
    \parbox[c]{\MoredataBoxWidth}{
    \includegraphics[width=\MoredataImageWidth]{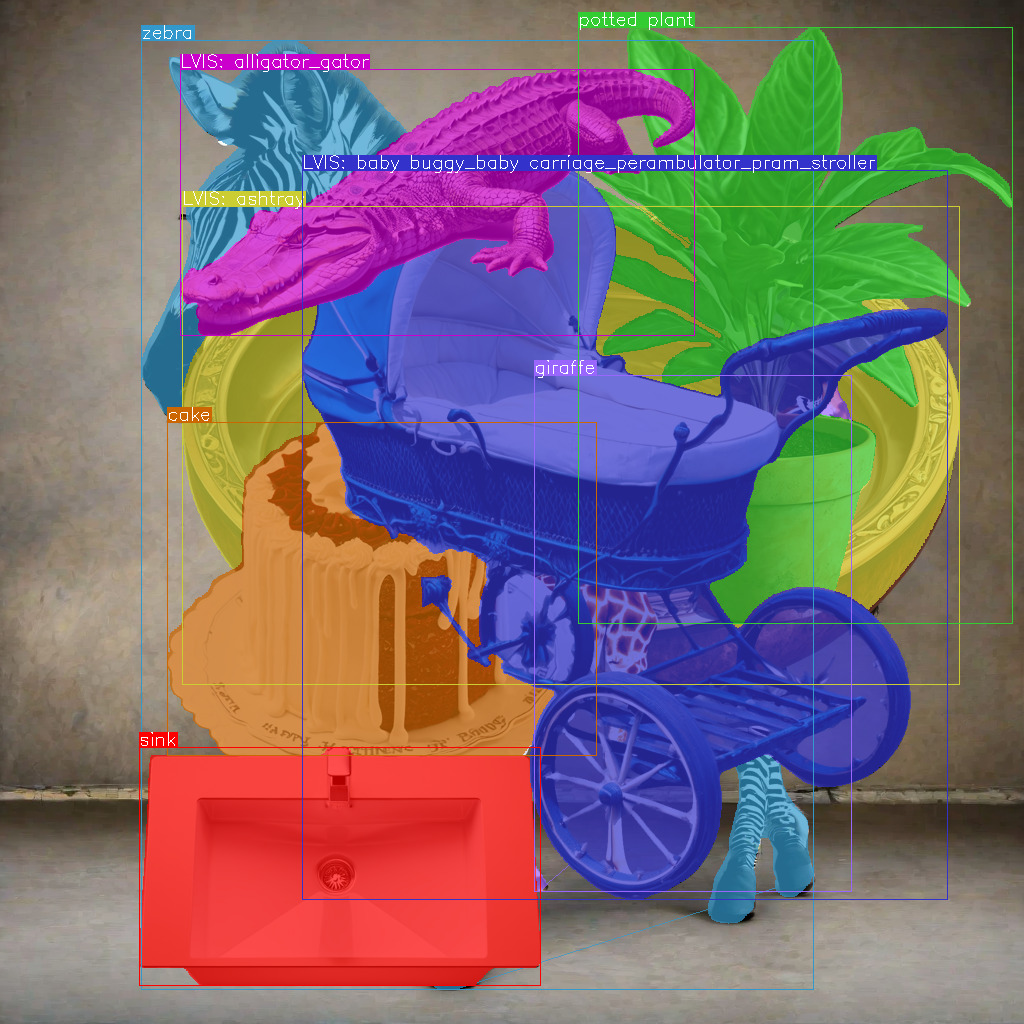} 
    } \\ 
    
  \end{tabular}
  \caption{\textbf{Additional Examples of Gen-n-Val Synthetic Data for COCO experiments.}}
  \label{fig:moredata1}
\end{figure*}

\begin{figure*}[ht!]
  \centering
  \begin{tabular}{@{}c@{}c@{}}

    \parbox[c]{\MoredataBoxWidth}{
    \includegraphics[width=\MoredataImageWidth]{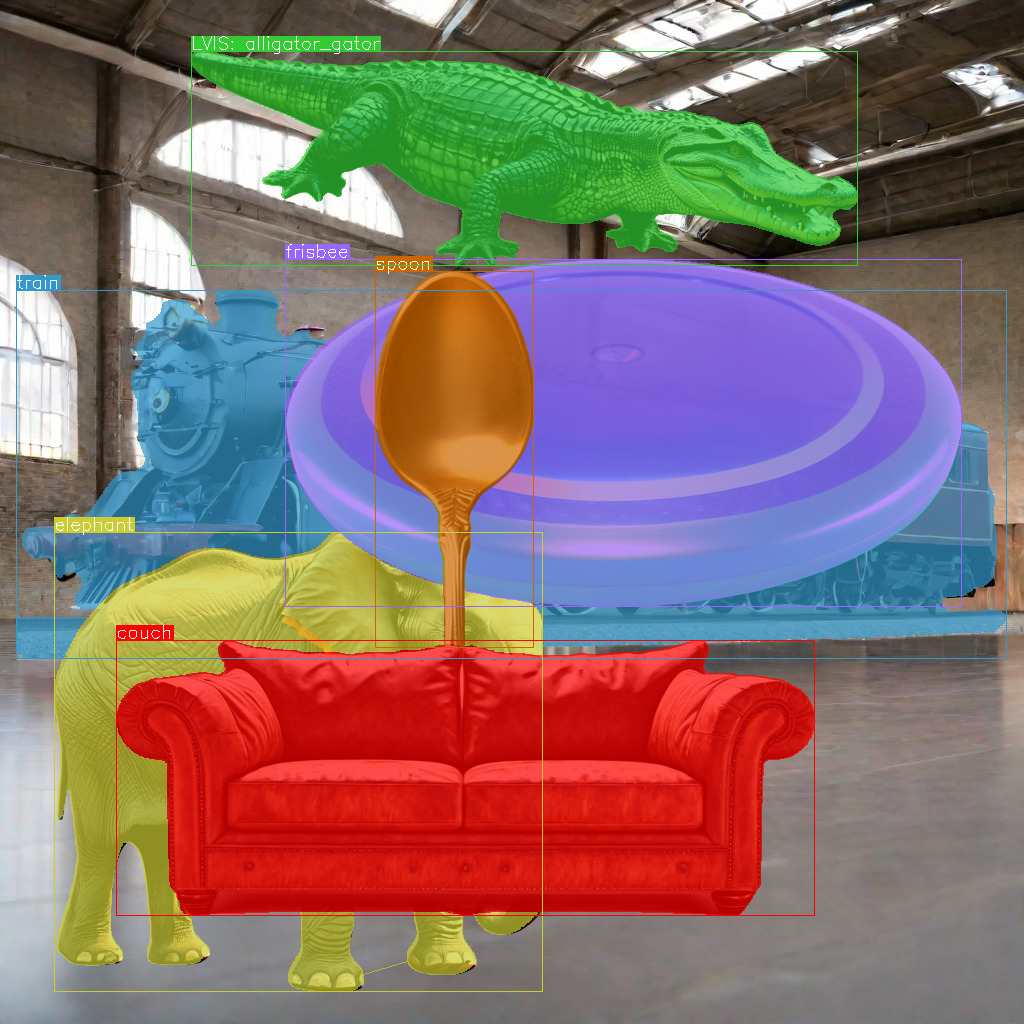} 
    } & 
    \parbox[c]{\MoredataBoxWidth}{
    \includegraphics[width=\MoredataImageWidth]{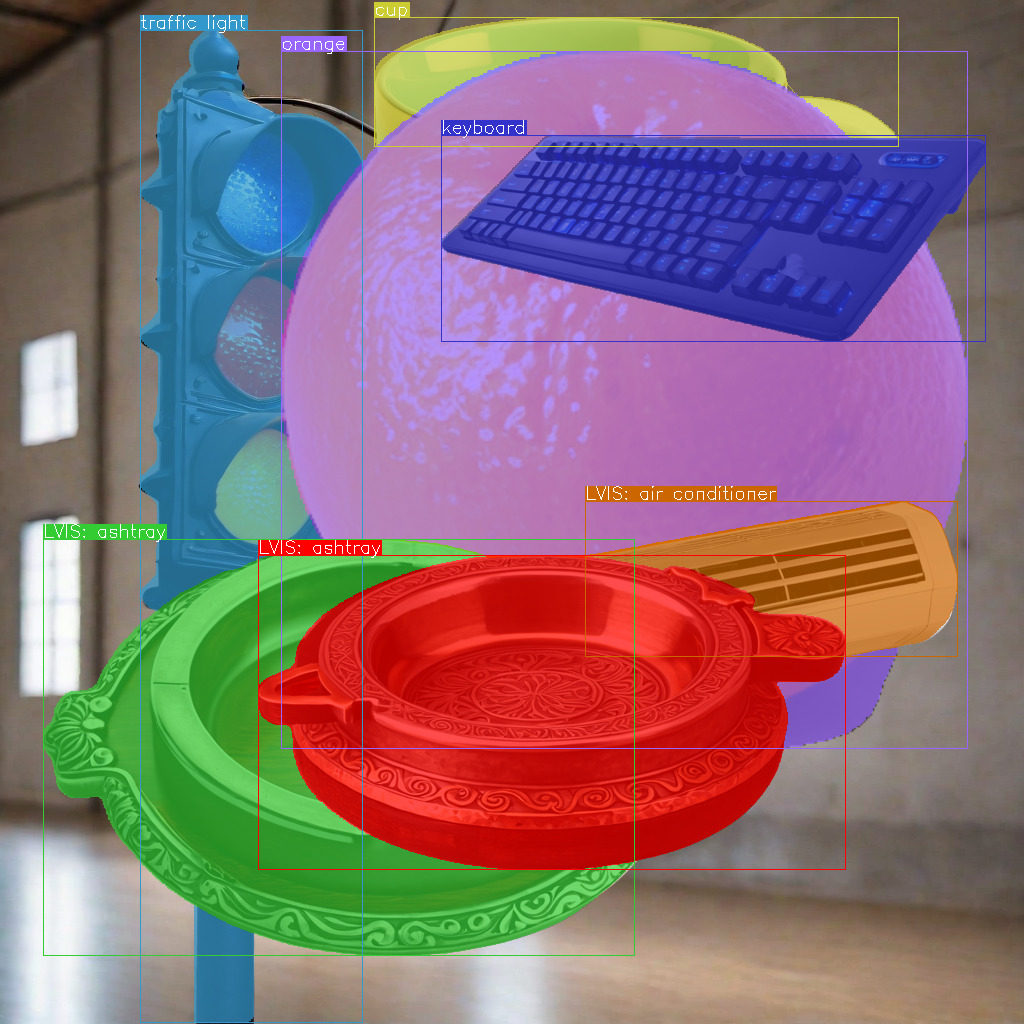} 
    } \\ 

    \parbox[c]{\MoredataBoxWidth}{
    \includegraphics[width=\MoredataImageWidth]{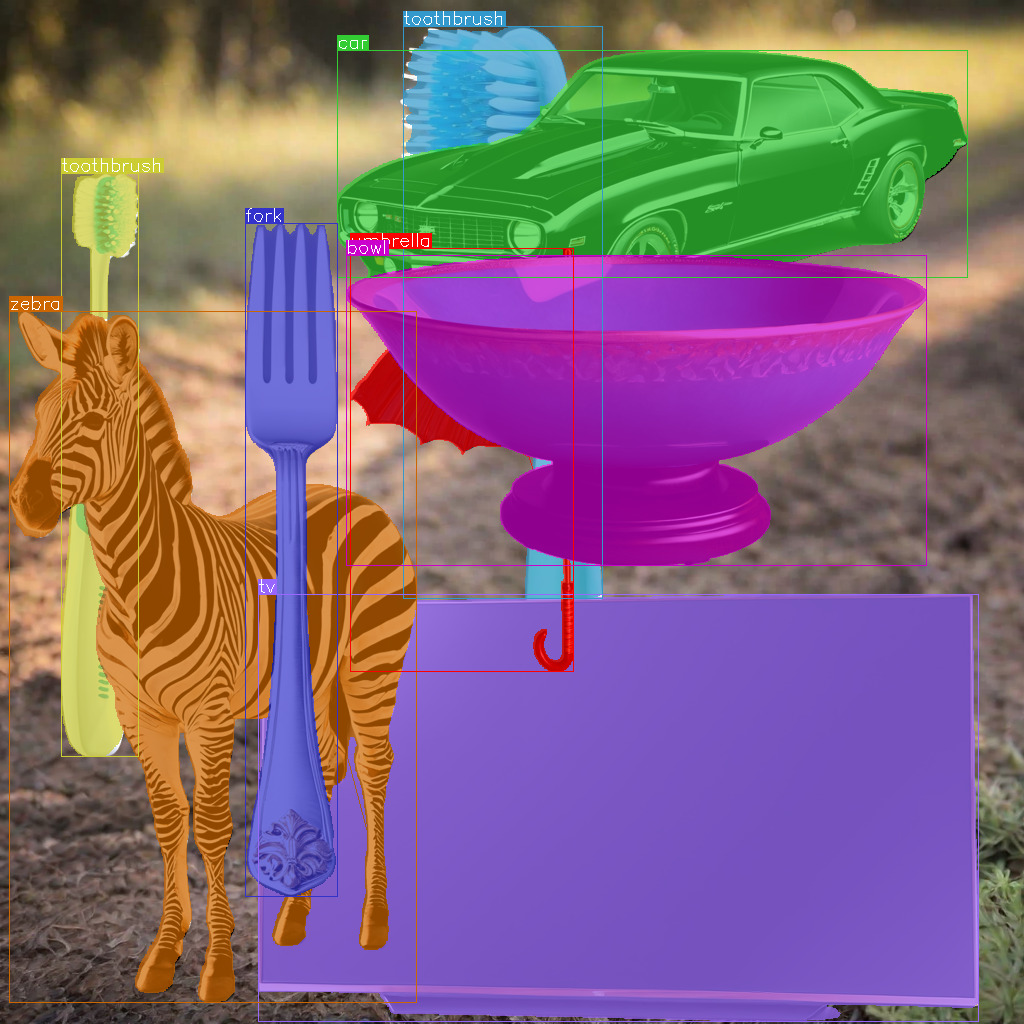} 
    } &
    \parbox[c]{\MoredataBoxWidth}{
    \includegraphics[width=\MoredataImageWidth]{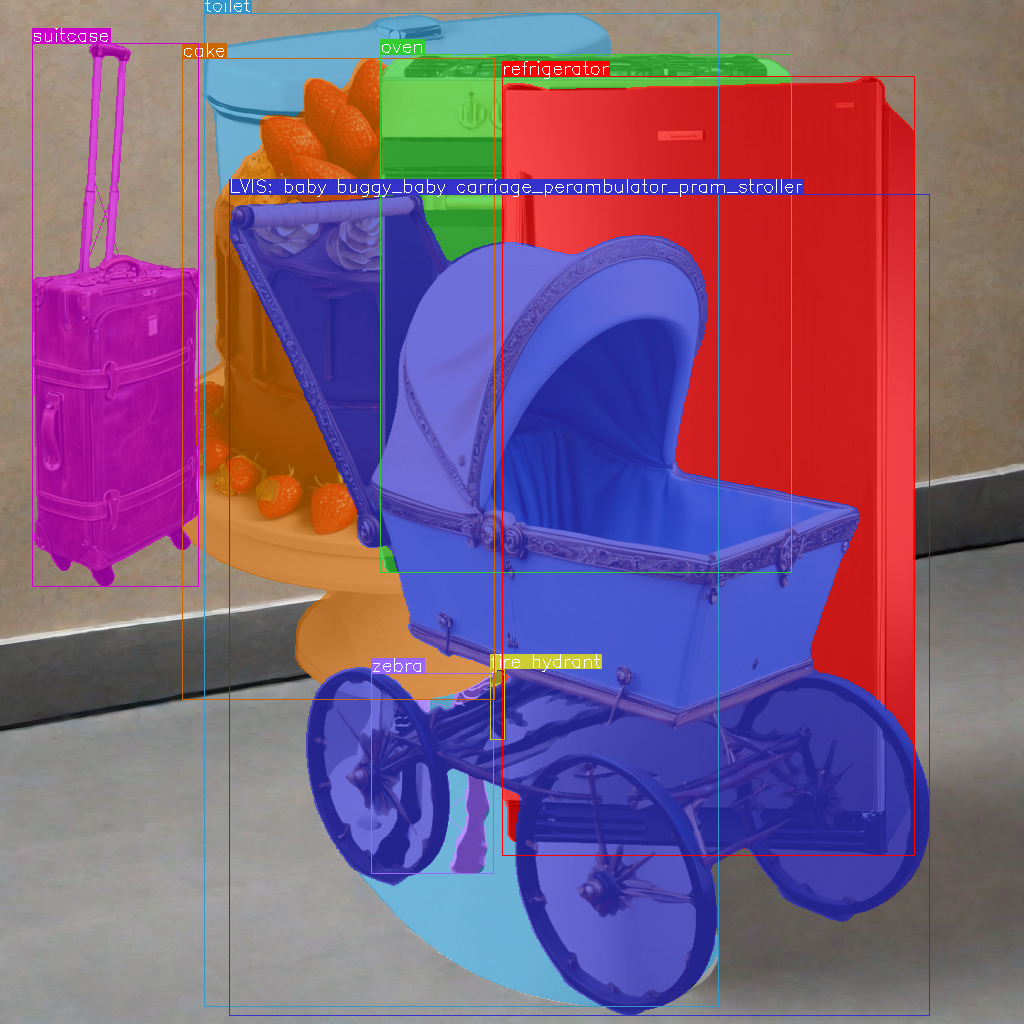} 
    } \\ 

    \parbox[c]{\MoredataBoxWidth}{
    \includegraphics[width=\MoredataImageWidth]{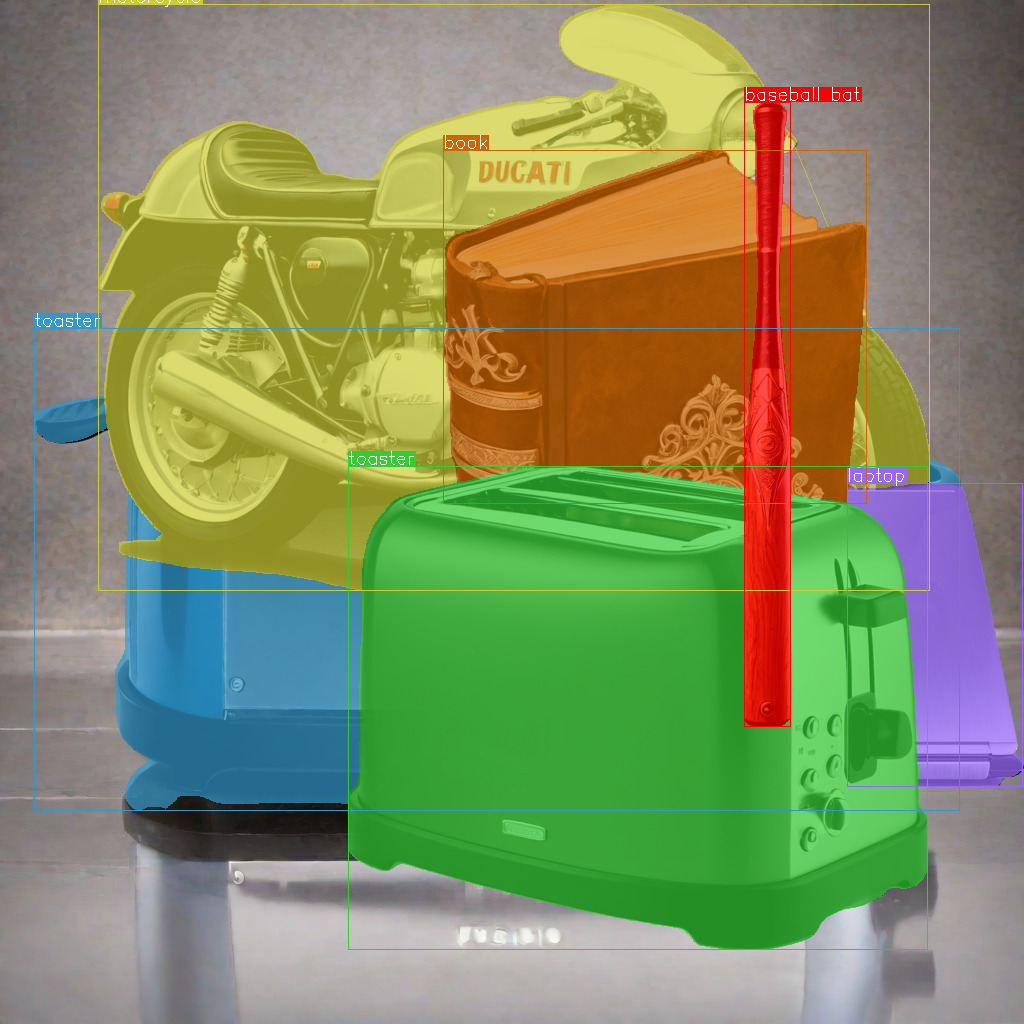} 
    } & 
    \parbox[c]{\MoredataBoxWidth}{
    \includegraphics[width=\MoredataImageWidth]{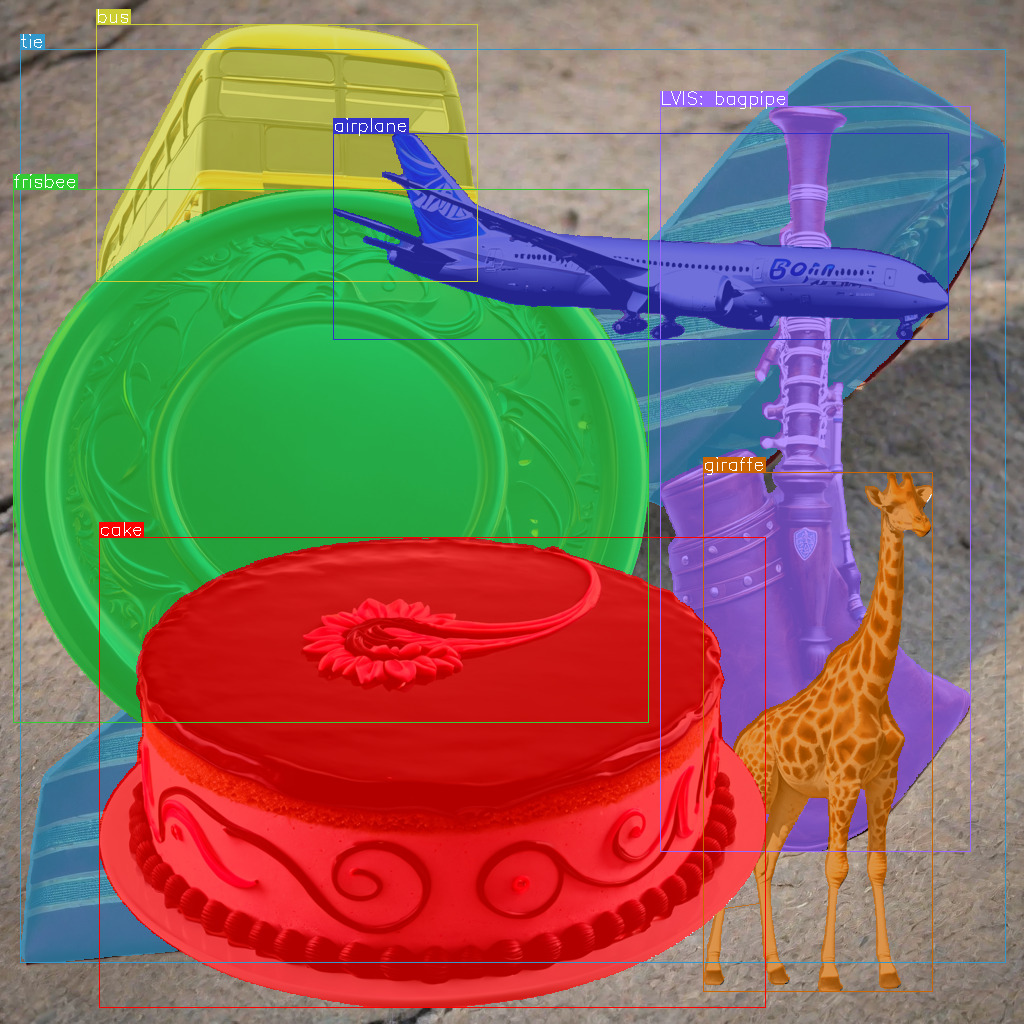} 
    } \\ 
    
  \end{tabular}
  \caption{\textbf{Additional Examples of Gen-n-Val Synthetic Data for COCO experiments.}}
  \label{fig:moredata2}
\end{figure*}

\begin{figure*}[ht!]
  \centering
  \begin{tabular}{@{}c@{}c@{}}

    \parbox[c]{\MoredataBoxWidth}{
    \includegraphics[width=\MoredataImageWidth]{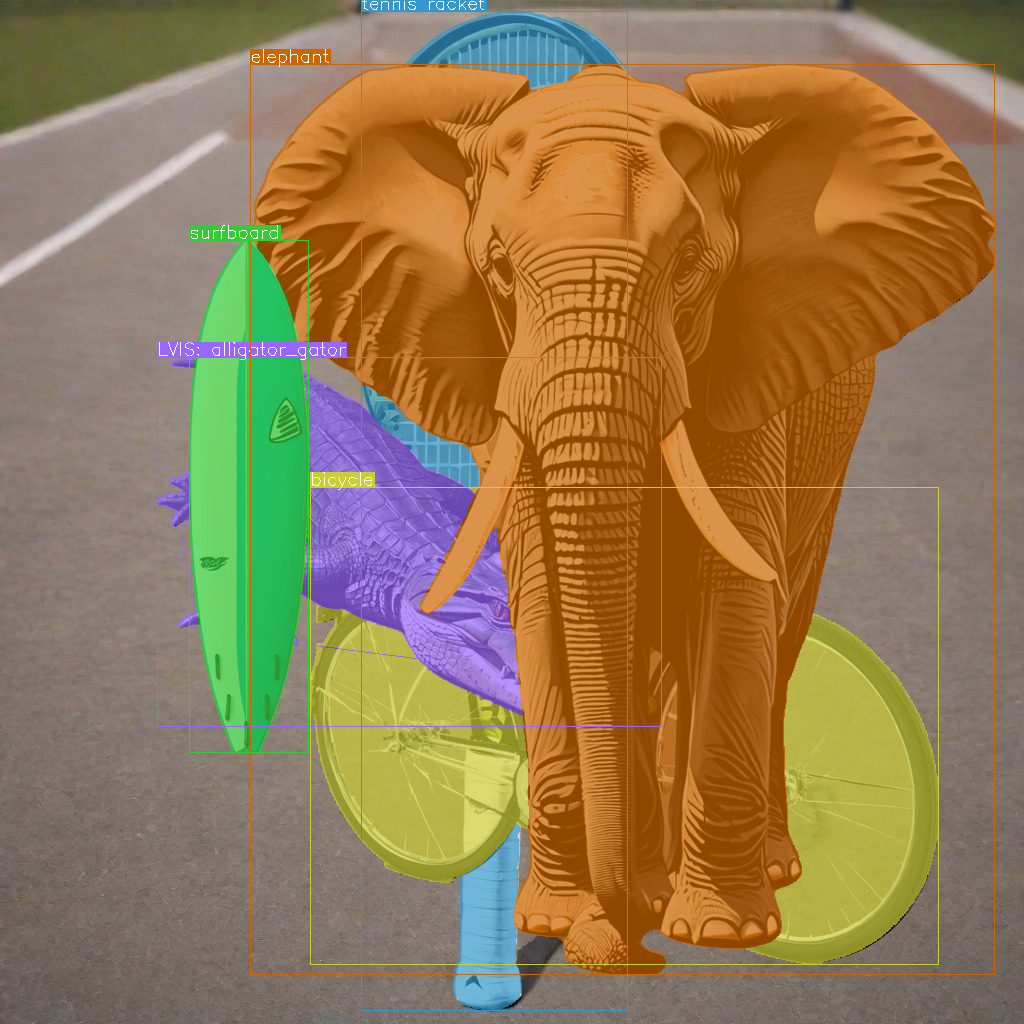} 
    } & 
    \parbox[c]{\MoredataBoxWidth}{
    \includegraphics[width=\MoredataImageWidth]{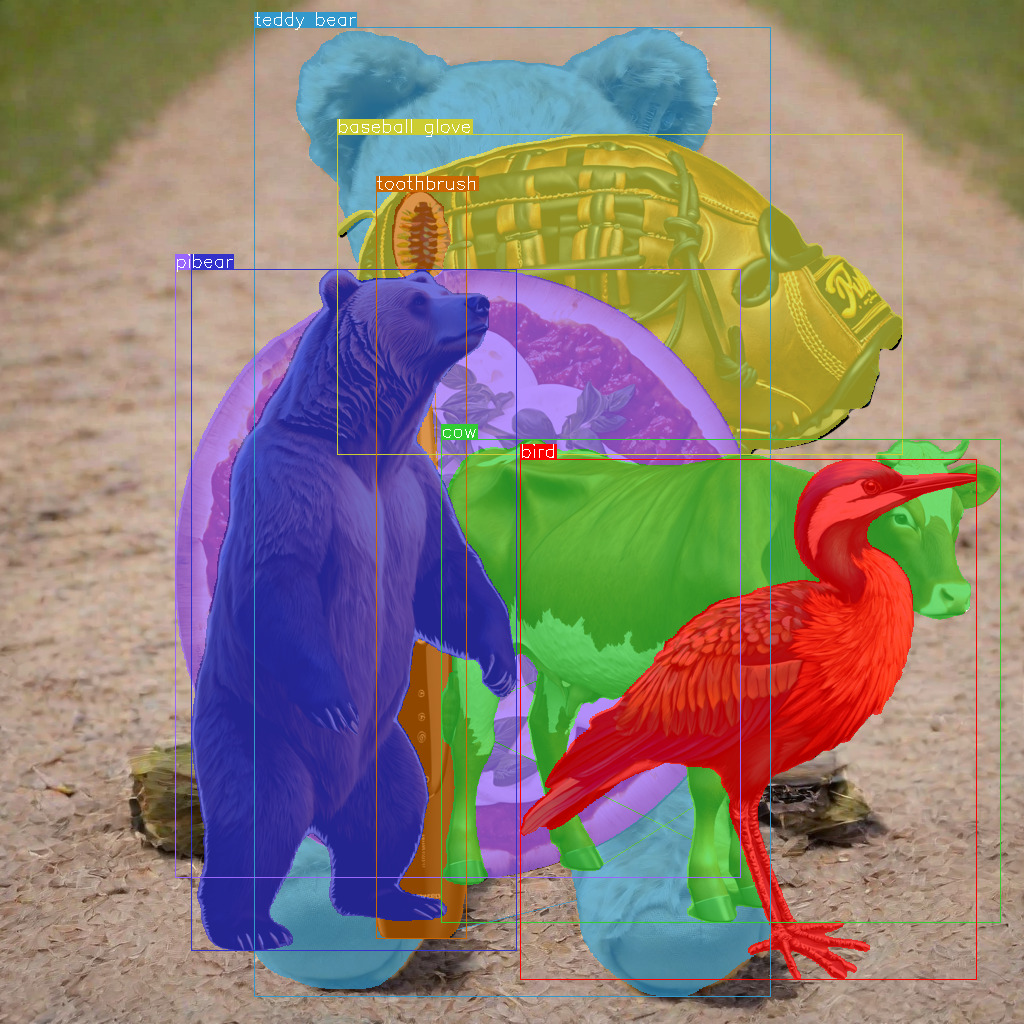} 
    } \\ 

    \parbox[c]{\MoredataBoxWidth}{
    \includegraphics[width=\MoredataImageWidth]{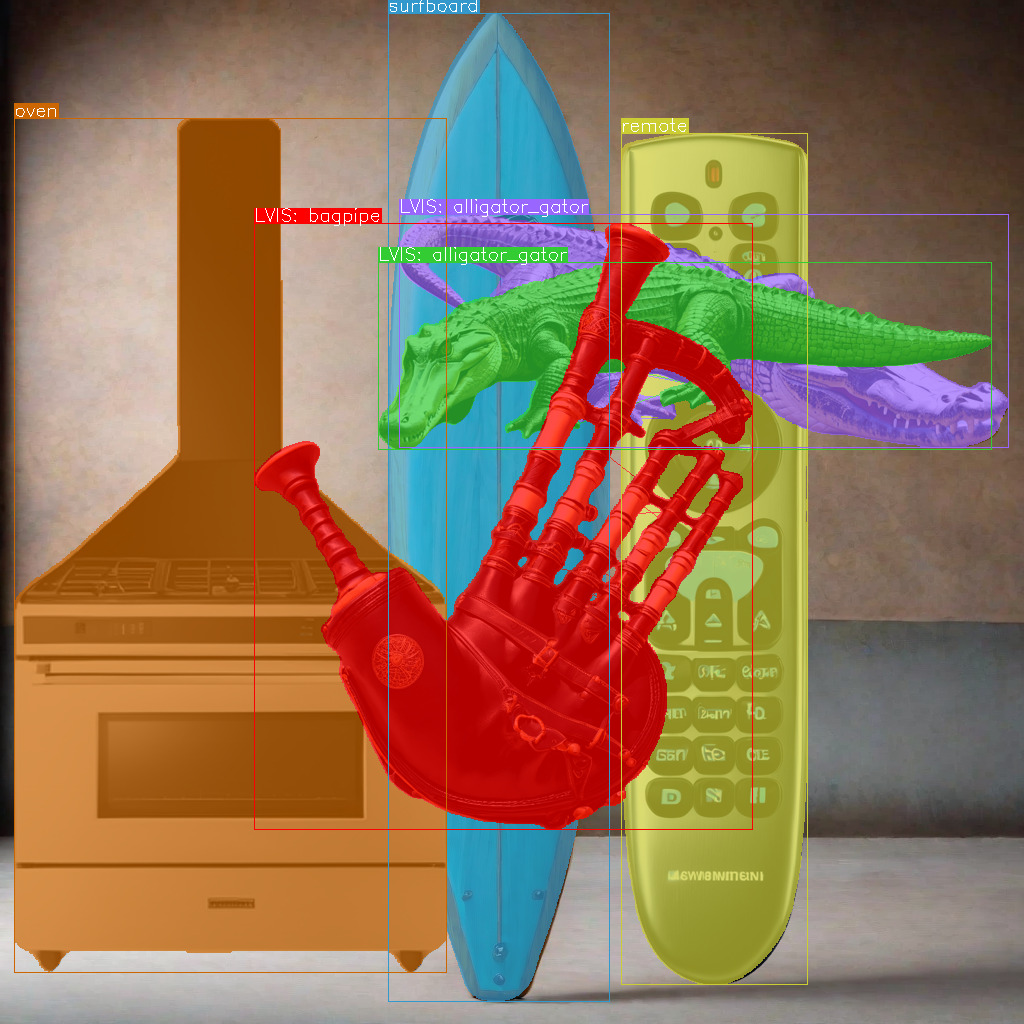} 
    } &
    \parbox[c]{\MoredataBoxWidth}{
    \includegraphics[width=\MoredataImageWidth]{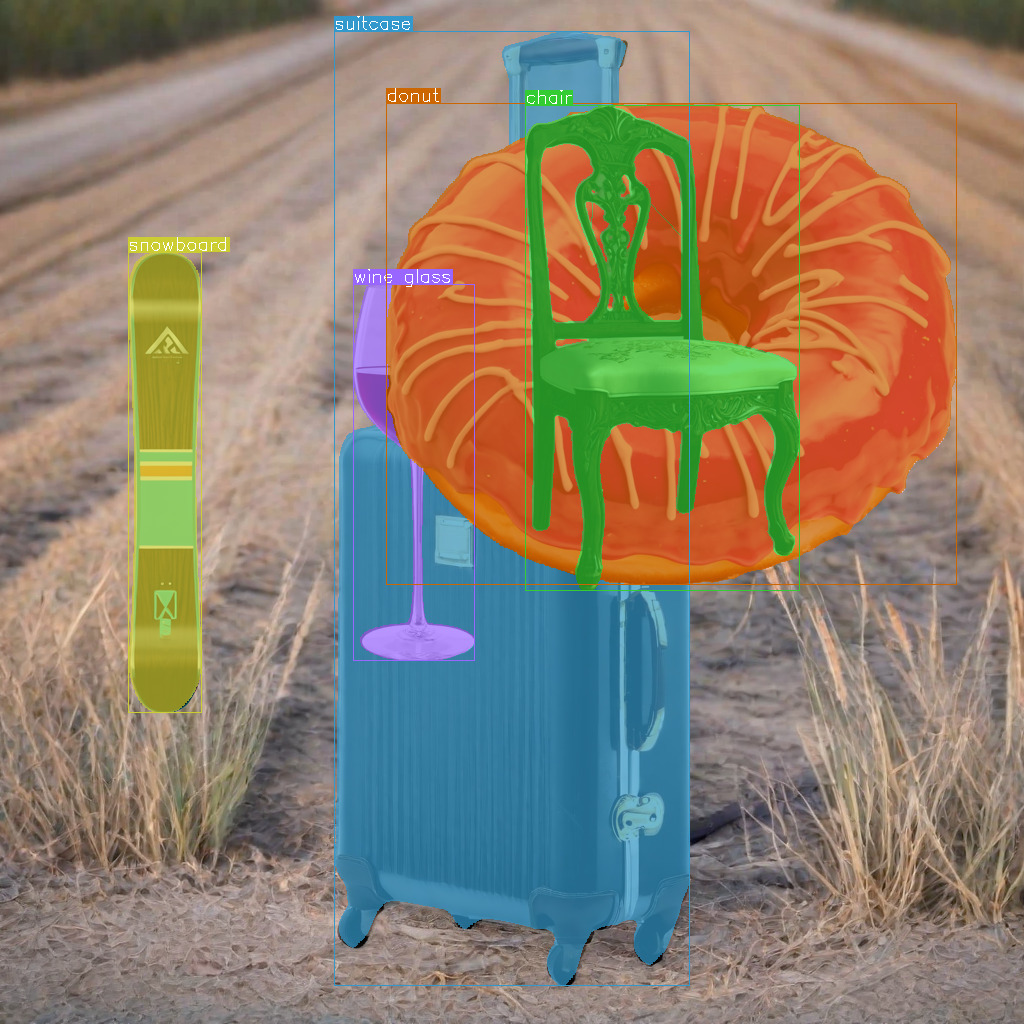} 
    } \\

    \parbox[c]{\MoredataBoxWidth}{
    \includegraphics[width=\MoredataImageWidth]{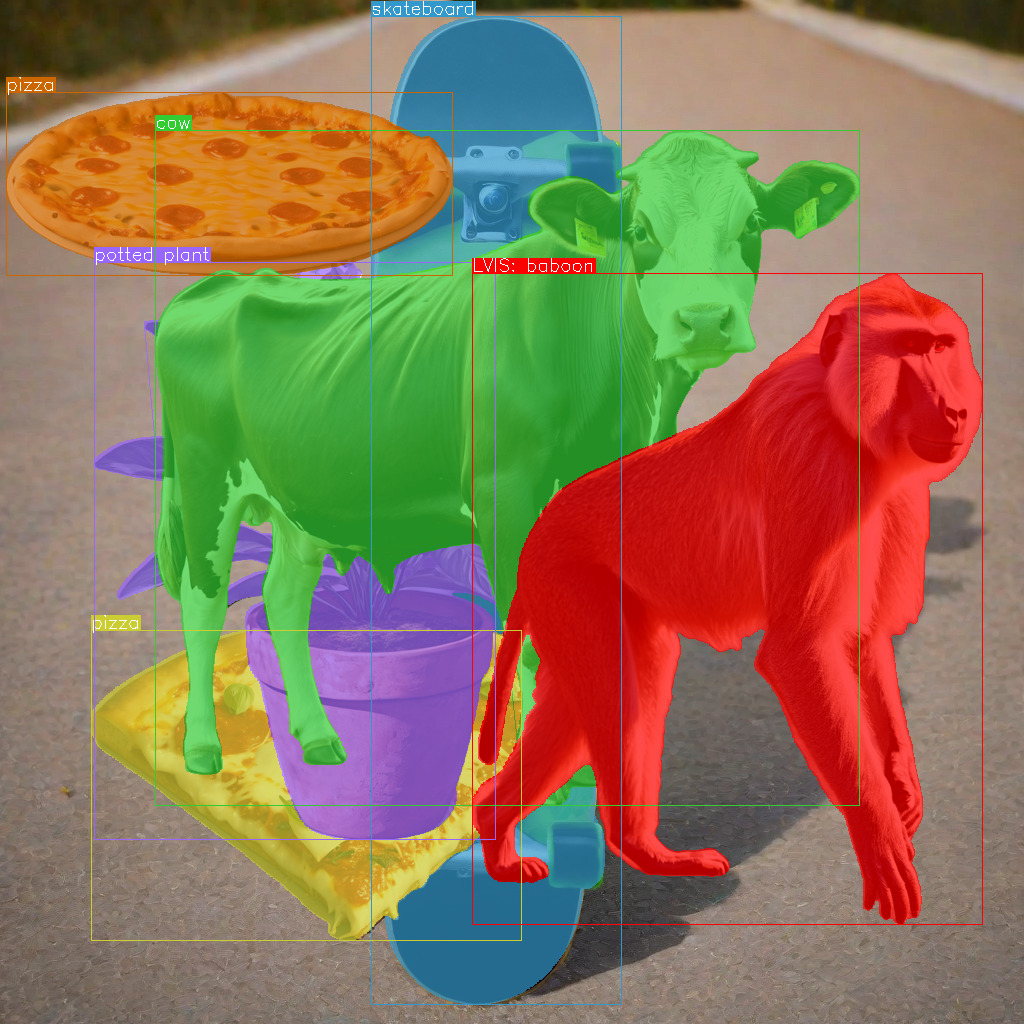} 
    } & 
    \parbox[c]{\MoredataBoxWidth}{
    \includegraphics[width=\MoredataImageWidth]{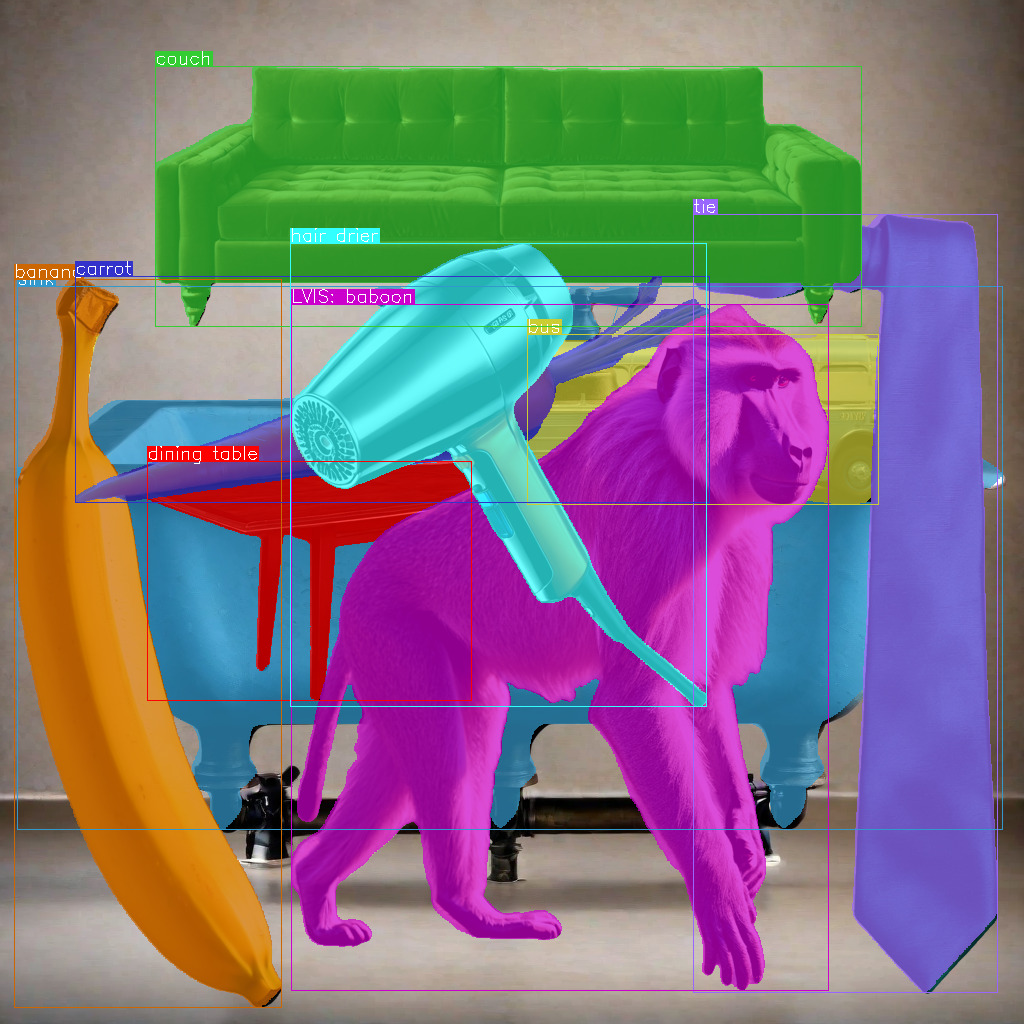} 
    } \\ 
    
  \end{tabular}
  \caption{\textbf{Additional Examples of Gen-n-Val Synthetic Data for COCO experiments.}}
  \label{fig:moredata3}
\end{figure*}

\begin{figure*}[ht!]
  \centering
  \begin{tabular}{@{}c@{}c@{}}

    \parbox[c]{\MoredataBoxWidth}{
    \includegraphics[width=\MoredataImageWidth]{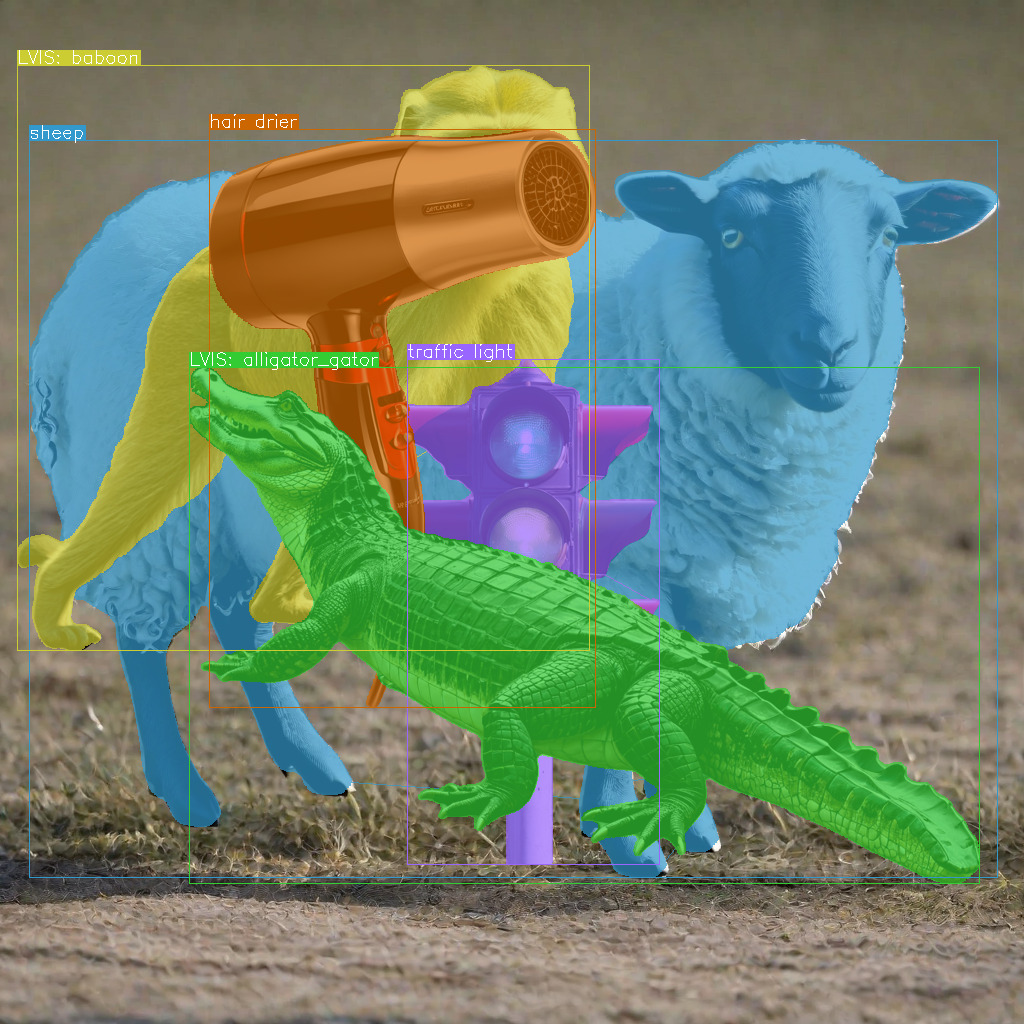} 
    } & 
    \parbox[c]{\MoredataBoxWidth}{
    \includegraphics[width=\MoredataImageWidth]{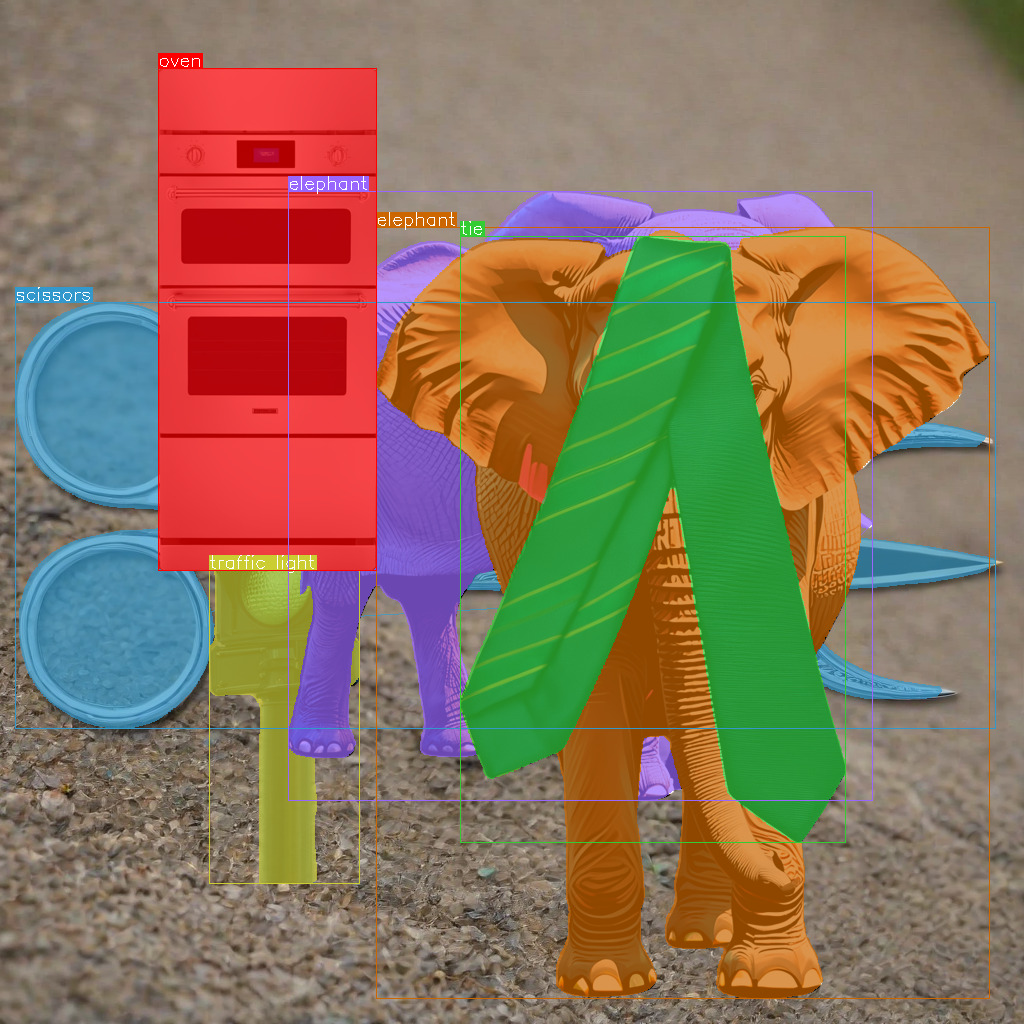} 
    } \\ 

    \parbox[c]{\MoredataBoxWidth}{
    \includegraphics[width=\MoredataImageWidth]{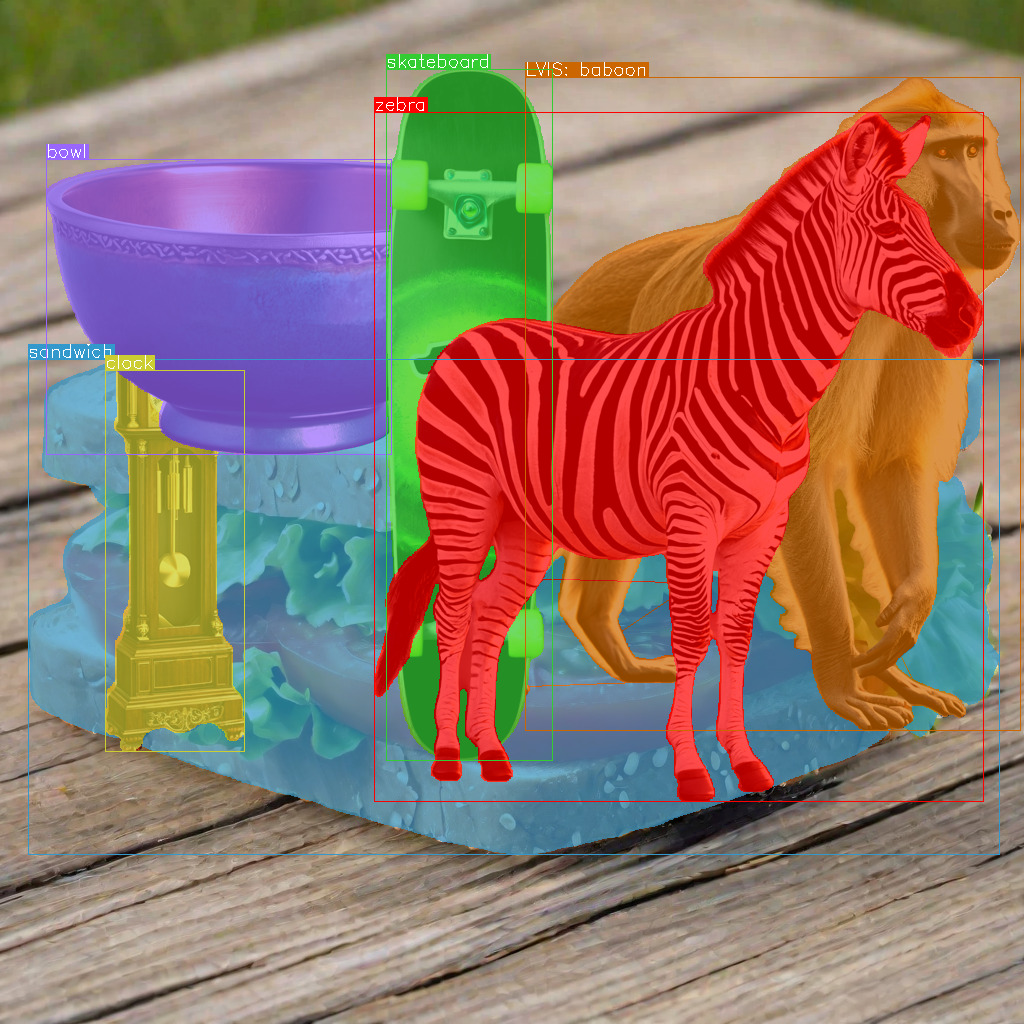} 
    } &
    \parbox[c]{\MoredataBoxWidth}{
    \includegraphics[width=\MoredataImageWidth]{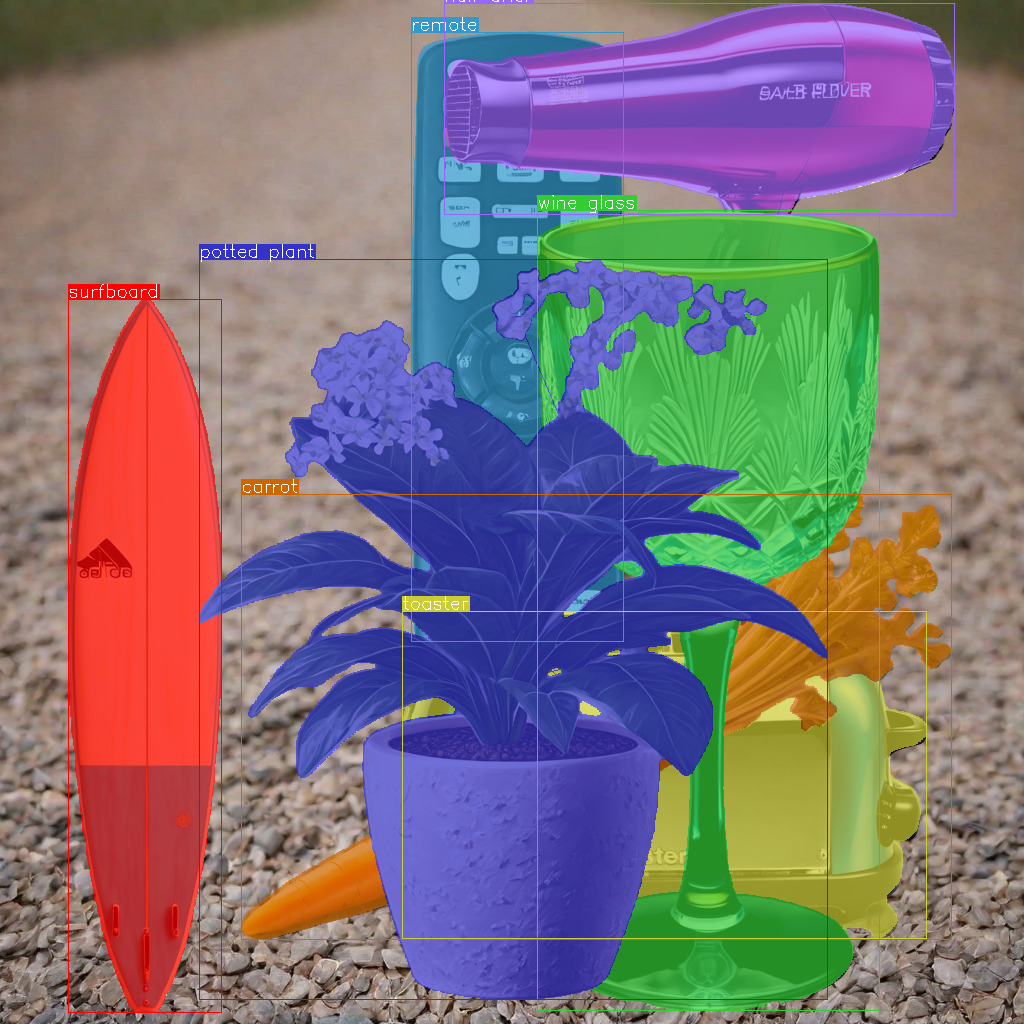} 
    } \\ 

    \parbox[c]{\MoredataBoxWidth}{
    \includegraphics[width=\MoredataImageWidth]{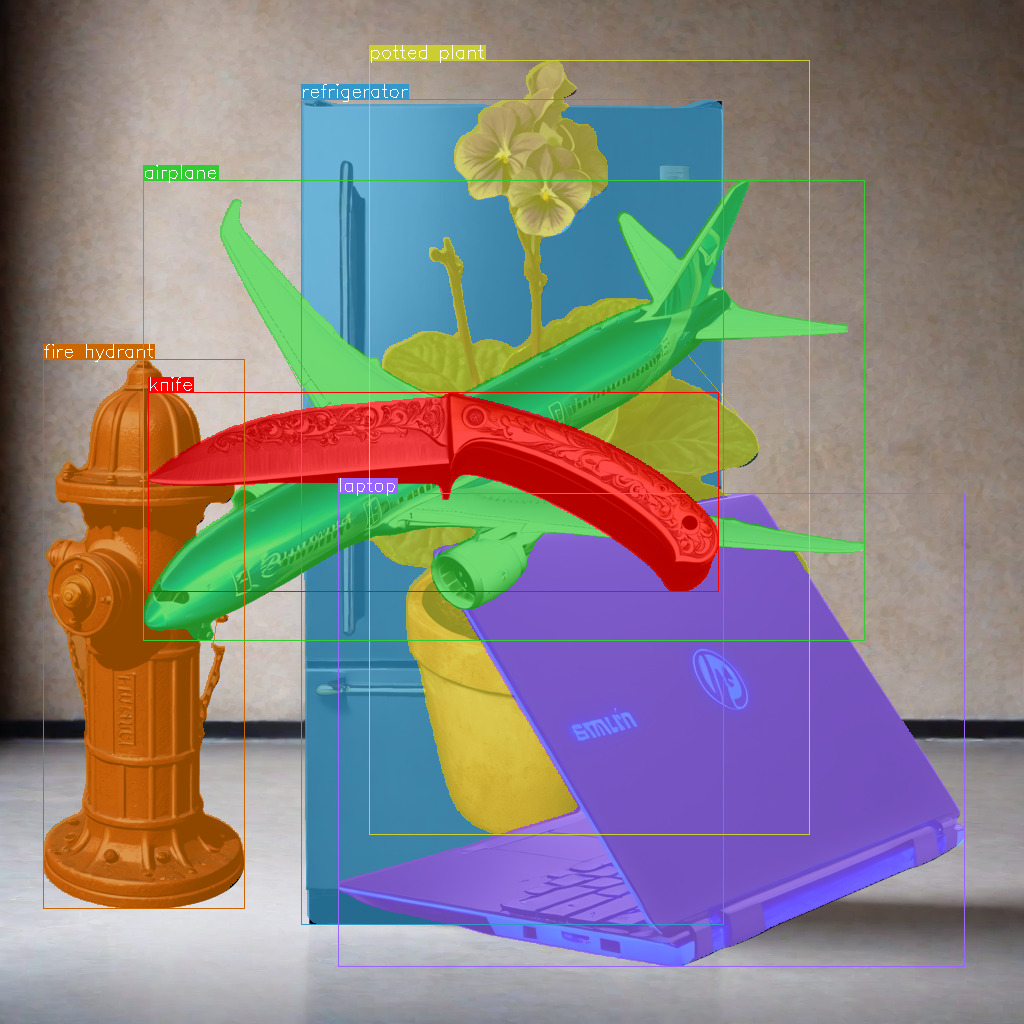} 
    } & 
    \parbox[c]{\MoredataBoxWidth}{
    \includegraphics[width=\MoredataImageWidth]{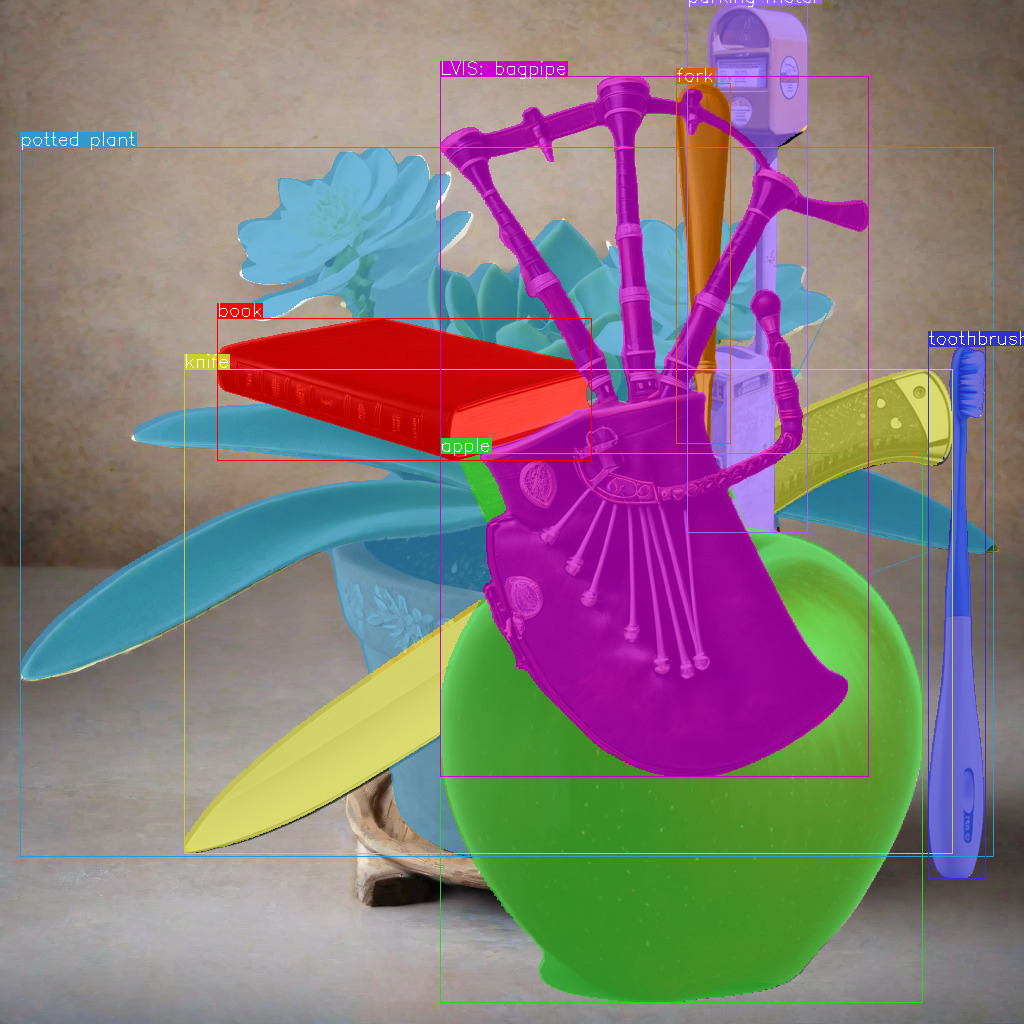} 
    } \\ 
    
  \end{tabular}
  \caption{\textbf{Additional Examples of Gen-n-Val Synthetic Data for COCO experiments.}}
  \label{fig:moredata4}
\end{figure*}

\begin{figure*}[ht!]
  \centering
  \begin{tabular}{@{}c@{}c@{}}

    \parbox[c]{\MoredataBoxWidth}{
    \includegraphics[width=\MoredataImageWidth]{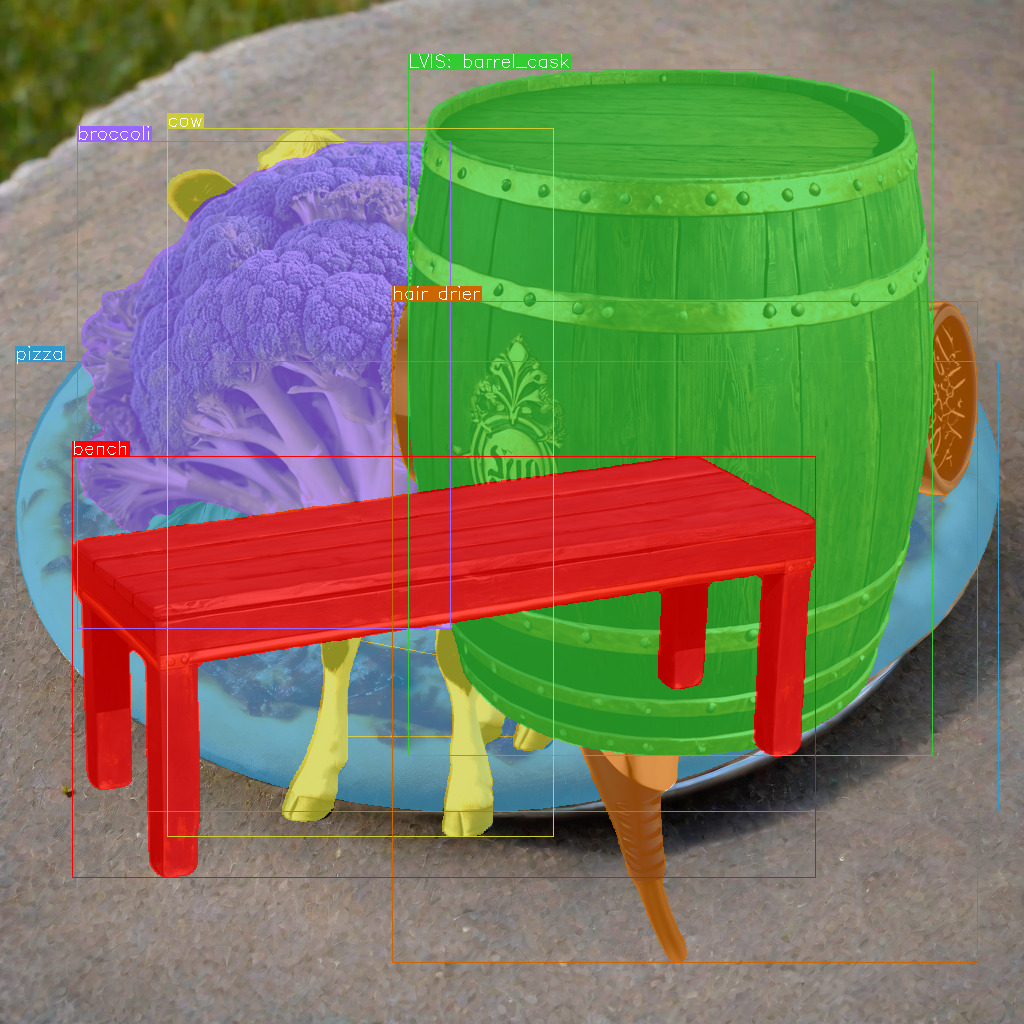} 
    } & 
    \parbox[c]{\MoredataBoxWidth}{
    \includegraphics[width=\MoredataImageWidth]{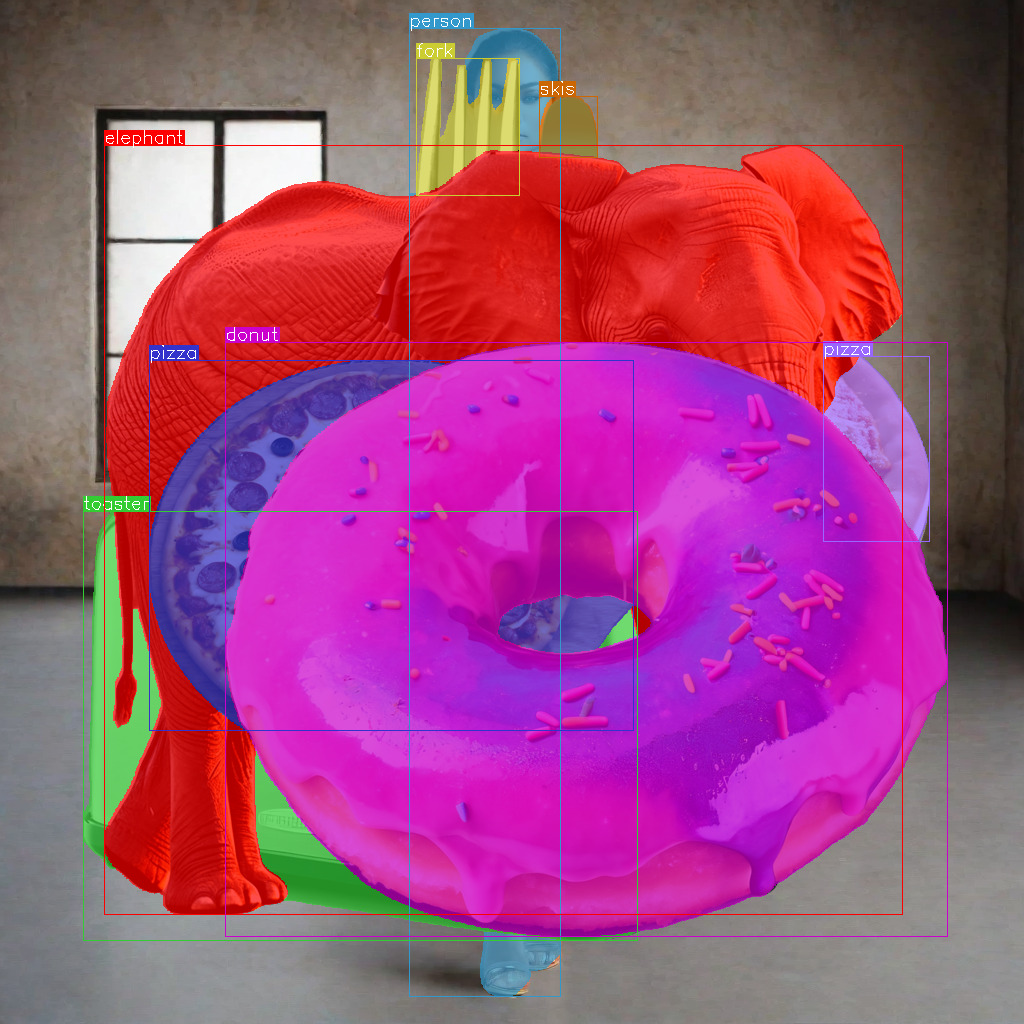} 
    } \\ 

    \parbox[c]{\MoredataBoxWidth}{
    \includegraphics[width=\MoredataImageWidth]{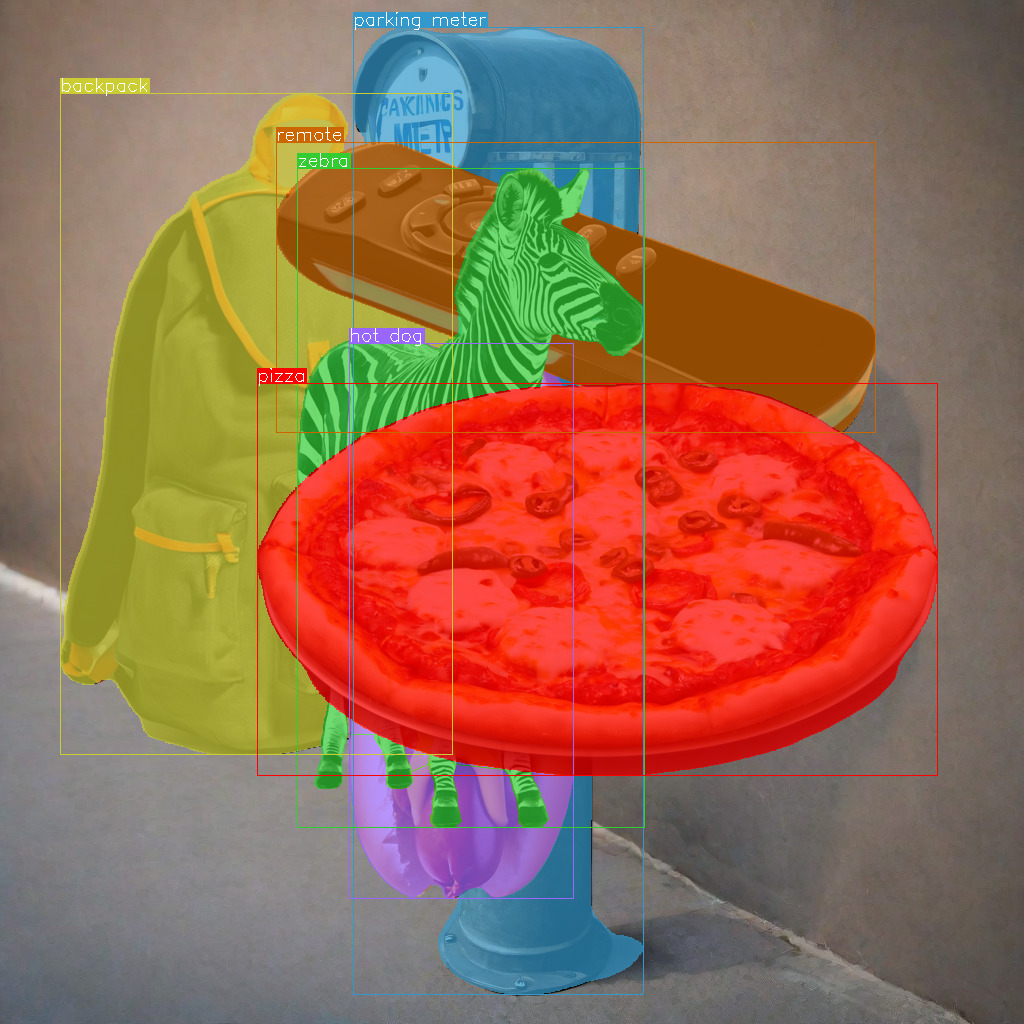} 
    } &
    \parbox[c]{\MoredataBoxWidth}{
    \includegraphics[width=\MoredataImageWidth]{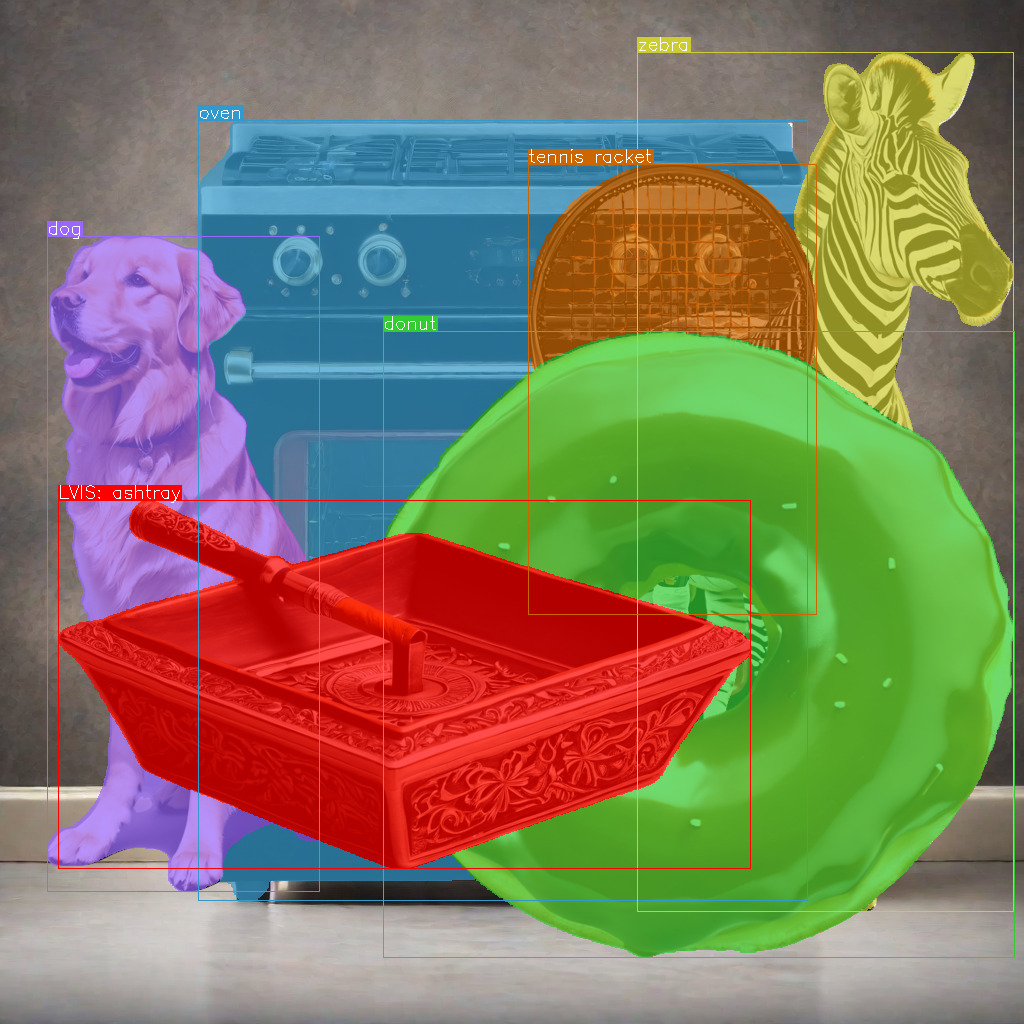} 
    } \\ 

    \parbox[c]{\MoredataBoxWidth}{
    \includegraphics[width=\MoredataImageWidth]{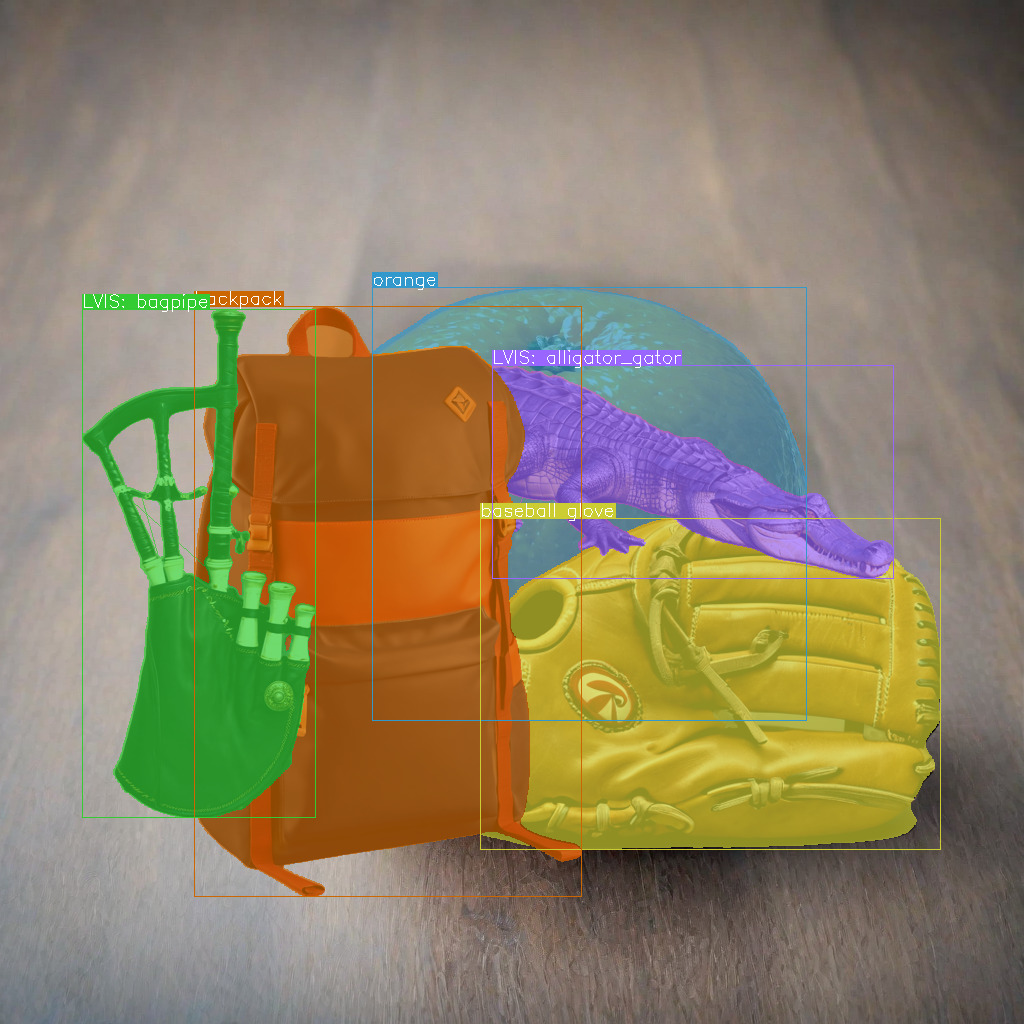} 
    } & 
    \parbox[c]{\MoredataBoxWidth}{
    \includegraphics[width=\MoredataImageWidth]{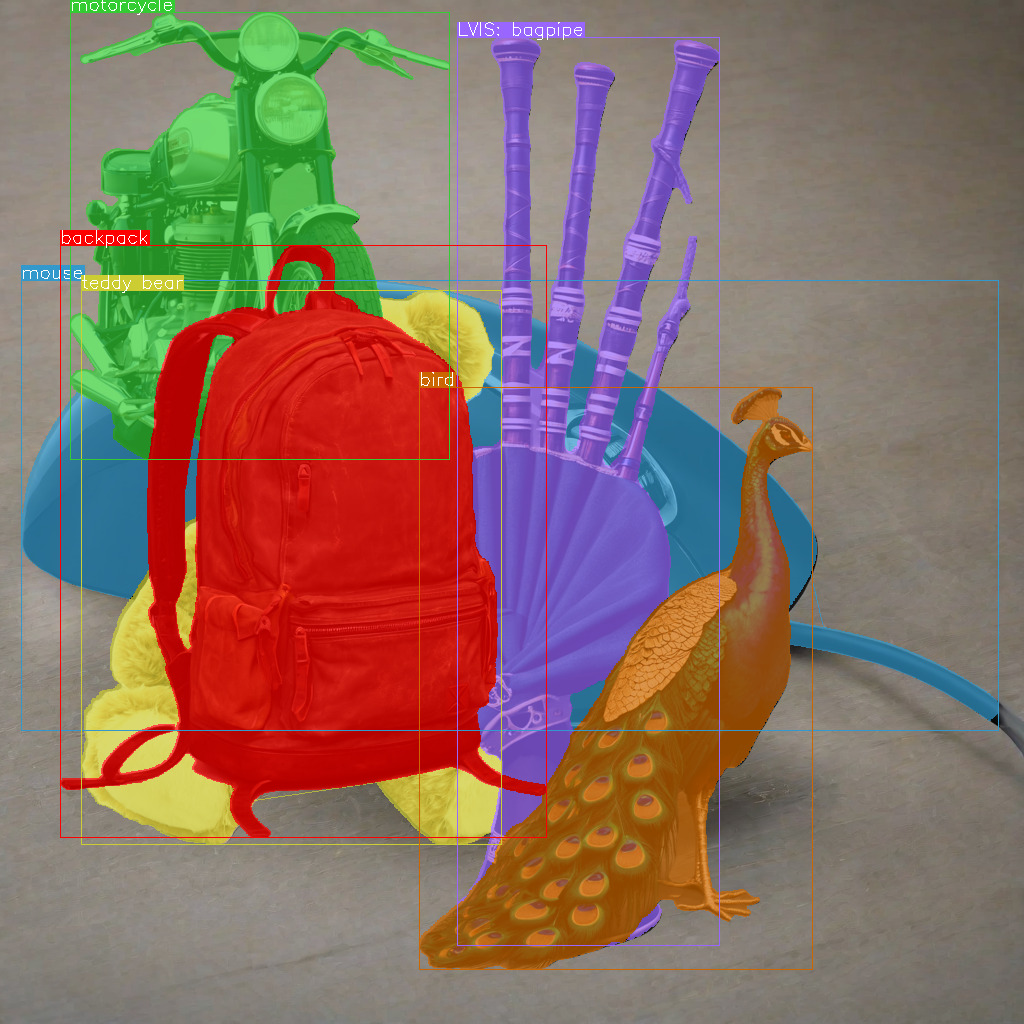} 
    } \\ 
    
  \end{tabular}
  \caption{\textbf{Additional Examples of Gen-n-Val Synthetic Data for COCO experiments.}}
  \label{fig:moredata5}
\end{figure*}

\begin{figure*}[ht!]
  \centering
  \begin{tabular}{@{}c@{}c@{}}

    \parbox[c]{\MoredataBoxWidth}{
    \includegraphics[width=\MoredataImageWidth]{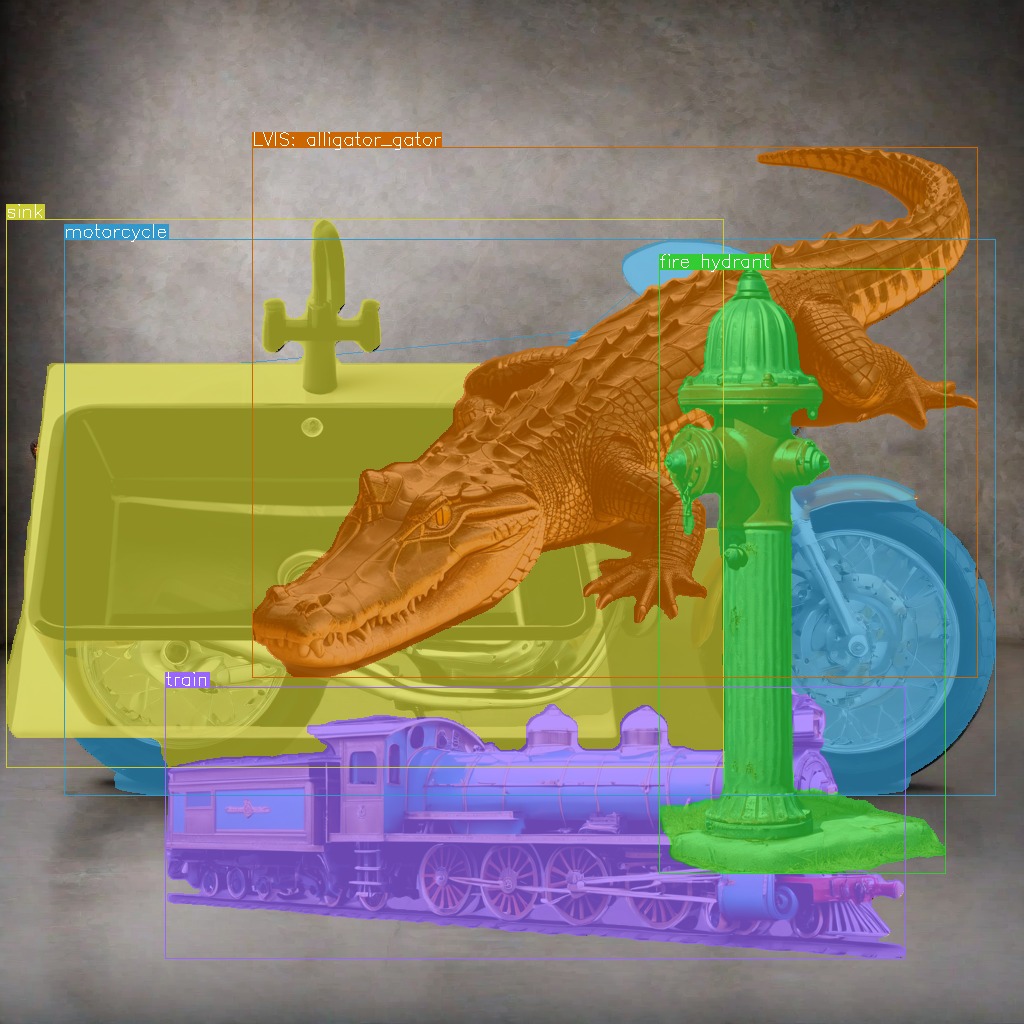} 
    } & 
    \parbox[c]{\MoredataBoxWidth}{
    \includegraphics[width=\MoredataImageWidth]{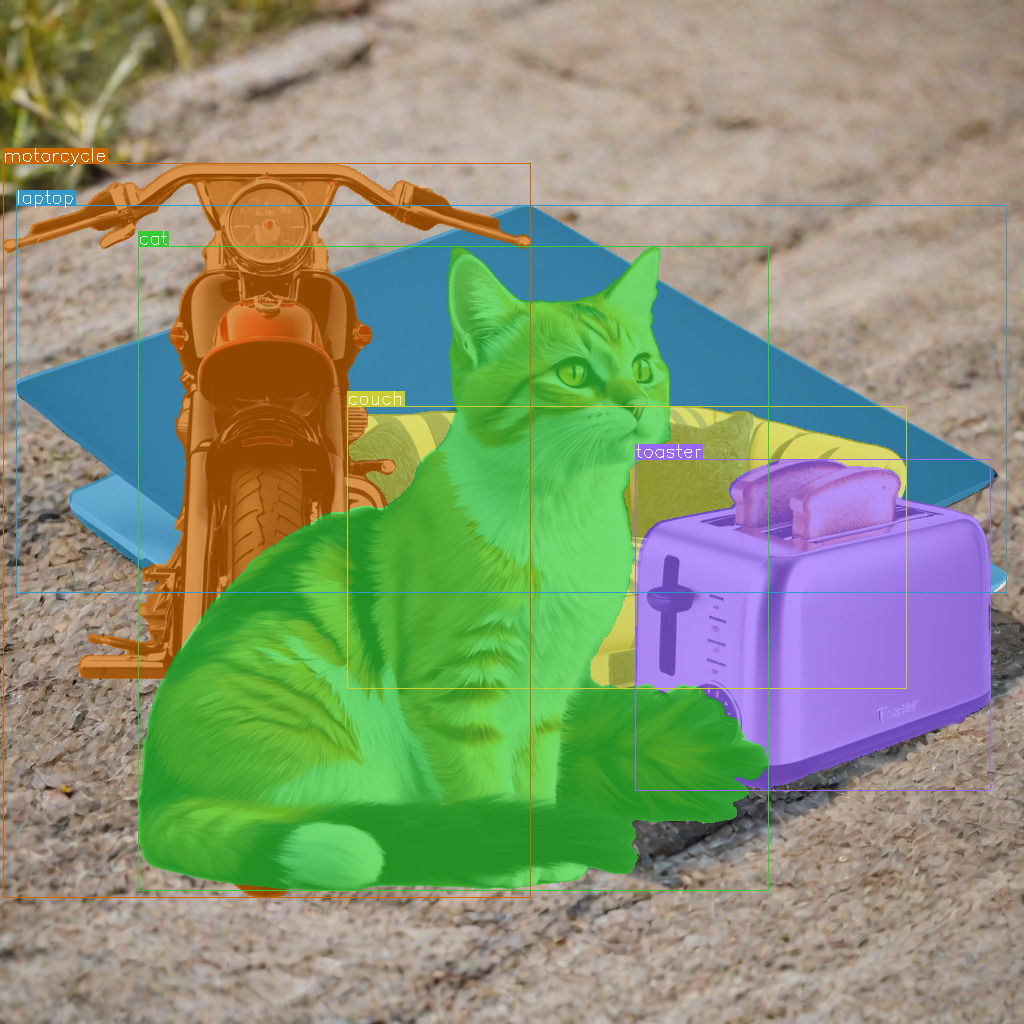} 
    } \\ 
    
    \parbox[c]{\MoredataBoxWidth}{
    \includegraphics[width=\MoredataImageWidth]{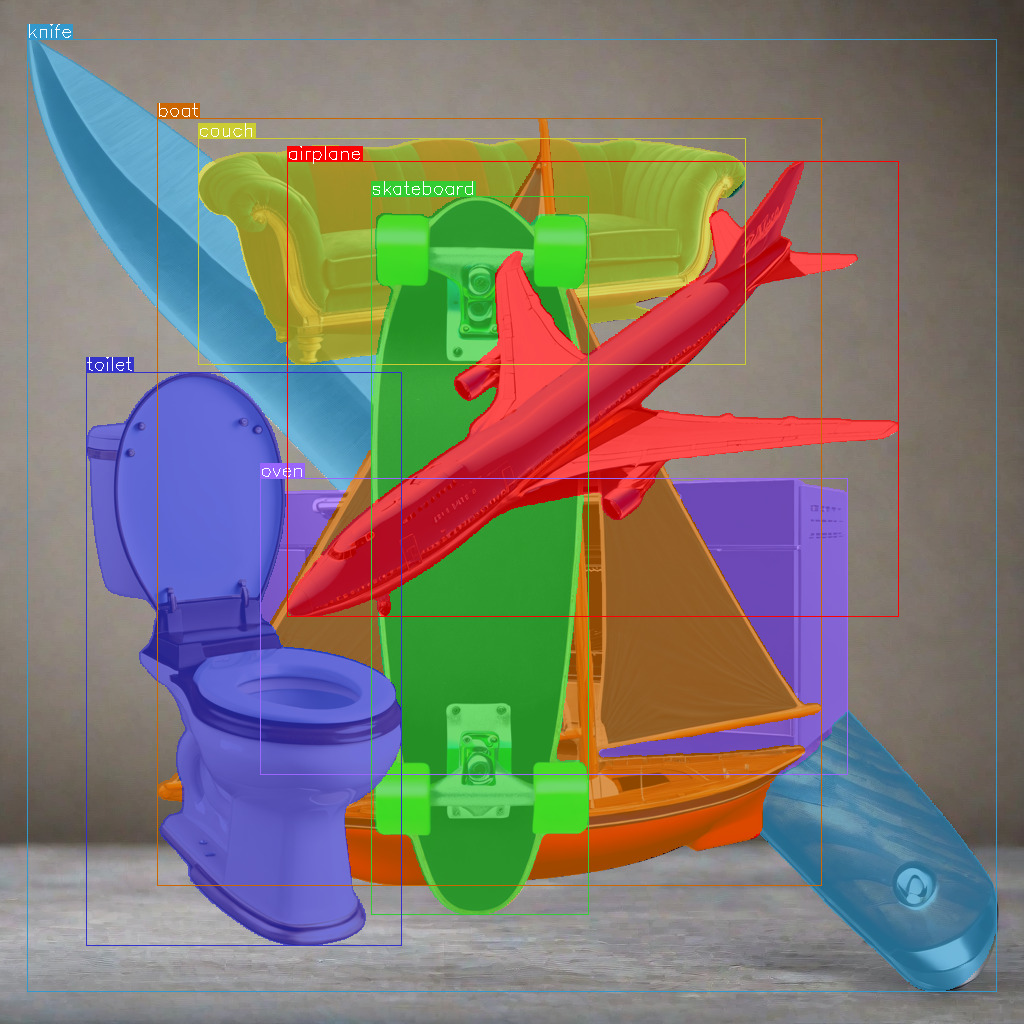} 
    } &
    \parbox[c]{\MoredataBoxWidth}{
    \includegraphics[width=\MoredataImageWidth]{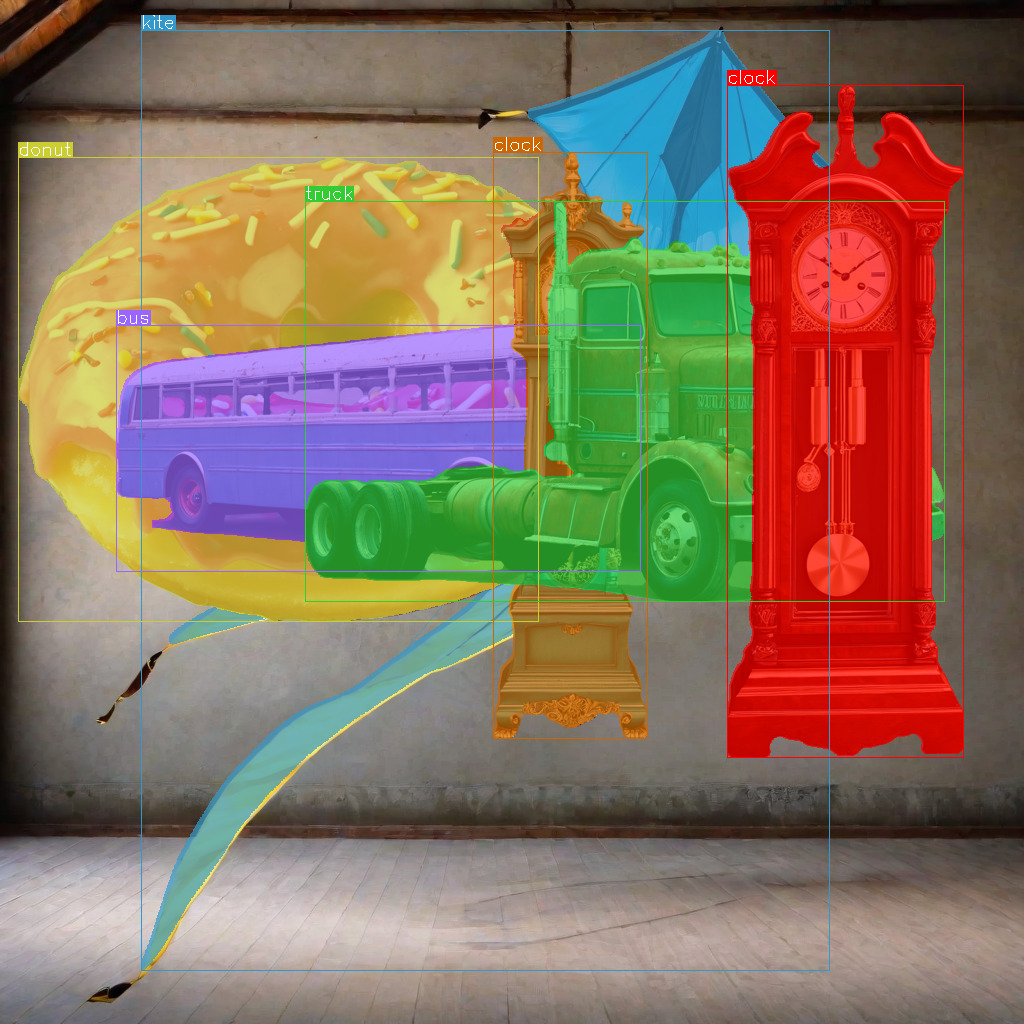} 
    } \\ 

    \parbox[c]{\MoredataBoxWidth}{
    \includegraphics[width=\MoredataImageWidth]{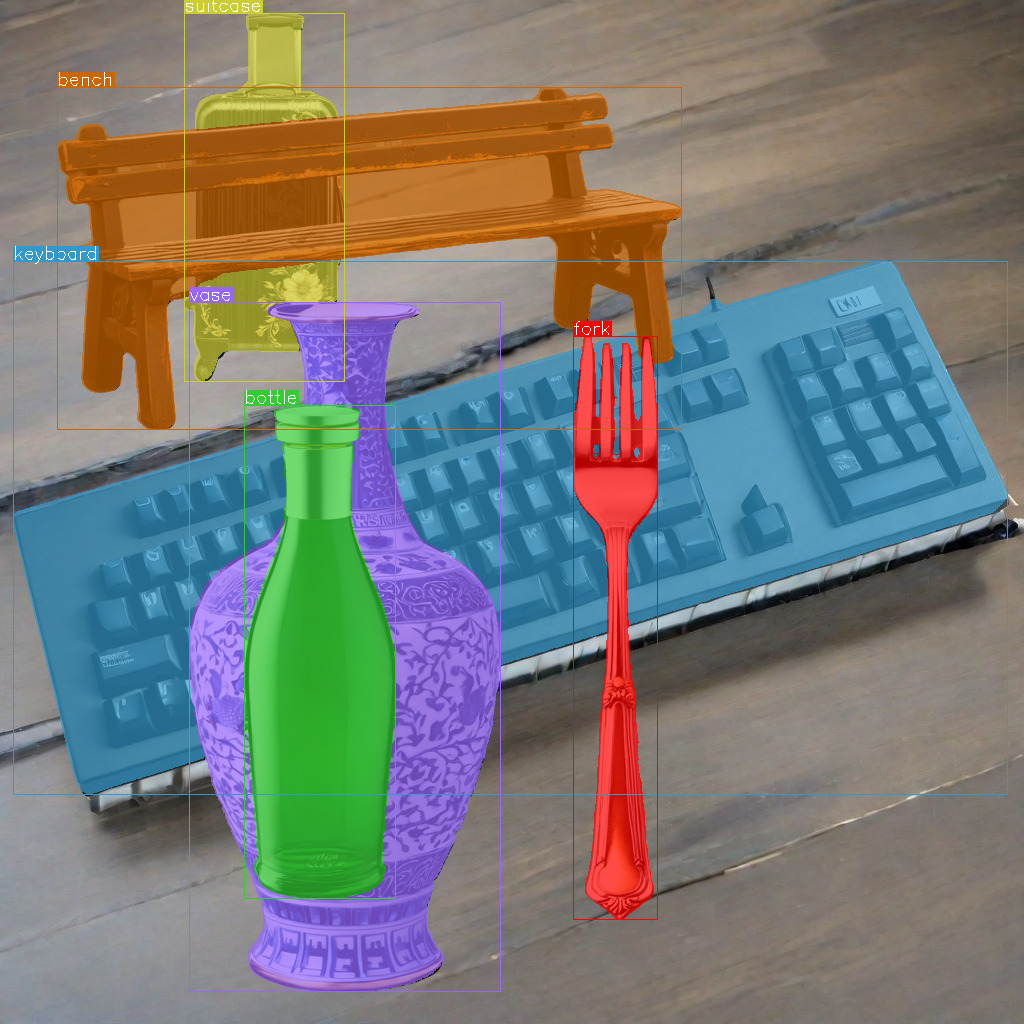} 
    } & 
    \parbox[c]{\MoredataBoxWidth}{
    \includegraphics[width=\MoredataImageWidth]{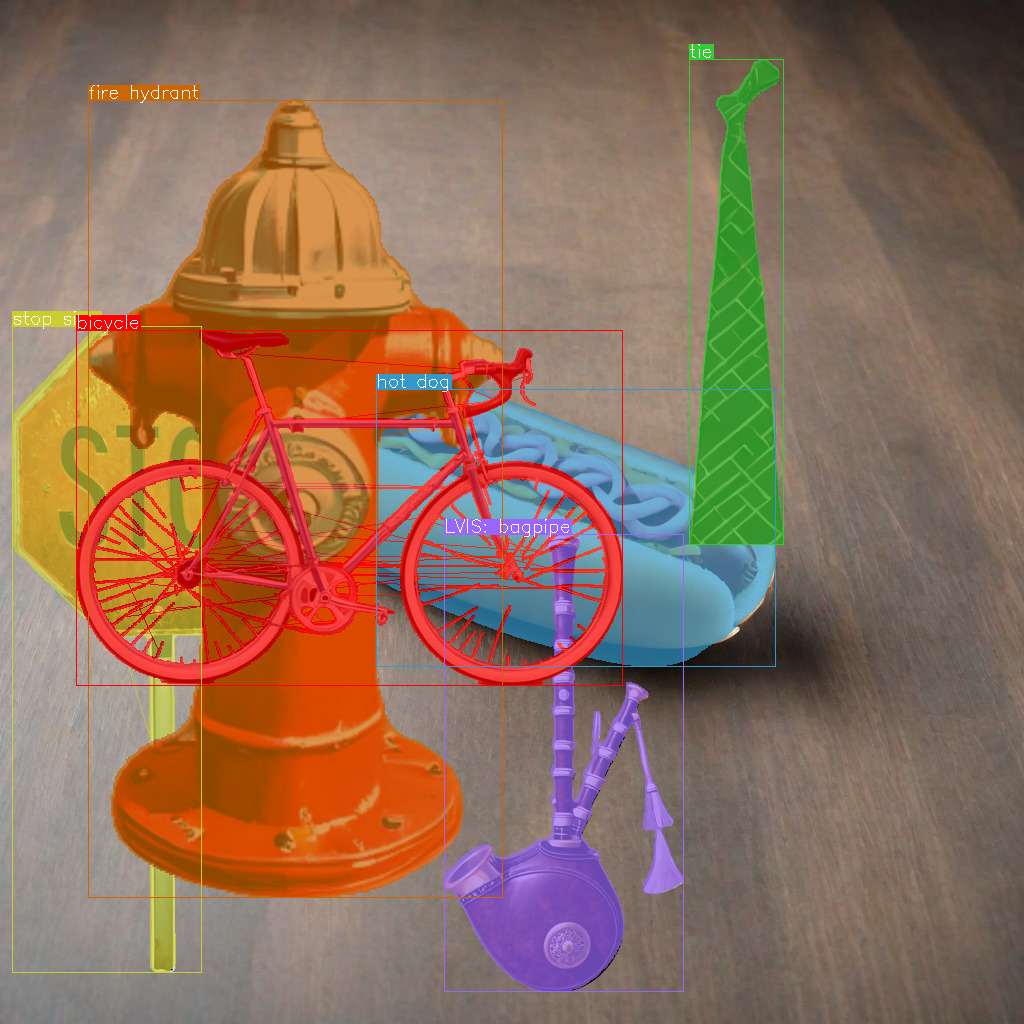} 
    } \\ 
    
  \end{tabular}
  \caption{\textbf{Additional Examples of Gen-n-Val Synthetic Data for COCO experiments.}}
  \label{fig:moredata5}
\end{figure*}

\begin{figure*}[ht!]
  \centering
  \begin{tabular}{@{}c@{}c@{}}

    \parbox[c]{\MoredataBoxWidth}{
    \includegraphics[width=\MoredataImageWidth]{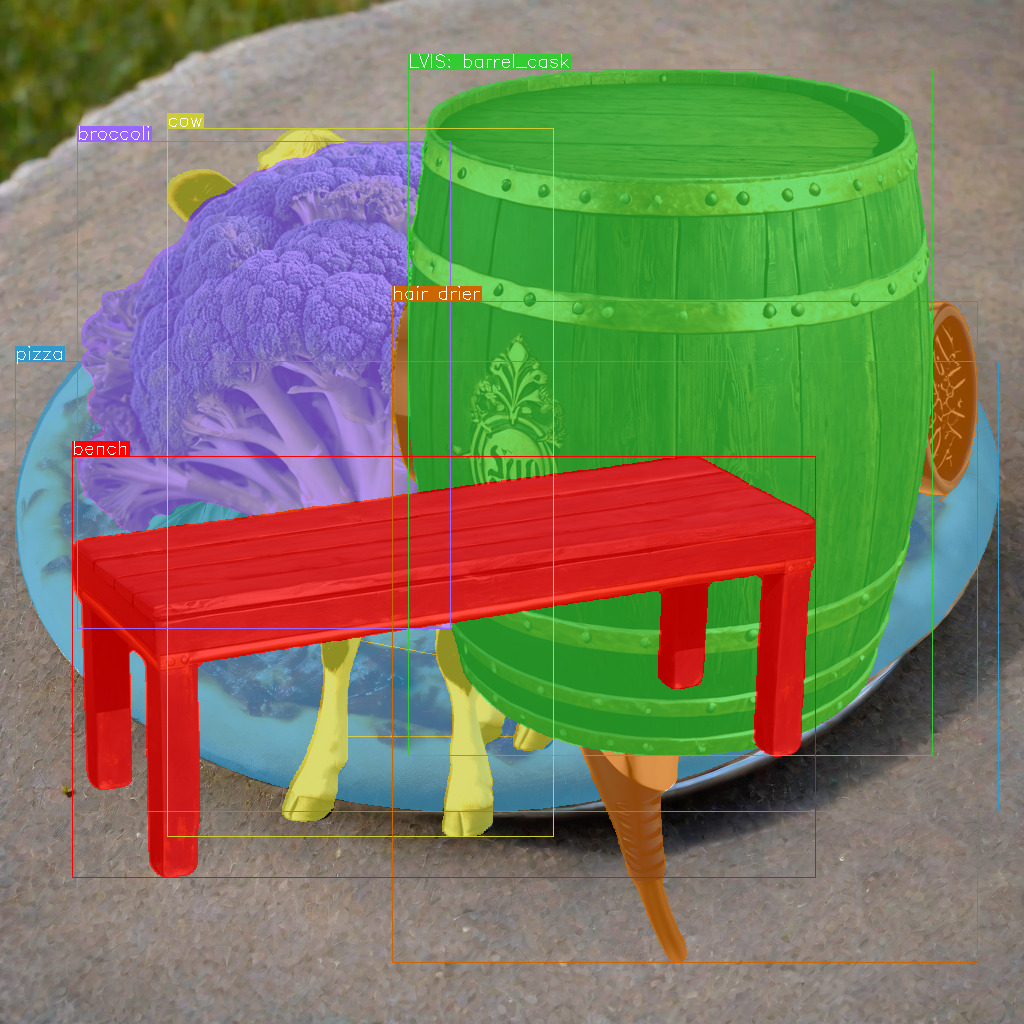} 
    } & 
    \parbox[c]{\MoredataBoxWidth}{
    \includegraphics[width=\MoredataImageWidth]{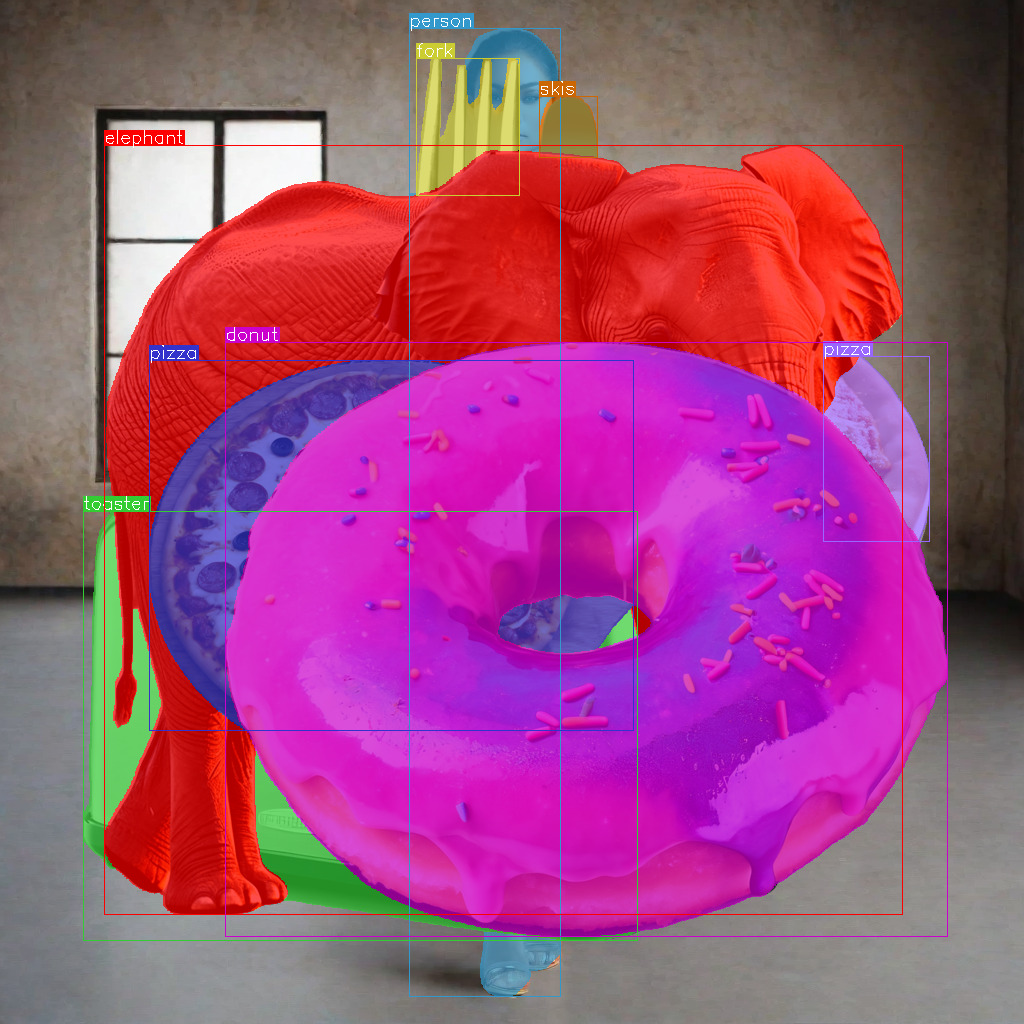} 
    } \\ 

    \parbox[c]{\MoredataBoxWidth}{
    \includegraphics[width=\MoredataImageWidth]{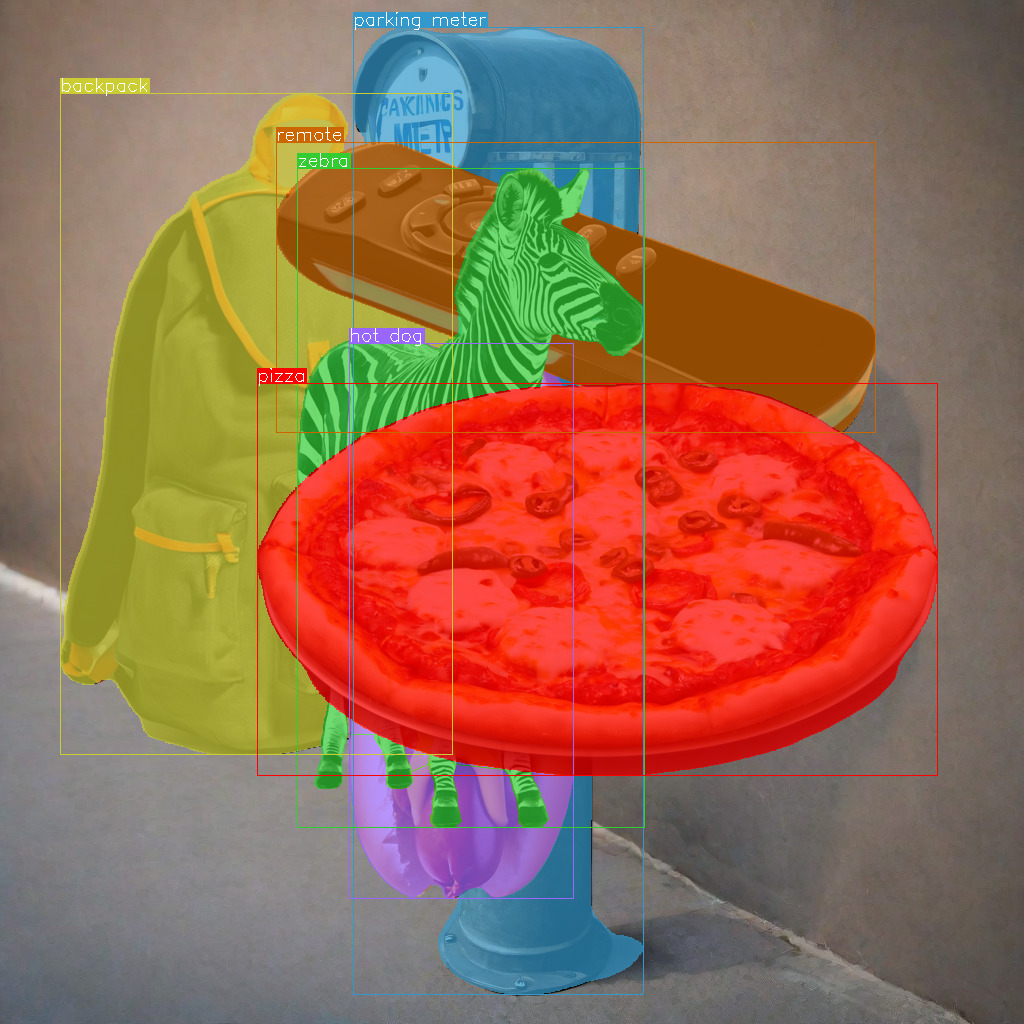} 
    } &
    \parbox[c]{\MoredataBoxWidth}{
    \includegraphics[width=\MoredataImageWidth]{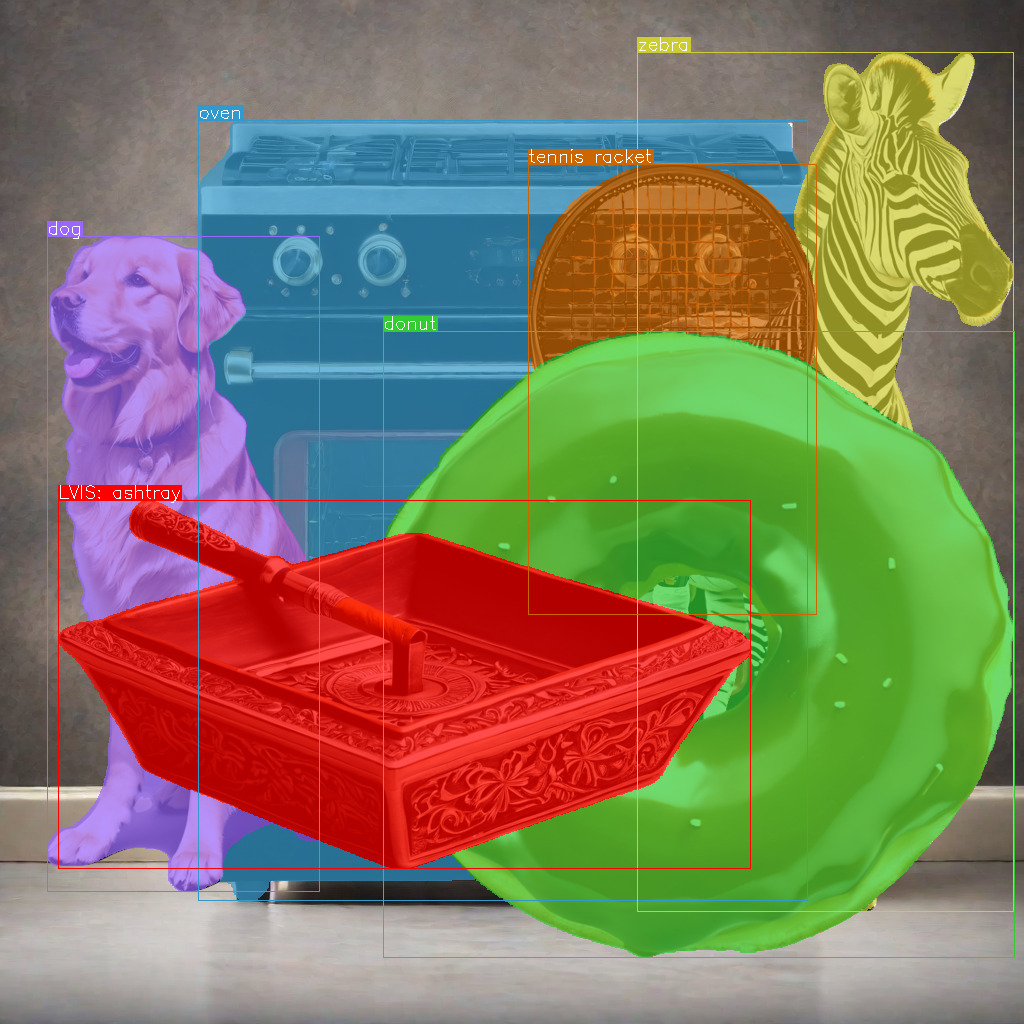} 
    } \\ 

    \parbox[c]{\MoredataBoxWidth}{
    \includegraphics[width=\MoredataImageWidth]{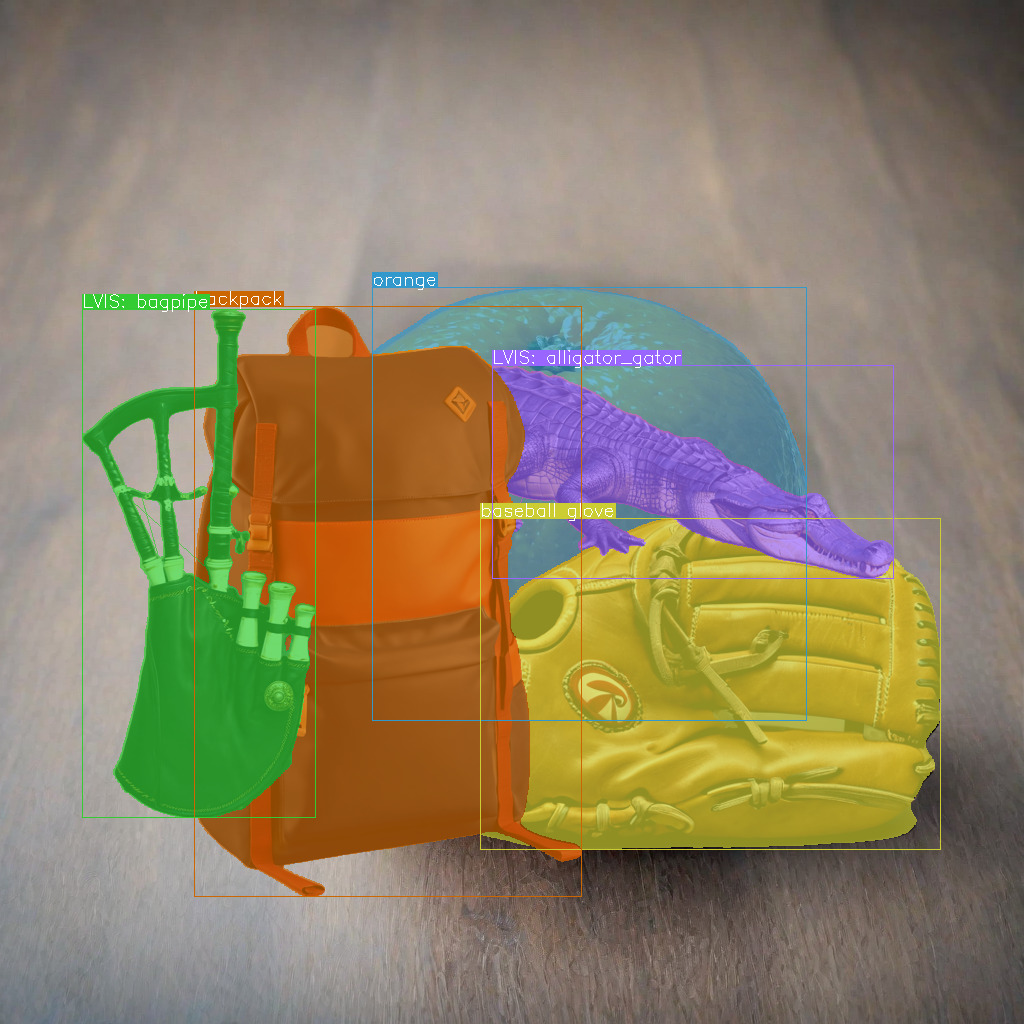} 
    } & 
    \parbox[c]{\MoredataBoxWidth}{
    \includegraphics[width=\MoredataImageWidth]{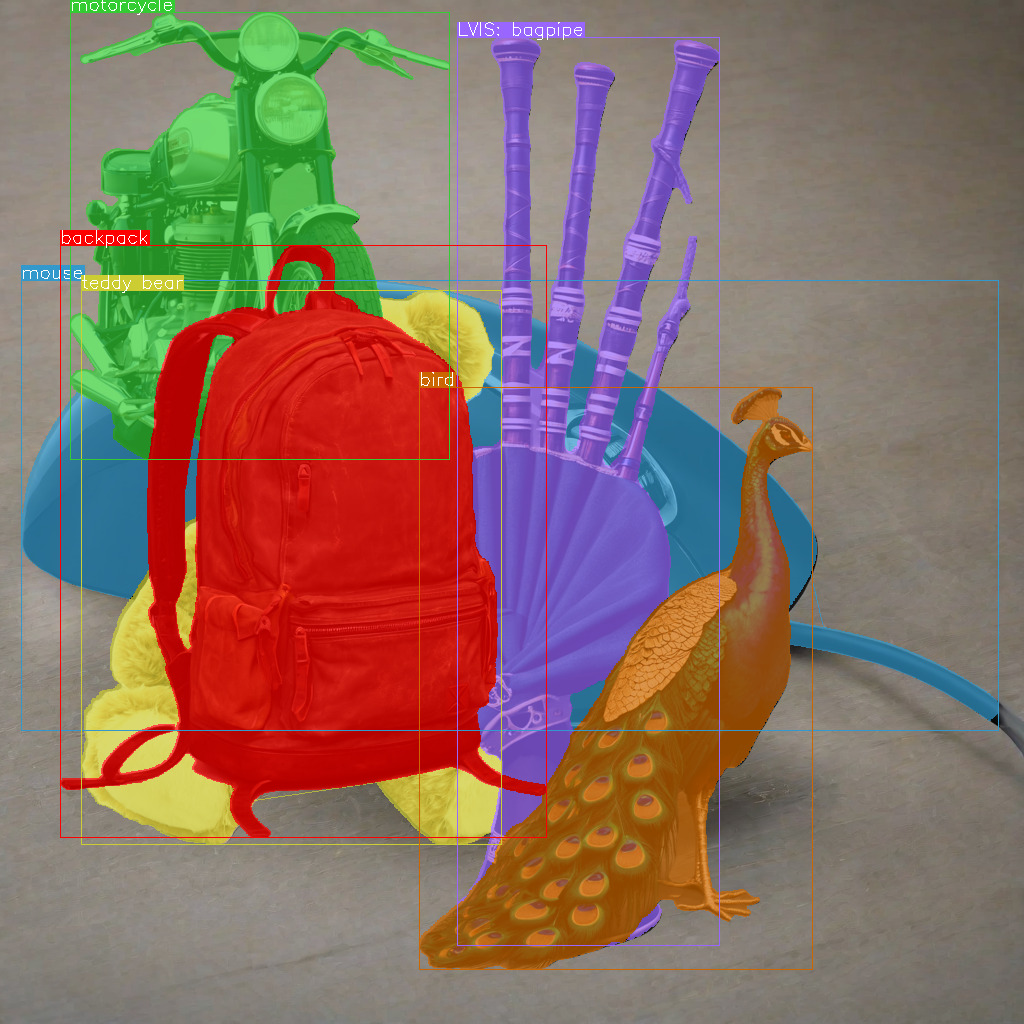} 
    } \\ 
    
  \end{tabular}
  \caption{\textbf{Additional Examples of Gen-n-Val Synthetic Data for COCO experiments.}}
  \label{fig:moredata6}
\end{figure*}

\begin{figure*}[ht!]
  \centering
  \begin{tabular}{@{}c@{}c@{}}

    Original LVIS & Augmented by Gen-n-Val \\

    \parbox[c]{\MoredataBoxWidth}{
    \includegraphics[width=\MoredataImageWidth]{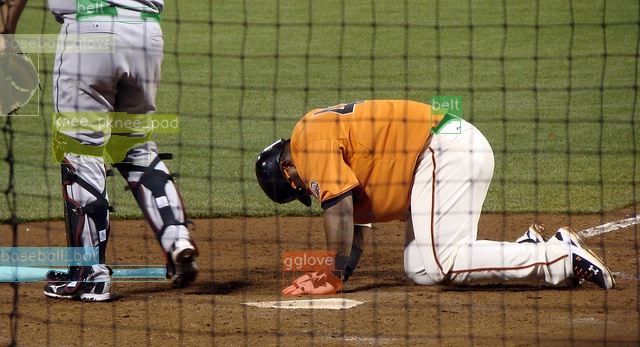} 
    } & 
    \parbox[c]{\MoredataBoxWidth}{
    \includegraphics[width=\MoredataImageWidth]{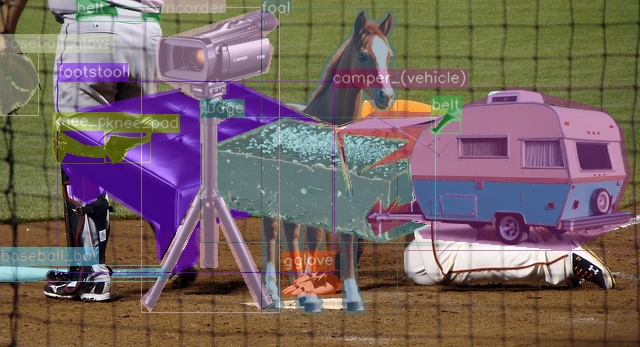} 
    } \\ 

    \parbox[c]{\MoredataBoxWidth}{
    \includegraphics[width=\MoredataImageWidth]{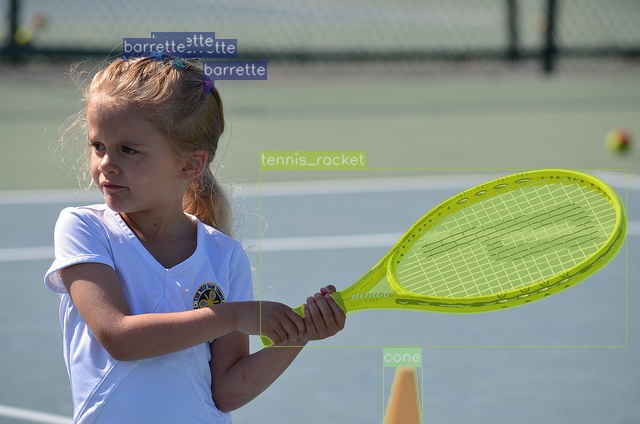} 
    } & 
    \parbox[c]{\MoredataBoxWidth}{
    \includegraphics[width=\MoredataImageWidth]{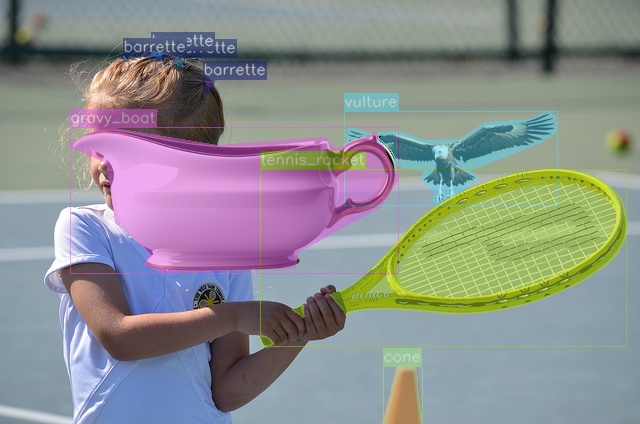} 
    } \\

    \parbox[c]{\MoredataBoxWidth}{
    \includegraphics[width=\MoredataImageWidth]{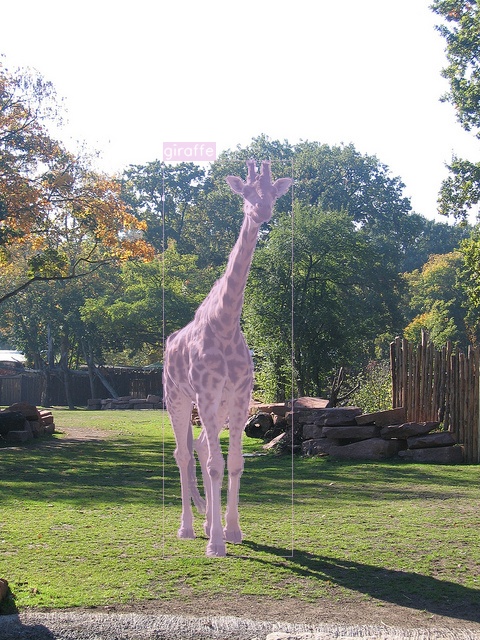} 
    } & 
    \parbox[c]{\MoredataBoxWidth}{
    \includegraphics[width=\MoredataImageWidth]{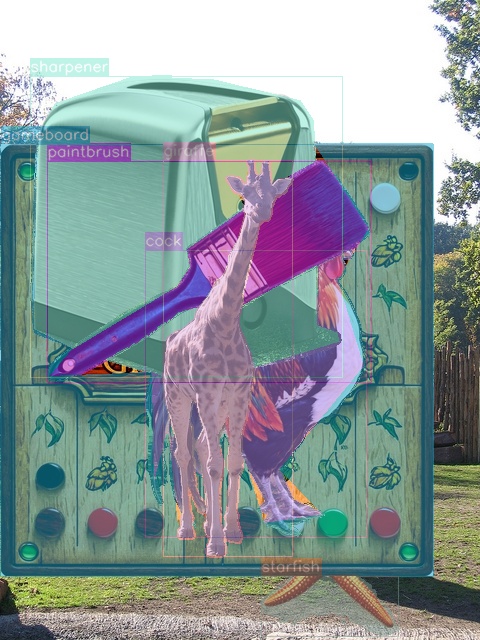} 
    } \\

    Original LVIS & Augmented by Gen-n-Val \\
    
  \end{tabular}
  \caption{\textbf{Additional Examples of Gen-n-Val Synthetic Data for LVIS experiments.}}
  \label{fig:lvisdata_1}
\end{figure*}

\begin{figure*}[ht!]
  \centering
  \begin{tabular}{@{}c@{}c@{}}

    Original LVIS & Augmented by Gen-n-Val \\

    \parbox[c]{\MoredataBoxWidth}{
    \includegraphics[width=\MoredataImageWidth]{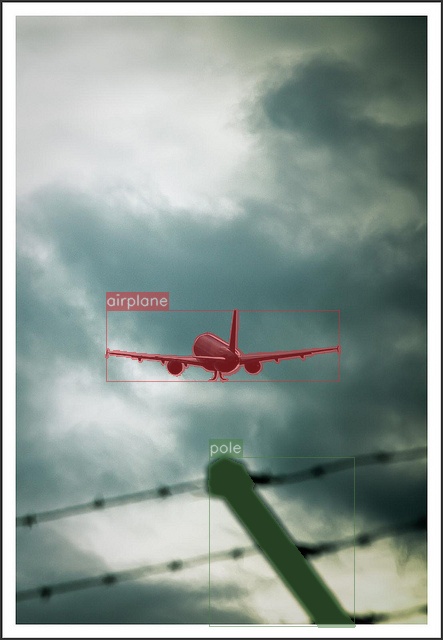} 
    } & 
    \parbox[c]{\MoredataBoxWidth}{
    \includegraphics[width=\MoredataImageWidth]{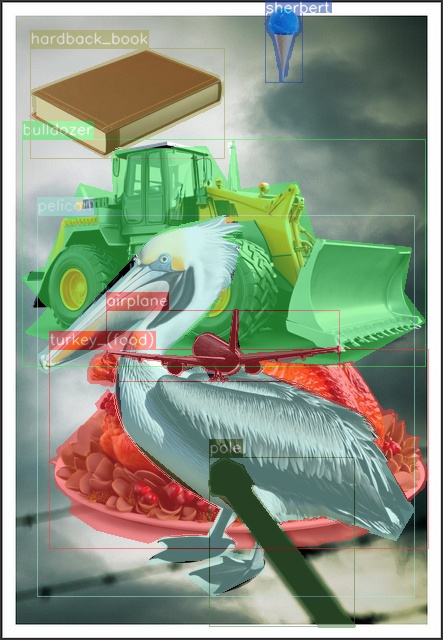} 
    } \\ 

    \parbox[c]{\MoredataBoxWidth}{
    \includegraphics[width=\MoredataImageWidth]{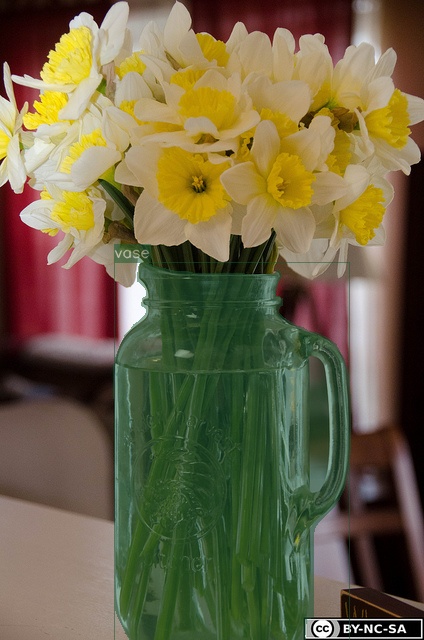} 
    } & 
    \parbox[c]{\MoredataBoxWidth}{
    \includegraphics[width=\MoredataImageWidth]{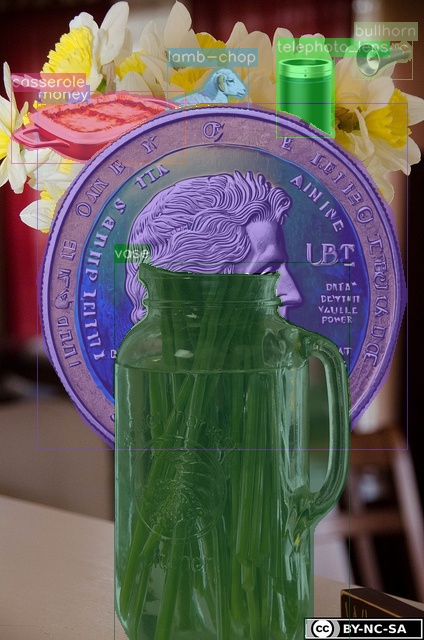} 
    } \\ 
    
    Original LVIS & Augmented by Gen-n-Val \\
    
  \end{tabular}
  \caption{\textbf{Additional Examples of Gen-n-Val Synthetic Data for LVIS experiments.}}
  \label{fig:lvisdata_2}
\end{figure*}

\begin{figure*}[ht!]
  \centering

  \begin{tabular}{@{}c@{}c@{}}

    % \hline
    Original LVIS & Augmented by Gen-n-Val \\

    \parbox[c]{\MoredataBoxWidth}{
    \includegraphics[width=\MoredataImageWidth]{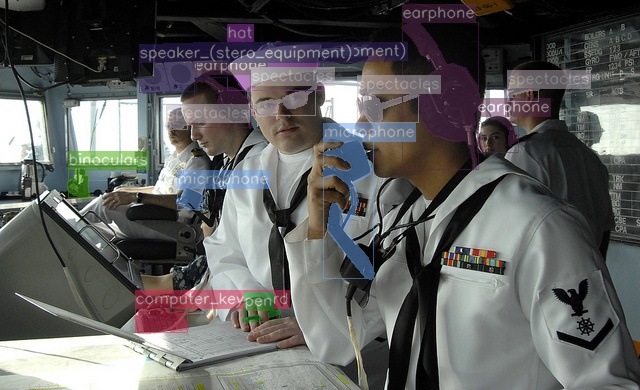} 
    } & 
    \parbox[c]{\MoredataBoxWidth}{
    \includegraphics[width=\MoredataImageWidth]{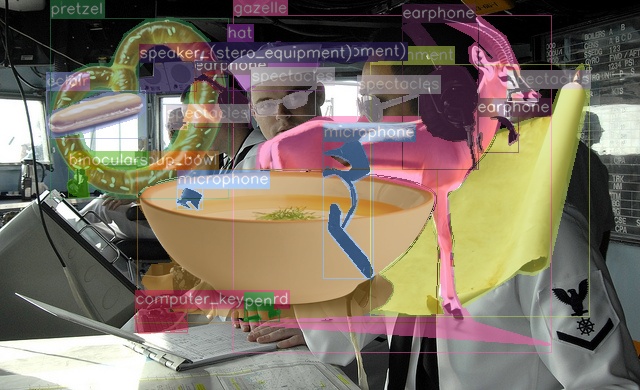} 
    } \\ 

    \parbox[c]{\MoredataBoxWidth}{
    \includegraphics[width=\MoredataImageWidth]{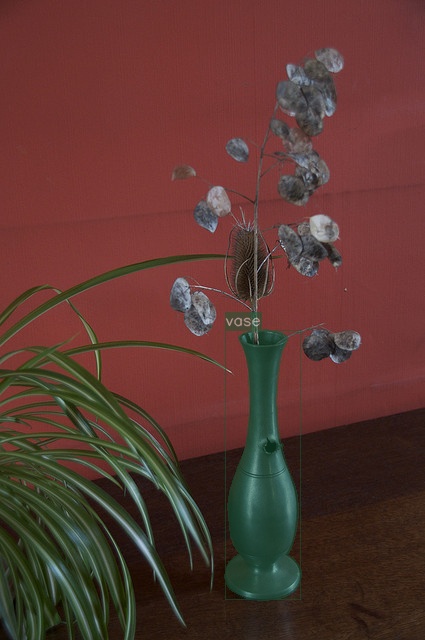} 
    } & 
    \parbox[c]{\MoredataBoxWidth}{
    \includegraphics[width=\MoredataImageWidth]{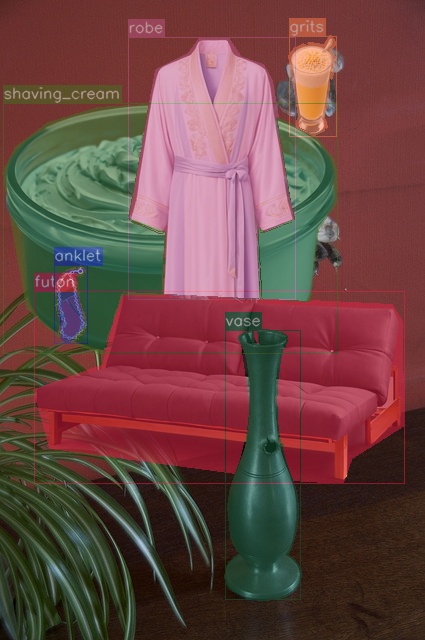} 
    } \\ 

    \parbox[c]{\MoredataBoxWidth}{
    \includegraphics[width=\MoredataImageWidth]{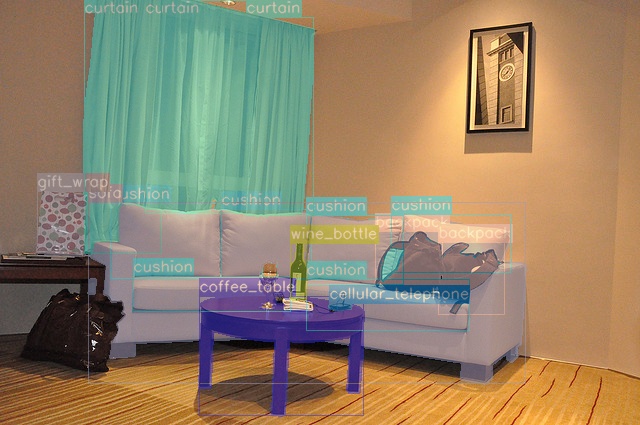} 
    } & 
    \parbox[c]{\MoredataBoxWidth}{
    \includegraphics[width=\MoredataImageWidth]{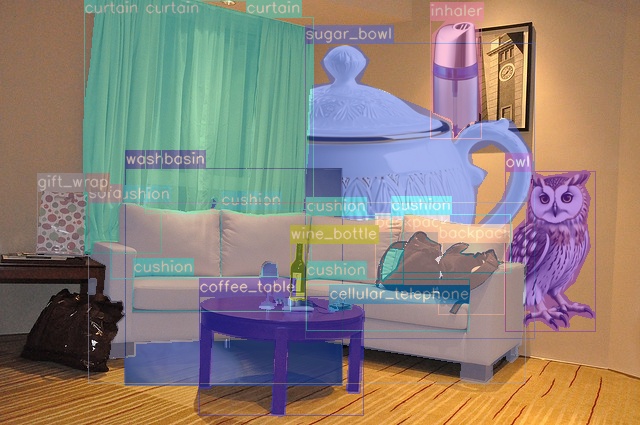} 
    } \\ 
    
    Original LVIS & Augmented by Gen-n-Val \\
    
  \end{tabular}
  \caption{\textbf{Additional Examples of Gen-n-Val Synthetic Data for LVIS experiments.}}
  \label{fig:lvisdata_3}
\end{figure*}